**PRACTICAL APPROACH TO KNOWLEDGE-BASED QUESTION ANSWERING WITH NATURAL LANGUAGE UNDERSTANDING AND ADVANCED REASONING**

WILSON WONG YIK SEN

A thesis submitted

in fulfillment of the requirements for the degree of Master of Science in Information and Communication

Technology

Faculty of Information and Communication Technology
KOLEJ UNIVERSITI TEKNIKAL KEBANGSAAN MALAYSIA
2005

# Dedication

To my soul mate, Joe and my family

*"In science, the credit goes to the man who convinces the world, not to the man to whom the idea first occurs"* by Sir William Osler

*"We can only see a short distance ahead, but we can see plenty there that needs to be done"* by Alan Turing

# Acknowledgements

Graduating with a Master's degree through research in any subject is a non-trivial task. With the support in certain aspects from the people mentioned below, life has been made a little easier. First of all, I would like to thank my supervisor Associate Professor Goh Ong Sing for giving me the opportunity to pursue this project. Going through this research alone is a very daunting task but with the companionship and assistance from other individuals in the University Press, things went along much smoother. Not leaving out people from the Faculty of Information and Communication Technology, Study Leave Unit and Postgraduate Study Centre, a very big hug for everyone. A special thanks to Dr. Mohammad Ishak Desa, Dr. Shahrin Sahib and Dr. Nanna Suryana. Thank you for always being there for my questions. I would also like to attribute whatever academic success that I have achieved to the Government of Malaysia. I would not have come this far without the opportunity and financial support by the Government since the very first day I pursued my undergraduate studies. Not forgetting, my family back in Penang. There are no words that could express my heartfelt-gratitude for the day-in day-out prayers and support from afar. With your blessings, I'm bound to achieve greatness from the very first day I left home. As for my pseudo-family members in Kuala Lumpur namely Shu Yueh, Ah Hoe, Sau Lun, Mr. and Mrs. Nakayama and others, a big thank you. On a more personal level, I would like to give a million hugs and kisses to my soul mate Liaw Sau Joe for sticking with me across ten thousand miles of ocean and being extremely forbearing with me when I get irritable from the pressure of this research. Finally, my appreciation goes to the city of Melaka for being such a great place to do research.

WILSON WONG

June 2005

# Table of Contents

**REFERENCES**

**APPENDIX**

# List of Figures

# List of Tables

# List of Abbreviations

| | |
|---|---|
| TREC | *T*ext *RE*trieval *C*onference, an annual conference and competition co-sponsored by the National Institute of Standards and Technology, and U.S. Department of Defense. |
| BBN | *B*olt, *B*eranek and *N*ewman, the last names of the three founders of BBN Technologies and the original name of the company. |
| SHRDLU | The name comes from the fact that the frequency order of letters in English is ETAOIN*SHRDLU*. As a result, the arrangement of the keys on Linotype typesetting machines was ETAOIN on the first column and *SHRDLU* the second. |
| QA | *Q*uestion *A*nswering |
| NLP/NLU | *N*atural *L*anguage *P*rocessing/*N*atural *L*anguage *U*nderstanding |
| IR | *I*nformation *R*etrieval |
| NaLURI | *Na*tural *L*anguage *U*nderstanding and *R*easoning for *I*ntelligence |
| MUSE | *MU*lti-*S*ource *E*ntity, an information extraction system to perform named entity recognition on diverse types of text with minimal adaptation. |
| FASTUS | *F*inite *S*tate *A*utomata-based Text *U*nderstanding *S*ystem, an information extraction system by Stanford Research Institute. |
| LISP | *LIS*t *P*rocessing, a functional programming language family originally developed as a practical computation model and later became the favored artificial intelligence research language. |
| HTML | *H*yper*T*ext *M*arkup *L*anguage |
| XI | *X* stands for cross-classification and *I* for inheritance |

# Related Publications

Wilson Wong, Shahrin Sahib & Ong-Sing Goh. *Evaluation of Response Quality for Heterogeneous Question Answering Systems*. Accepted to the IEEE/WIC/ACM International Conference on Web Intelligence (WI 2005), University of Technology of Compiegne, France, 19-22 September 2005.

Wilson Wong, Shahrin Sahib & Ong-Sing Goh. NaLURI: Question Answering with Natural Language Understanding and Network-based Advanced Reasoning. Submitted for review to the International Conference on Intelligent Technologies (InTech 2005) in July 2005.

Wilson Wong, Shahrin Sahib & Ong-Sing Goh, "Question Answering Approaches in the 21$^{st}$ Century: A Survey", Submitted to the ACM Computing Survey for review in April 2005.

Wilson Wong, Shahrin Sahib & Ong-Sing Goh. *Response Quality Evaluation in Heterogeneous Question Answering System: A Black-box Approach.* Unpublished manuscript, Kolej Universiti Teknikal Kebangsaan Malaysia, 2005.

Wilson Wong, Halizah Basiron, Shahrin Sahib and Ong-Sing Goh, "Intelligent Responses through Network-based Answer Discovery with Advanced Reasoning", To appear in the Proceeding of the IASTED International Conference on Computational Intelligence (CI 2005), Alberta, Canada, 4-6 July.

Wilson Wong, Ong-Sing Goh, Mohamad-Ishak Desa and Shahrin Sahib, "Online Cyberlaw Knowledge Base Construction Using Semantic Network", Ed. M.H. Hamza. In Proceeding of the IASTED

International Conference on Applied Simulation & Modeling (ASM 2004), Rhodes, Greece, 28-30 June 2004. ISBN: 0-88986-401-2, pp. 347-352.

Wilson Wong, Ong-Sing Goh and Mokhtar Mohd-Yusof, "Syntax Preprocessing in Cyberlaw Web Knowledge Base Construction", Ed. M. Mohammadian. In Proceeding of the International Conference on Computational Intelligence for Modelling, Control and Automation (CIMCA 2004), Gold Coast, Australia, 12-14 July 2004. ISBN: 174-088-1893, pp: 174-184.

# Abstract

The complexity of natural language and the open-domain nature of the World Wide Web have caused modern-day question answering systems to rely only on information retrieval techniques and shallow natural language processing tasks. This approach has brought about serious drawbacks namely restriction on the nature of question and response. This restriction constitutes the first problem addressed by this research. Through recent academic works, many researchers have begun to acknowledge the problem and agreed that the solution comes in the form of a new approach based on natural language understanding and reasoning in a knowledge-based environment. Due to the infancy stage of this new approach and practical consideration, the current practices vary greatly and are mostly based on only low-level natural language understanding, minimalist representation formalism and conventional reasoning approach without advanced features. As a result, not only were these systems found to be inadequate to solve the first problem but have also created the second problem, that is the limitation to scale across domains and to real-life natural language text. This research hypothesized that a practical approach in the form of a solution framework which combines full-discourse natural language understanding, powerful representation formalism capable of exploiting ontological information and reasoning approach with advanced features, will solve both the first and second problem without compromising practicality factors. The solution framework is implemented as a system called *"Natural Language Understanding and Reasoning for Intelligence"* (NaLURI). More importantly, two evaluations and their results are presented to demonstrate that the inclusion of more demanding features into a question answering system will not only allow for a wider range of questions and better response quality, but does not affect the response time, hence approving the hypothesis of this research.

# Abstrak

*Kerumitan dalam pengendalian bahasa tabii serta sifat-sifat domain terbuka dalam Jaringan Sedunia menyebabkan sistem soal jawab zaman moden tiada pilihan selain daripada bergantung sepenuhnya pada teknik carian maklumat berasaskan kata kunci serta pemprosesan bahasa tabii yang sempit. Pendekatan seperti ini menyebabkan kebolehan sistem untuk menjawab soalan terhad. Batasan ini merupakan masalah pertama yang cuba diselesaikan dalam penyelidikan ini. Melalui penulisan akademik yang terkini, para penyelidik mula mengakui kewujudan masalah tersebut dan bersetuju bahawa jalan penyelesaiannya terangkum dalam satu pendekatan baru yang berasaskan teknik-teknik pemahaman bahasa tabii dan taakulan yang berasaskan pengetahuan. Disebabkan oleh pelbagai pertimbangan dari segi isu-isu praktikal, amalan sedia ada kebanyakannya berasaskan hanya pemahaman bahasa tabii sempit, perwakilan formalisma yang minimum dan pendekatan taakulan konvensional tanpa ciri-ciri canggih. Akibatnya, sistem-sistem ini bukan sahaja didapati tidak sesuai untuk menyelesaikan masalah pertama dalam penyelidikan ini, malah amalan-amalan tersebut telah membawa kepada masalah kedua iaitu pembatasan dalam merentasi domain dan teks bahasa tabii yang sebenar. Penyelidikan ini mencadangkan satu pendekatan praktikal dalam bentuk rangka kerja yang akan menyelesaikan masalah pertama serta masalah kedua melalui pemahaman bahasa tabii dan wacana yang lengkap dengan perwakilan formalisme yang optimum seperti rangkaian semantik yang mampu mengeksploitasi maklumat ontologi untuk membolehkan pengenalan ciri-ciri canggih ke dalam pendekatan taakulan. Rangka kerja penyelesaian tersebut direalisasikan melalui "Natural Language Understanding and Reasoning for Intelligence" (NaLURI). Dua penilaian bagi menguji isu-isu praktikal juga dibuat bagi menunjukkan bahawa pengenalan ciri-ciri canggih ke dalam sistem soal jawab bukan sahaja akan meningkatkan kualiti soal-jawab, malah tidak akan menjejaskan masa tindak balas.*

# Chapter 1

# Introduction

## 1.1 The Past, Present and Future of Question Answering

The common idea in question answering is to be able to provide responses to questions written in natural language (i.e. English) by finding the answer in some sources (e.g. web pages, plain texts, knowledge bases) or by generating explanations in the case of failures (i.e. which is only possible through intelligent approaches). Unlike information retrieval applications like web search engines, the goal is to find a specific answer (Lin *et al.,* 2003) rather than flooding the users with documents or even best-matching passages as most information retrieval systems currently do. With the increase in the number of online information seekers, the demand for automated question answering systems has risen accordingly. There are many ways of looking at question answering depending on the approaches towards the various dimensions (Hirschman & Gaizauskas, 2001). The different dimensions include question, response, technique, information source, domain and evaluation as depicted in Figure 1.1.

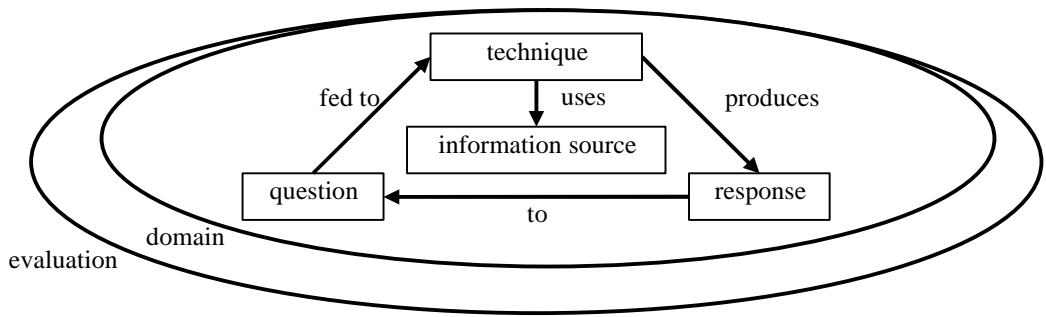

Figure 1.1: Dimensions of question answering

As different type of questions pose dissimilar level of challenges, the type of questions supported by a question answering system can be used to determine the strength of the system. Questions can be formulated in five ways (Moldovan *et al.*, 2002) namely factual questions (e.g. *"Where is Kuala Lumpur"*), questions requiring simple reasoning (e.g. *"Why did the accident happen?"*, synthesis-based questions (e.g. *"What are the daily activities of the victim a week before he was murdered?"*), dialogue-based questions (e.g. *"Who is the defendant in that case?"*) and finally, speculative questions (e.g. *"Is the idea of raising fuel price justified?"*).

Unlike questions, there are no definitions of what encompasses an exact response. Clearly an answer has to be correct to be of any use, but this still leaves a lot of scope for different systems to present the same answer in many different ways. Nevertheless, from the techniques employed for producing answers, one can almost predict the structure of the response. Systems that use unstructured texts as their source of answers for example, will usually return a short extract from the text as responses. The major question with such systems is how long the returned answer should be.

The following two dimensions, namely technique and information source, are the common aspects used to differentiate between the various types of question answering systems since the beginning of the question answering era. The technique used is usually highly related to the type of information source used. If the information source is free-text, then the technique will most likely be based on some information retrieval approach. On the other hand, if the information source is knowledge base or database, then the approach will be either logic-based or some language sanctioned by the knowledge base or database.

Domain is one of the dimensions that determines the focus or direction of a question answering system. Open-domain question answering practices techniques based on probabilistic measures and has a wider range of information source. It is very likely that the techniques are more logic-based and well-founded with relatively limited sources for question answering that focuses on certain domains as compared to open-domain. A domain-oriented question answering system deals with questions under a specific domain and can be seen as a richer approach because natural language processing systems can exploit domain knowledge and ontologies. Advanced reasoning such as providing explanation for answers and generalizing questions is not possible in open-domain systems. Open-domain question answering systems need to deal with questions about nearly everything and it is very difficult to rely on ontological information due to the absence of wide and yet detailed world knowledge. On the other hand, these systems have much more data to exploit in the process of extracting the answers (Clarke *et al.*, 2001).

The last dimension, which is evaluation, can be rather subjective especially when dealing with different types of natural language systems in different domains. Surprisingly, the literatures on evaluation are relatively sparse given its state of importance and are mostly available in form of evaluating general natural language systems (King, 1996). It is easy to evaluate systems in which there is a clearly defined answer, however, for most natural language questions there is no single correct answer. Only the question answering systems based on shallow natural language processing and information retrieval that have the corpora and test questions readily available for example, can use recall and precision as evaluation criteria. The question answering track of TREC is a good example (Voorhees, 2003). The task of evaluating the system can be more subjective and difficult for other domain-oriented question answering.

### 1.1.1 Domain-Oriented Question Answering in Cyberlaw

Like many other question answering systems based on natural language understanding and reasoning, the choice of domain tend to be focused in certain areas. Some examples are the question answering system for biomedicine by Zweigenbaum (2003), question answering system for weather forecast by Chung *et al.*

(2004) and question answering system for the tourism domain by Benamara (2004). These domain-oriented question answering systems not only act as real-life example of the success of the natural language understanding and reasoning approach, but the systems itself proved to be a worthwhile attempt in providing intelligent assistant for domain experts.

While there is already a number of domain-oriented question answering systems out there based on a wide range of techniques in natural language understanding and reasoning covering unique domains, a system for the domain of Cyberlaw is yet to exist. This late emergence can be attributed to the fact that unlike other conventional domains which have been around for quite some time such as medicine, tourism and other branches of science and economics, Cyberlaw only surfaces during the boom of the various activities related to the World Wide Web like e-commerce, e-banking, etc. The description of an initial effort towards a question answering system for the Cyberlaw domain is presented by Wong *et al.* (2004a).

Activities involving the use of Internet and information technology have increased tremendously over the years. This is particularly true as more companies and countries are attempting to use technologies like e-commerce, e-marketing, e-government, telemedicine and many more to achieve better efficiency and a paperless environment. With the increasingly important role played by Information Technology and the Internet, security threats and human misconducts will follow suit, creating a whole new paradigm of online information on Cyberlaw (Zahri & Ahmad-Nasir, 2003). In Malaysia alone for example, the year 2004 sees a rise in cyber crime from 856 cases in January to 1393 cases in December (Anon., 2004). As more parts of our life become acquainted to technology, a unified and easily accessible source of knowledge on Cyberlaw will be a valuable asset to legal practitioners, legal students, academicians, enthusiasts and others alike.

Despite the hype surrounding the field of Cyberlaw, there is yet to be any definite description of the scope of Cyberlaw but nonetheless, Grossman (1999) offers a credible argument which describes Cyberlaw as a fusion between computer law, Internet law, e-commerce law, intellectual property law, traditional contract law, criminal law, litigation of technology related disputes and much more. In a nutshell, Cyberlaw concerns what technology does or can do. To illustrate this point further, the following examples should be considered. We can earn money with technology, thus we have e-commerce law and traditional contract law in the mix. We can find new ways to digitally lose privacy online and we may relate that to an Internet law issue. Technology can also give us new ways to commit crimes and infringe copyrights, so we have to include criminal law and intellectual property law.

### 1.1.2 The Early Days of Question Answering

Some of best-known question answering systems in the early days were designed to provide natural language front ends to databases. These systems operated under extremely limited domains. Some of the best-known were BASEBALL (Green *et al.*, 1963), LUNAR (Woods, 1973) and LIFER (Hendrix *et al.*, 1978). The BASEBALL system was designed to answer questions about baseball games which had been played in the American league over a single season, while LUNAR was designed to enable lunar geologists to conveniently access, compare and evaluate the chemical analysis data on lunar rock and soil composition that was accumulating as a result of the Apollo moon mission. Similarly, LIFER employs a front-end natural language interface to connect to databases allowing users to ask questions about United States' navy ships. The other two systems that had an equal share of fame during the 1970s are SHRDLU by Winograd (1972) and GUS (Bobrow *et al.*, 1977). GUS was designed to simulate a travel advisor and has access to a database containing limited information about airline flight times. SHRDLU, the better known between the two, was created for manipulating geometrical blocks in a confined world. The difference between LUNAR and systems like SHRDLU and GUS is the latter's ability to carry out a dialogue.

Problems started to surface when some researchers during the early days attempted to apply their limited natural language interface to more general English text. This is mainly due to the expensive requirements for understanding and reasoning, and given the state-of-the-art technology during the early days, their approaches were only feasible in very limited domains (Hirschman & Gaizauskas, 2001). Moreover, the researchers in the field of natural language understanding during the early days were only starting to solve isolated problems in the lower level of linguistic analysis (Mueller, 1999). Due to the limiting factors in understanding natural language during the late 70s, a shift in the question answering approach began to take place. Rather than focusing on natural language understanding, researchers have opted for approaches that are known to be effective during that time, allowing them to move beyond domain restriction and exploit open-domain, natural language information (Fischer, 2003). These systems employ what is known to work best with free-text documents, namely information retrieval which typically relies on statistical methods to process the keywords in a query and calculate a relevance ranking. This ranking is used to search an open domain of texts to return a list of documents that possibly contain an answer. The process requires little or no linguistic knowledge because it relies primarily on word frequencies. However, many researches like Cardie *et al.* (2000) have examined and agreed that even weak linguistic knowledge can significantly improve the results of a question answering system. Thus, the marriage between shallow natural language processing and information retrieval for open-domain information marks the beginning of the modern-day question answering systems.

One of the first few systems that exhibit the modern-day question answering characteristics is MURAX by Kupiec (1993). MURAX employs an encyclopedia as the open-domain source and an information retrieval system for accessing it. Shallow linguistic analysis is performed using a part-of-speech tagger and finite-state recognizers for matching syntactic patterns. FAQ Finder by Burke *et al.,* (1997) is also another type of modern-day question answering system that works on frequently-asked questions found in newsgroups using the SMART information retrieval systems (Buckley, 1985). In 1999, many more systems from this camp like YorkQA (Alfonseca *et al.*, 2001) began to appear when TREC-8 and its subsequent conferences provided large corpuses as the underlying source for developing and evaluating question answering systems.

As the demand for a better search and retrieval solution to the ever-growing World Wide Web increased, researchers began to look into the exploitation of information on the World Wide Web as the source for question answering. With the wide availability of web search engines, modern-day question answering systems using classical information retrieval were very quickly extended to the World Wide Web. Some of the well known systems that exploit the web search engines are like Webclopedia (Hermjakob, 2001), AnswerBus (Zheng, 2002b) and MULDER (Kwok *et al.*, 2001). This modern-day question answering approach has indeed lived up to its name and has flourished until the present day. Consequently, many of the current researches tackle the problem of question answering from the dimension where the technique is based on the marriage of shallow natural language processing and information retrieval, and the information source using either TREC corpora or the World Wide Web.

### 1.1.3 Limitations of Modern-Day Question Answering

Through a review of the existing question answering systems based on shallow natural language processing and information retrieval which is discussed in Chapter 3, it can be seen that the ubiquitous ways of accessing information on the World Wide Web is web search engines. This has provided many modern-day question answering systems with an easy way out. These modern-day question answering systems have become too reliant on the use of web search engines and the idea that endless source of open-domain information is just a click away. This has resulted in the obsession towards shallow natural language processing and information retrieval (Diekema *et al.,* 2004), leaving no room for other more drastic approaches. Even though we can consider this advancement as something to be proud of but, they are far from being the ultimate solution in question answering because problems related to the handling of natural language are far from being solved using mere shallow natural language processing and information retrieval.

Undoubtedly, we acknowledge that open-domain question answering is a hard task because no restriction is imposed either on the question type or on the user's vocabulary. No matter how simple one might regard these systems as, they work considerably well given the size of the World Wide Web. There are three reasons for the success of question answering systems even in the face of openness in domain. Firstly, the questions handled by these systems are limited to factual questions namely those questions using interrogative words like *"who"*, *"when"*, *"where"* and *"what"*. These questions are the simplest in the hierarchy of questions (Moldovan *et al.*, 2002) and once the system is able to detect these interrogative words, it already has half the answer. The other half remains in its ability to detect proper names or named-entity, formulate queries using these words and let the web search engines do their work. Like the phrase from the television series the 'X-Files', *"the truth is out there"*, the probability of getting the required facts from the World Wide Web is extremely high due to the strength of existing search engines like Google and Yahoo. Consequently, these systems can actually find the answer without knowing what it is or in which domain it belongs to. Secondly, the answers produced by these systems require no justification. As answers are extracted directly from the web pages using question keywords, these systems cannot provide traceability for the origin of the answer, and justifications or explanations on why the answer is as such. Despite these drawbacks, the approach of directly extracting answers from the web pages without alteration works well for the above-mentioned questions because these types of questions ask for facts, not comments. Lastly, an endless source of open-domain information is just a click away via a wide range of information retrieval tools for the World Wide Web readily available like Google, whose credibility is well-known. Thus, for every accurate answer, half of the credit should go to the web search engines.

Despite the practicality of being able to handle open domain questions, the modern-day question answering approach has resulted to great restrictions on the nature of question and response whereby the users are restricted to ask only factual questions, and the responses produced are merely extracted snippets and the validity cannot be verified. Voorhees (2003) even reaffirmed that the problems of modern-day question answering systems are mainly due to the limitations of their techniques based on shallow natural language processing and information retrieval.

### 1.1.4 The Key to Resolving the Question and Response Restriction

With the prevalence of shallow natural language processing and information retrieval in question answering, many would consider that the lure of achieving natural language understanding in question answering have come to an end. Nevertheless, the need to handle natural language, which many considered as the sole inherent problem in question answering, has put modern-day question answering in a state of restlessness. Various problems due to the richness of natural language continue to haunt question answering systems which are based on shallow natural language processing and information retrieval.

Strzalkowski *et al.* (1999) examined the approaches of modern-day question answering which employs shallow natural language processing and information retrieval, and noted that they were far from adequate. When problems that involve question answering systems arise, researchers attempt to look for isolated solution from natural language understanding to solve the problem. This does not mean such systems actually attempt natural language understanding. To actually achieve understanding of natural language, more is required than just the adoption of several natural language processing tasks. However, when the question *"what if I ask for questions that require more than just facts?"* arises, the community from recent Question Answering Roadmap (Maybury, 2003) instantly knows that such requirement is far from reach using the current approach of modern-day question answering.

The ability to handle non-factual questions and generate intelligent responses requires a totally different approach if the system desires a stable, reliable functioning. The overall participants from the Question Answering Roadmap workshop felt that the lacks of adoption of general natural language processing and reasoning were limiters to progress (Maybury, 2003). This has led to a new line of approach in question answering through the heavy use of knowledge base with understanding and reasoning (Moreale & Vargas-Vera, 2003). Through the Question Answering Roadmap, the scope of question answering has been expanded along several dimensions including multiple question types, multiple answer types, multiple media, multiple languages, interactive dialog with the user to refine/guide the question answering process, multiple answer perspectives, and ultimately, answers which provide an evaluation or judgment based on retrieved data. This shift is also demonstrated through the proposed future directions of many researchers. Kwok *et al.* (2001), for example, has acknowledged through its future work section that their implementation of system lacks the use of syntactic and semantic information. By making more use of this information, they can actually improve their recall. In the future work section of Breck *et al.* (1999), they have considered including the ability to coreference and also to enrich the semantic representations extracted from questions and documents. Another important future direction of theirs is to upgrade the nature of the answers to move beyond simple answer retrieval into full-blown answer synthesis. Last but not least, from the writings of Nyberg & Mitamura (2002, p.1), *"QA systems are expanding beyond information retrieval and information extraction, to become full-fledged, complex NLP applications…"*. In a way, these researchers do agree that a higher level of natural language understanding and reasoning is necessary to improve the quality of a question answering system.

Hence, enlightened by the fact that any major leap in question answering simply will not be achieved without attempting a deeper understanding of at least parts of natural language texts, some researchers have decided to get to the root of the problem. This has led to a new line of approach in question answering through the heavy use of knowledge base with reasoning while maintaining the openness of the domain and information source (Moreale & Vargas-Vera, 2003; Maybury, 2003). Thus, the idea of understanding the natural language information presented on the World Wide Web and the ability to reason for intelligent answers has been the focus of this resurrection, giving rise to the approach towards question answering based on natural language understanding and reasoning. Although different in many ways, but START (Katz, 1997) and WEBCOOP by Benamara (2004) are two good examples from this new approach.

### 1.1.5 The Lack of a Comprehensive Solution

Therefore, in order to improve upon the existing question answering facilities, we will need to approach the problem using natural language understanding and reasoning. A review on the existing systems of similar nature (i.e. START and WEBCOOP) which is discussed in Chapter 3 looks for comparisons and rooms for improvements, reveals that there is no single ideal architecture that comprehensively integrates both natural language understanding and advanced reasoning that are necessary for adequately solving the restrictions on the nature of question and response without instigating new problems. Hence, on top of the contributions by these systems, such lack also shows that there are still rooms for improvements.

Many approaches by various current researches towards natural language understanding have only reached some level of the semantic analysis and not to the level of discourse analysis. Moreover, these systems store the output of natural language understanding using inexpressive representation formalisms that cannot fully exploit intrinsic properties like inheritance, generalization, etc and also, ontological information. Besides, due to the minimalist approach in the representation formalism where there is no need to handle ontological information, and to capture intrinsic properties, the reasoning approach is limited to merely rule-based and no advanced reasoning features are possible. Apparently, such approaches may be beneficial in terms of the processing time but the ease of scalability across domains and to real-life natural language text is questionable.

Hence, the solution to the problems in this research is not as simple as including natural language understanding and reasoning, but the consideration for the various levels of natural language understanding, choice of representation formalism and the reasoning technique with advanced features is also important. A thorough consideration made during the design of the solution is important to make sure that during the course of solving one problem, additional unforeseen problems are not introduced.

### 1.1.6 Existing Measures for Evaluation of Question Answering

During the review of the evolution of question answering in Section 1.1.2, we have shown that one of the contributing factors to the prevalence of the approach based on shallow natural language processing and information retrieval is due to practical consequences that may result from the introduction of higher-level of natural language understanding into question answering. Even for systems that actually attempt natural language understanding and reasoning like START and WEBCOOP, the aim for practicality and simplicity in responses or other aspects of the system has guided them to avoid higher-level of natural language understanding and reasoning. This is the reason behind the variations in practices by the various question answering systems based on natural language understanding and reasoning. WEBCOOP sacrifices a comprehensive analysis of web pages by focusing only on important information to ignore the complexity of natural language understanding. START focuses the parsing to only single sentences to avoid the fuss in full semantic and discourse analysis. START opts for simple matching of logical expressions instead of advanced reasoning. Full natural language understanding and advanced reasoning are usually not practical and laborious to scale across multiple domains.

Conclusively, for the validity of the solution of this research, an evaluation must be performed to assess the response time and response quality of the prototype to ensure that the implementation of comprehensive natural language understanding and reasoning into question answering does not compromise practicality factors.

In order to promote a fair environment for evaluation and comparison between systems, several existing measures were considered. While not all evaluation measures are applicable to domain-oriented question answering systems, they are still discussed to demonstrate the scarcity of standard measures. The most notable evaluating for question answering has to be the question answering track in the TREC evaluation (Voorhees, 2003). Evaluation in TREC is essentially based on the F-measure to assess the quality of response in terms of precision and recall. Such mode of evaluation is tailored for all question answering systems based on shallow natural language processing and information retrieval like AnswerBus where information retrieval is the backbone of such systems. To enable F-measure, a large query and document ensemble is required where the document collection is manually read and tagged as correct or incorrect for one question out of a list of predefined. There are several inherent requirements with F-measure that makes it inappropriate for evaluations of domain-oriented question answering systems based on understanding and reasoning:

- assessments should average over large corpus or query collection;
- assessments have to be binary where answers can only be classified as correct and incorrect; and

- assessments would be heavily skewed by corpus, making the results not translatable from one domain to another.

The first requirement actually makes it extremely difficult to evaluate domain-oriented systems like START and NaLURI due to the absence of large quantity of domain-related documents collection. Besides, like most other systems based on understanding and reasoning, NaLURI uses knowledge base as information source instead of a large document collection, making F-measure impossible. For modern-day question answering systems, the large corpus requirement has been handled by TREC. Secondly, responses produced in question answering system based on understanding and reasoning such as START and NaLURI are descriptive in nature and thus, cannot be merely classified into correct and incorrect. Moreover, the classification is manually done by human experts, making the results extremely subjective and non-definite. Lastly, most systems based on understanding and reasoning actually has domain portability as their main aim by starting out as a domain-restricted system and slowly grows or moves to other domains. The characteristic of F-measure that skews according to domains makes it inappropriate for evaluation of such systems.

There are also other measures but are mostly designed for general tasks related to natural language processing like translation, database query, etc. Facemire (1989) proposes that a simple number scale be established for the evaluation of natural language text processing systems. This metric is to be based on human linguistic performance, taken as 1.0, and is the simple average of four subcomponents which are the size of the lexicon, the speed and accuracy of the parse and the overall experience of the system. The author has also oversimplified matters by equating the ability of understanding to mere sentence parsing. Also, the use of the criteria of speed and accuracy in parsing has limited the ability of the metric to move on with time. As the computing strength increases in terms of hardware and software, the factor of speed and accuracy can no longer be discriminative enough to separate one system from another.

Unlike the previous model, a general model is provided by Guida & Mauri (1984) that acts as a basis of a quantitative measure for evaluating how well a system can understand natural language. But how well a system can understand natural language only provides for half of the actual ability required to generate high-quality responses. Hence, such a general model is inadequate for more specific application of natural language understanding like question answering.

Srivastava & Rajaraman (1995) have also attempted to devise an experimental validation for intelligence parameters of a system. The authors concluded that intelligence of a question answering system is not a scalar value but rather, a vector quantity. The set of parameters that define intelligence are knowledge content of a system, efficiency of a system and correctness of a system. In this approach, the respondent is an entity that has the answer in mind and the questioner must attempt to guess what is in the mind of the respondent with the help of the least number of questions. The questioner who manages to figure out the answer using the minimal number of questions is considered intelligent. Hence, this approach is not suitable for evaluating the quality of responses in a standard setting of question answering system.

Allen (1995) and Nyberg & Mitamura (2002) have also suggested a type of black-box evaluation where a system is evaluated to observe its efficiency at producing quality or desirable answers. Diekema *et al.* (2004) further characterized the black-box evaluation and suggested that systems can be evaluated on their answer providing ability that includes measures for answer completeness, accuracy and relevancy. The authors also state that evaluation measures should include more fine grained scoring procedures to cater for answers to different types of questions. The authors give examples of answers that are explanations, summaries, biographies or comparative evaluations that cannot be meaningfully rated as simply right or wrong. We consider this black-box approach as comprehensive in assessing how well question answering systems produce responses required by users and how capable these systems are in handling various types of situations and questions. Despite the merits of the evaluation approach, none of the authors provide further details on the formal measures used for scoring and ranking the systems under evaluation.

## 1.2 Problem Statement

Question answering systems tend to be rather complex, and can be approached differently depending on the dimensions as has been observed in the first section. In a Master's research, it is very difficult to thoroughly investigate all aspects of a question answering system. Therefore, certain boundaries or scope have to be set with respect to the dimensions related to the problem and solution. For this research, we have limited ourselves to only look at the problem from the dimensions of question and response, and solution from the dimension of technique.

By summarizing Section 1.1.3, the problem of this research lies in the dimensions of question and response. Formally, the initial problem is:

- obsession on the use of low-level syntax analysis and web search engines to harvest open-domain online information has resulted in restrictions on the nature of question and response.

We have then shown in Section 1.1.4 that the solution to the restrictions on the nature of question and response lies in the dimension of technique, which is based on natural language understanding and reasoning. Following that, existing solutions are reviewed and by summarizing Section 1.1.5, there is no single comprehensive solution that is fit to solve the initial problem. From the approaches by existing systems based on natural language understanding and reasoning, a resultant problem is present:

- low-level natural language understanding, minimalist choice of representation formalism and conventional reasoning without advanced features will pose difficulty in scaling across domains and to real-life natural language text.

## 1.3 Hypothesis

Based on the problems and the scope defined, the solution is only concerned with improving question answering from the aspects of diversity of questions and quality of responses without introducing scalability problem. By considering the resultant problem, it is obvious that the solution to the initial problem is not as simple as introducing natural language understanding and reasoning into question answering. The design of the solution must put into consideration three aspects namely full-discourse natural language understanding, powerful and expressive representation formalism like semantic network, and network-based reasoning that supports advanced reasoning.

Firstly, the solution to the problems by this research must adopt a natural language understanding approach that not only covers the necessary aspects of semantic analysis, but also to include the crucial aspects in discourse analysis. As a result, such approach will ensure that the question answering system can handle both questions and information from natural language text from any information source and domain.

Secondly, with the solution based on powerful and expressive representation formalism like the semantic network, facts produced by the full-discourse natural language understanding, including intrinsic properties between entities can be captured and expressed with the help of ontological information. Cyberlaw was chosen out of the many domains for use in the design of ontology and gazetteer. It is vital to point out at this point that the main focus of this research is the dimensions of question, response and technique, and not the dimension of domain (i.e. Cyberlaw). Cyberlaw was selected for use in this research purely for its novelty and the abundance of news available online.

Thirdly, with the network-based representation formalism, alternative approaches for reasoning, other than rule-based that can fully exploit the formalism's expressiveness can be adopted in the solution. Also, with the ontological commitment supported by the network-based representation formalism, the integration of advanced reasoning features can be done.

In a nutshell, the hypothesis of this research states that the problems described in Section 1.2 will be solved through the innovative introduction of natural language understanding and advanced reasoning into a question answering system in the form of the following three criteria:

- full-discourse natural language understanding approach is required to ensure the ease of scalability across domains and to real-life natural language text;
- representation formalism that has good expressive capability for intrinsic properties with support of ontological information, like the semantic network is required for representing rich output from natural language understanding; and
- reasoning approach capable of fully exploiting what the representation formalism has to offer and to work with the advanced reasoning features is required.

## 1.4 Research Methodology

After formulating the hypothesis, the main objective of this research is none other than to prove the validity of the hypothesis. To achieve this objective, this research is conducted in five distinct phases. Figure 1.2 depicts how the research is conducted by illustrating the problems understudy, how these problems surfaced through a series of reviews, the resulting solution design and implementation to overcome the problems, and finally, evaluation.

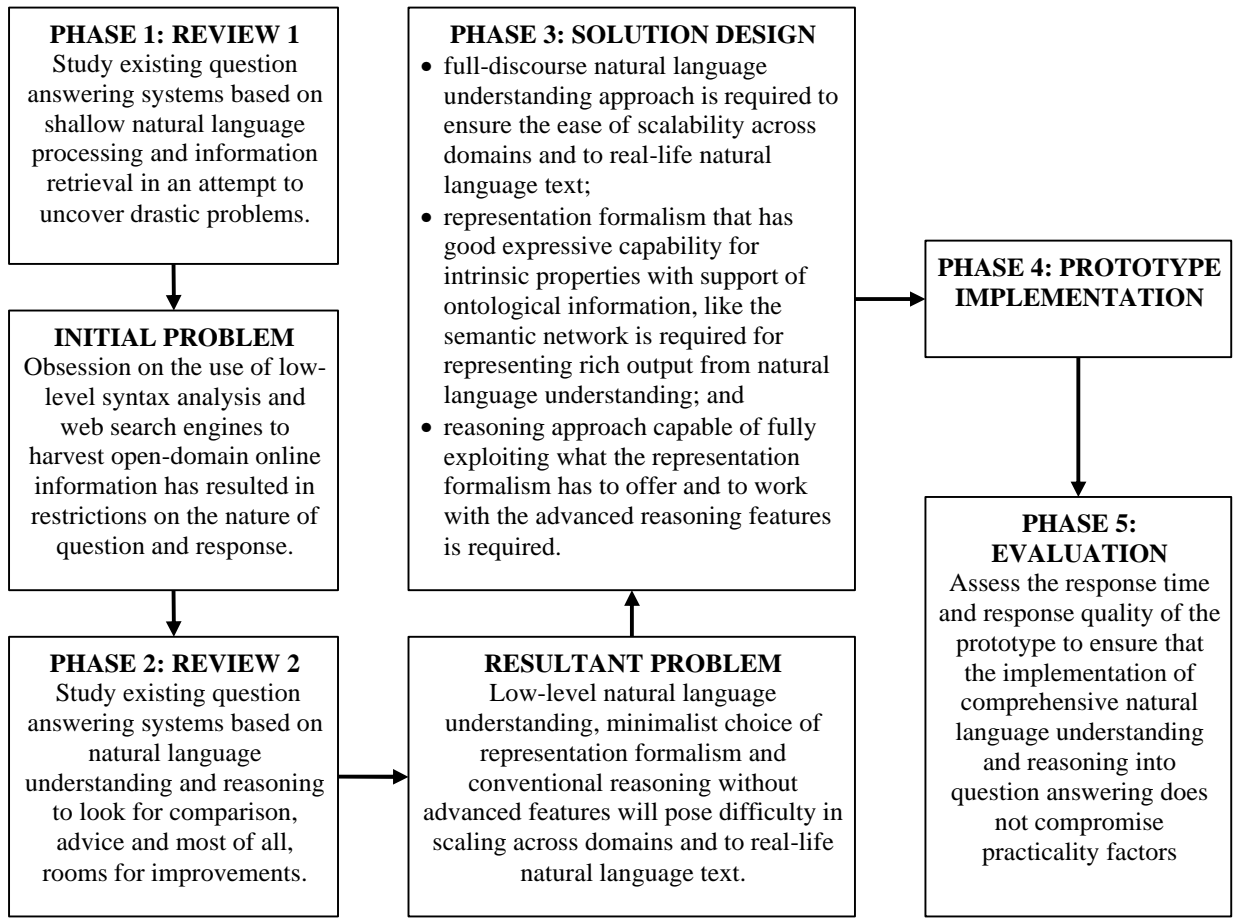

Figure 1.2: Research methodology

Phase 1 and 2 are performed to analyze and disclose the existing practices in question answering systems. These reviews will be performed on four prominent online systems (i.e. two from each approach). These systems are scrutinized to a very detailed level, and the merits and shortcomings are made visible. The shortcomings constitute the problem statement in this research whereas the merits are innovatively adopted for use in the design of the solution framework. Phase 1 and 2 are documented in Chapter 3.

From the problems excavated during the activities in Phase 1 and 2, a hypothesis is formulated. Phase 3, 4 and 5 is all about realizing and proving the hypothesis. Phase 3 and 4 are carried out to design and develop a prototype question answering system that will be used in Phase 5 for evaluation. Phase 3 is carried out in three parts namely the design of knowledge base and gazetteer, natural language understanding mechanisms, and reasoning mechanisms. The process of realizing Phase 3 is documented in Chapter 4, whereas the development of the prototype in Phase 4 based on the design from Phase 3 is discussed in Chapter 5. Finally, Phase 5 is administered to evaluate the system to ensure that response time and response quality is not affected, and the process and results are then explicated in Chapter 6.

The first two chapters namely Chapter 1 and Chapter 2 are prepared to bring the readers into the field of question answering and the related fields (i.e. natural language understanding and knowledge representation and reasoning), the proliferation of problems and the potential solutions. Certain notable approaches in the fields of natural language understanding, and knowledge representation and reasoning discussed in Chapter 2 are also considered for use during the design of the solution framework. Also, the last chapter, Chapter 7, is prepared to summarize the entire dissertation and highlights a list of contributions from this research. Besides, limitations and the possible solutions, and the future directions for this approach of question answering are also included.

# Chapter 2

# Background Information

## 2.1 Introduction

As highlighted in Chapter 1, the design of the solution framework must put into consideration three aspects namely full-discourse natural language understanding, powerful and expressive representation formalism like semantic network, and network-based reasoning that supports advanced reasoning. Hence, to promote a common understanding regarding the theoretical background of these three aspects and also to act as reference materials at any point in time, relevant introductory information is organized and put forward.

## 2.2 Natural Language Understanding

The answer to our intelligence is our ability to make use of and understand languages. Linguistic competency is seen as the dividing line that separates us from other primates, setting the foundation for civilizations and cultures. Thus, providing machines with the ability to understand natural languages has always been the dream of the scientific community, even to the extend of equating the ability of natural language understanding as the key to artificial intelligence.

The initial researches on transforming natural language segments into computer-understandable representation were focused on solving in-depth problems, mainly on the lower linguistic levels like morphemes and syntax rather than tackling the problem of full natural language understanding. This was mainly due to the underestimation of the level of complexity of natural language understanding by early researchers, causing them to break free from the big picture.

The existing strategies, algorithms and current research on natural language understanding, or better known as natural language processing, can be classified according to a model used in linguistics that divides language into six separate levels namely phonetic, morphology, syntax, semantic, pragmatic and discourse (Allen, 1995). Natural language processing deals with the various computational treatments of meanings conveyed through natural language and construction of computational models for natural language. Each of these computational models are crafted to deal with problems at different levels of linguistic and only systems that aims to achieve natural language understanding will employ complete natural language processing tasks. Certain applications for information extraction, information retrieval, question answering, machine translation, text summarization and speech understanding will employ techniques from natural language processing to solve isolated problems.

The process of understanding text-based natural language involves different stages of analysis as shown in Figure 2.1 and the success of their implementation constitutes full natural language understanding.

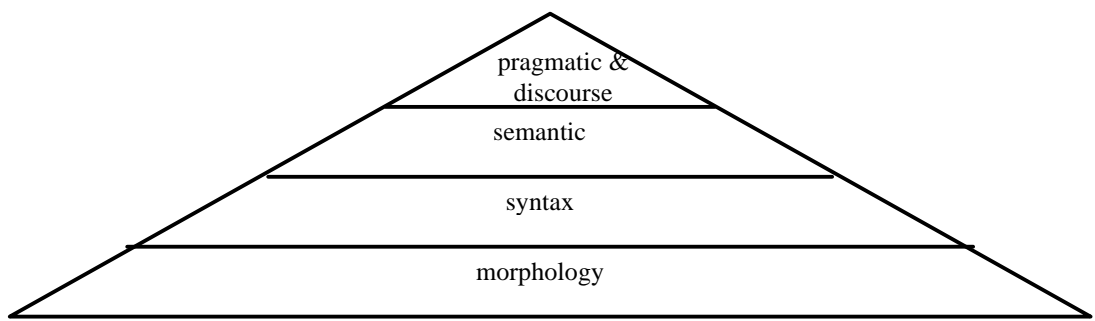

Figure 2.1: Stages of analysis in natural language understanding

Most problems in natural language understanding especially those in the morphology and syntax level have been solved in a sufficient degree as to proceed to the next higher levels. The ability to handle natural language on those higher levels constitutes most of what is known as understanding of natural language. While certain progresses in semantic and discourse analysis exist, it is still at infancy stage. Only through advancement in these higher levels can the notion of understanding achieve greater realization. Much research on the hard problem of in-depth understanding by computer was performed circa 1970s, interest shifted in the 1990s to information extraction and word sense disambiguation (Mueller, 1999). Now that a degree of success has been achieved on these easier problems, it is time to return to in-depth understanding.

**2.2.1 Morphological Analysis**

Morphological analysis is the study of elements within words, where individual words are analyzed into their components and non-word tokens such as punctuation are separated from the words. It involves identification of word stem, identification of prefixes and suffixes, thereby possibly identifying the tense and number and classification into noun, verb and so on. Part-of-speech tagging is probably the most common form of morphological classification of words although not much morphological information is produced.

Part-of-speech tagging is the task of assigning an appropriate part-of-speech to each word in a sentence. For example, noun (e.g. court, company), verb (e.g. sue), pronoun (e.g. he, they), articles (e.g. the, a), demonstratives (e.g. that, those this), adjectives (i.e. describe nouns), adverbs (i.e. describe verbs), prepositions (e.g. in, from, to, out), coordinating conjunctions (e.g. and, or, but), subordinating conjunctions (e.g. if, because, that) and so on. Part-of-speech tagging is a process of morphological analysis and due to its location on the lowest level of linguistics, it is usually taken as the first step in natural language processing language at the sentence level. Its output is vital for tasks such as sentence parsing and word sense disambiguation on the higher level. The two major approaches to this task are rule-based approaches (Brill, 1992) and statistical approaches (Cutting *et al.*, 1992).

## 2.2.2 Syntactic Analysis

Syntactic analysis is the study of elements within sentences. It involves transforming linear sequences of words into structures that exhibits how the words relate to one another. The process of converting the flat list of words of the sentence into a structure that defines the units represented by that list is known as sentence parsing. Sentence parsing is seen as the second phase of natural language understanding, which utilizes the output from part-of-speech tagging.

Computing the constituents of a sentence requires two things namely the grammatical model, which is a formal specification of the permissible structures in the language and the parsing technique, which is the method of analyzing a sentence to determine its structure according to the grammar. The rules of parsing and its constraints are declaratively defined in what is known as grammar. The constraints are word order, number agreement and case agreement. For example, *"client the key"* is an illegal constituent in the sentence *"He gave the client the key"*. The most common way to represent grammar is as a set of production rules.

Sentences have structures and are made up of constituents known as phrases. A phrase consists of a head and modifiers. Based on the example Figure 2.2 below, the head of the noun phrase is *"ruling"* and the modifiers are *"the long-awaited"* and *"by the federal court"*.

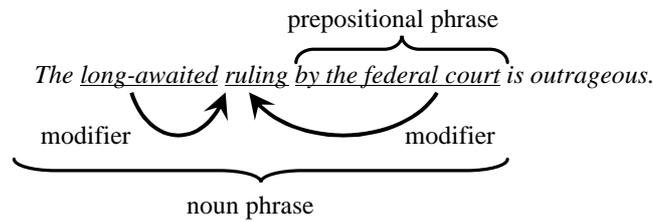

Figure 2.2: Grammatical structure for the sentence *"The long-awaited ruling by the federal court is outrageous"*

The predominant properties that characterize each particular variety of phrase and which establish the role it plays in a sentence are determined by the properties of the principal or head word that it includes. The head word of the subject or object of a sentence, for example, is a noun so that these constituents are categorized as noun phrases. These categories are known as grammatical categories and the relationship between them are known as grammatical relationships.

Two of the widely studied and implemented parsing grammars are link grammar (Sleator & Temperley, 1993) and principle-based grammar (Lin, 1993). While Link Grammar Parser[1] is based on link grammar, Principar (Lin, 1994) and its descendant Minipar[2] are principle-based grammar parsers. Minipar adopted some of the ideas of the Minimalist program (Chomsky, 1995) such as bare phrase structure and economy principles.

### 2.2.3 Semantic Analysis

Semantic analysis involves assigning meanings to the structures created by syntactic analyzers. One of the well-known examples, *"Colorless green ideas sleep furiously"* by Chomsky (1985) is syntactically correct but would certainly be rejected as it is semantically unusual. Semantic analysis requires the mapping of individual words into appropriate objects in the knowledge base and the creating of correct structures that best represent the meanings when individual words are combined together. The tasks in this step can be grouped into two research directions namely word sense disambiguation and named-entity tagging. Much success at this natural language understanding level comes from other fields namely knowledge representation and information extraction that strive for the same goal, that is to have a meaning representation of natural language sentences. Thus, tasks like word sense disambiguation and named-entity tagging can be seen appearing in these fields.

---

[1] The executable version is downloadable via http://www.link.cs.cmu.edu/link/ftp.html
[2] The executable version is downloadable via http://www.cs.ualberta.ca/~lindek/minipar.htm

Named entities are names that denote unique entities. Named entity is also known as proper nouns or proper names in linguistics. The meaning of a proper noun, other than what it refers, is generally arbitrary or irrelevant. For example, someone might be named *"Tiger Woods"* despite being neither a tiger nor woods. Proper nouns are capitalized in English and most or all other languages that use the Latin alphabet. Sometimes, the same word can appear as both a common noun and a proper noun, where one such entity is special.

Many literatures implement named-entity recognition as part of an information extraction and see it as an indispensable component (Maynard *et al.*, 2001). Named-entity recognition is also an essential prerequisite for many other tasks based on natural language understanding. Named-entity recognition involves identification of proper names in texts and classifying them into a set of predefined categories of interest. Named-entity recognition constitutes the first step in extracting structured information that is meaningful and thus, many see it as a task in semantic analysis. Three universally accepted categories are person, location and organisation. Usually, the other categories will either be a specialization of the three categories or entirely specific to a particular domain. Many may consider named-entity recognition as a trivial task of matching words against a predefined list of entities but in reality, it is totally a different case. One of the problems is the existence of variations in named entity. For example, a person's name

*"Alexandre Petrovsky"* can also be referred to as *"Mr. Petrovsky"*, *"Alexandre"* or simply *"Alex"*. Ambiguity that exists in named entity types has also created a considerable amount of problems. To illustrate, *"Washington"* can be a person's name and also a location. *"Price Waterhouse"* can be the all-famous accounting firm and although less likely, a person's name. Ambiguity that exists with common words like *"may"* can also be problematic.

Much of the initial work on finding names was based on either the use of list lookup, triggering approach or carefully hand-crafted pattern. MUSE by Maynard (2003) offers an example of a system that employs the list lookup approach. Such system recognises only entities stored in its lists or better known as gazetteers. It is simple, fast, language independent, easy to retarget. The disadvantages are the costs involved in collection and maintenance of the gazetteers and problems in handling name variants. As for the triggering approach (Gaizauskas *et al.*, 1995), internal evidence or structure that often exists in names is utilized. These components can be either stored or guessed. For example, the name *"Penang Hill"* can be identified through the rule *"Capitalized Word + {Hill, Mountain}"*. In the pattern matching approach, patterns are constructed by hands and used to match phrases. This approach is adopted in FASTUS (Appelt *et al.*, 1993). To illustrate, this phrase:

*"Guerrillas attacked Merino's home in San Salvador 5 days ago with explosives"*

will match the pattern:

*"<Perp> attacked <HumanTarget>'s <PhysicalTarget> in <Location> <Date> with <Device>"*

Another task in semantic analysis is word sense disambiguation, which involves the association of a given word in a text or discourse with a definition or meaning which is distinguishable from other meanings potentially attributable to that word. The task therefore necessarily involves two phases namely the determination of all the different senses for every word relevant to the text or discourse under consideration and the assignment of each occurrence of a word to the appropriate sense.

To accomplish the first phase, many of the recent work on word sense disambiguation rely on predefined senses, which includes a list of senses such as those found in everyday dictionaries and a group of features, categories or associated words. However, with the improving reliability of part-of-speech taggers, the work of word sense disambiguation has since focused largely on distinguishing senses among homographs belonging to the same grammatical category rather than all words. That is, for homographs with different parts of speech (e.g. *"court"*), part-of-speech tagging accomplishes word sense disambiguation. Hence, the task of word sense disambiguation is considered a level above syntax analysis, which is in the semantic analysis phase that deals with semantic and the contextual information in which the words exist.

As for the second phase, the assignment of words to senses, is accomplished by reliance on two major sources of information namely context of the word to be disambiguated and external knowledge sources like lexical and encyclopedic resources as well as hand-devised knowledge sources. The various methods for association of words with senses are used to determine the best match between the current context and one of these sources of information. These methods can be broadly classified into artificial intelligence methods, knowledge-based methods and empirical methods (Ide & Véronis, 1998).

### 2.2.4 Pragmatics and Discourse Analysis

Pragmatics and discourse are not as closely defined as the other linguistic areas. It can be said that both are concerned with the use and understanding of language beyond the boundaries of a single sentence. Discourse analysis is concerned with how the meaning of individual sentences may depend on the sentences that precede them and may influence the sentences yet to come. The entities involved in the sentence must either have been introduced explicitly or they must be related to entities that were. Thus, the overall discourse must be coherent. The information about the dependency among sentences is especially important for resolving pronouns, making anaphora resolution one of the important tasks in discourse analysis. Pragmatics analysis, on the other hand, is concerned with how sentences are used in different situations and how the uses affect the interpretation of the sentence. Very little formalisms for representing pragmatics or discourse have been developed because research in the areas of pragmatics and discourse is still at a basic level (Hu & Atwell, 2003; Fischer, 2003).

In this example, *"The defendant has not arrived yet but he should be here any minute"*, the reference *"he"* is the anaphor and *"The defendant"* is the antecedent. The process of determining the antecedent of a referencing anaphor is known as anaphora resolution. Note that the reference is only called anaphor if it comes after the antecedent. In the case where the reference comes before the antecedent, the reference is known as cataphor. For example, *"Because he was still underage, Matthew was trialed in a juvenile court"*. Anaphora resolution must not be confused with coreference resolution where in the latter, the aim is to resolve the anaphor and its antecedent used as referring expressions to the same referent in the real world. The relation between the anaphor and the antecedent is not the same as the one between the anaphor and its referent. In the example above, the referent is "*The defendant"* as a person in the real word whereas the antecedent is "*The defendant"* as a linguistic form.

There are three most widespread types of anaphora namely pronominal anaphora, definite noun phrase anaphora and one-anaphora (Hobbs, 1977). Pronominal anaphora is the most common type of anaphora realized by anaphoric pronouns such as *"she"*, *"they"* and so on. It should be pointed out that not all pronouns in English are anaphoric. For instance, *"it"* can often be non-anaphoric. Consider the expressions such as *"It is important"* and *"It is necessary"*. The next type of anaphora is definite noun phrase anaphora.

A typical case of this type is when the antecedent is referred by a definite noun phrase representing either the same concept or semantically close concepts such as synonyms and hypernyms. Consider the example, *"Police were everywhere during the trial. The law enforcers were there to make sure that everything was in order."* The last type is one-anaphora. It is realized by the use of *"one"* noun phrase. For example, *"If you could not make it for this sitting, you can try the evening one"*. Anaphors can also be differentiated based on the sentence in which the anaphor and antecedent are located. Intrasentential anaphors refer to an antecedent which is in the same sentence as the anaphor and intersentential anaphors refer to an antecedent which is in a different sentence from that of the anaphor.

Most of the anaphora resolution systems deal with resolution of anaphors which have noun phrases as their antecedents because, identifying anaphors which have verb phrases, clauses, sentences or even paragraphs/discourse segments as antecedents, is a more complicated task. Typically, all noun phrases preceding an anaphor are initially regarded as potential candidates for antecedents. Usually, a search scope has to be identified and most approaches look for noun phrases in the current and preceding sentence. However, an ideal anaphora resolution system should extend its scope of search. Antecedents which are seventeen sentences away from the anaphor have already been reported (Mitkov, 1998). Assuming that the scope of search for a specific approach has already been specified, the noun phrases preceding the anaphor within that scope are identified as candidates for antecedents and a number of anaphora resolution factors are employed to track down the correct antecedent.

Approaches to anaphora resolution usually rely on syntactic information like gender and number agreement, semantic consistency and parallelism and some even uses commonsense knowledge. Because of the need for syntactic and semantic information and the involvement of language understanding beyond the boundary of single sentence, anaphora resolution is considered as a task in discourse analysis. Algorithms for anaphora resolution can be broadly categorized into syntax-based approaches (Liakata, 2000), semantic-based approaches (Wilks, 1985), statistical approaches (Itai & Dagan, 1990), artificial intelligence approaches (Mitkov, 2001) and finally hybrid systems.

## 2.3 Knowledge Representation and Reasoning

The following is a typical conversation with the Eliza chat robot (Weizenbaum, 1966).

> *User: Men are all alike.*
>
> *Patient: In what way?*
>
> *User: They're always bugging us about something or other.*
>
> *Patient: Can you think of a specific example?*
>
> *User: Well, my boyfriend made me come here.*
>
> *Patient: Your boyfriend made you come here?*
>
> *User: He says I'm depressed much of the time.*
>
> *Patient: I am sorry to hear you are depressed*
>
> *User: It's true. I am unhappy*

At first sight, Eliza and other systems of the same nature (Mauldin, 1994) appeared extremely sophisticated for being able to carry out such human-like conversation by emulating a therapist. A closer look at how the program works shows just how simplistic it really is. ELIZA operates through sequences of pattern matching and string replacement. For example the pattern *"XXX do you YYY me"* matches the sentence *"Why do you hate me?"*. On encountering this sentence the program selects a transformation from the list of possible transformations for this pattern, for example, *"You like to think I YYY you, don't you?"* and produces the output *"You like to think I hate you, don't you?"*. Clearly this is not a robust system and there are many examples of ELIZA producing complete gibberish due to stringently applying these transformation rules. It should be obvious from the example of ELIZA that the only way to provide the ability for computers to reason and answer questions based on its understanding is by giving them ways of representing knowledge of the world about which they converse. Since it is not possible to represent everything in the real world, a representation must necessarily focus on some things while ignoring others. This makes representations imperfect approximation to reality and thus, by selecting any representation, a set of decisions are actually made about which part of the world to see and how to reason with what is seen. That is, selecting a representation means making a set of ontological commitments and defining a set of sanctioned inferences. As an approximation of the world is made, any knowledge generated or manipulated in the domain of the system is collectively referred to as domain knowledge.

### 2.3.1 Components of Knowledge Representation

The graphical representation of the two components of knowledge representation is illustrated in Figure 2.3. Formally, knowledge representation formalism allows the definition of two components of knowledge namely definitional and derivational (Davis *et al.*, 1993).

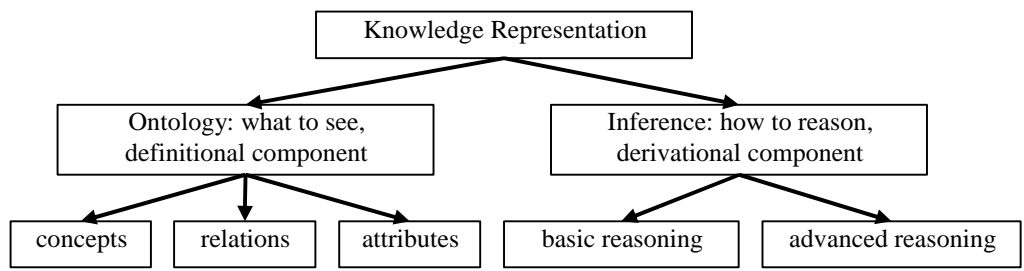

Figure 2.3: Components of a knowledge representation formalism

While some researchers like Clark (1989) may refer to the components as semantic role and computational role, they are essentially the same. The definitional component (i.e. semantic role) is used for representing knowledge that are explicit namely concepts, instances of concepts, relations between concepts and attributes of concepts in a hierarchical manner known as ontology while the derivational component (i.e. computational role) allows a common understanding of the structure of knowledge through the use of knowledge base. The derivational component defines how the knowledge should be interpreted through a set of sanctioned inferences or in short, how to reason. The capacity of reasoning provides the dynamism required to make the knowledge base useful. Without this ability, the knowledge base would be just another plain, unstructured memory.

### 2.3.2 Knowledge Representation Formalisms

Many people have misconceptions concerning knowledge representation formalism and conventional data structures. It is necessary to note that part of what makes a language representational is that it carries meaning, in summary, there is a correspondence between its constructs and things in the real world. A semantic network, for example, is a representation while a graph is a data structure. While every representation must be implemented in the machine by some data structure, the representational property is in the correspondence to something in the real world and in the constraint that correspondence imposes. Some of the major formalisms or languages for knowledge representation are first-order logic, semantic networks, frames and other derivative efforts such as XI (Gaizauskas & Humphreys, 1996) and KL-ONE (Burkert & Forster, 1991).

First-order logic is commonly used because of its mathematical basis to avoid excessive complexity. First-order logic has a long history which dates back to as far as Aristotle. It involves the use of standard forms of logical symbolism which have been familiar to philosophers and mathematicians for many decades. Most simple sentences, for example, "*Company A is big*" can be represented in terms of logical formula in which a predicate is applied to one or more arguments. In the example, "*big*" will be the predicate and "*Company A*" is the argument resulting in *big(Company A)*. Due to the well-understood analytical techniques and ease in expressing the formulae of first-order logic in artificial intelligence languages such as LISP, it has been a very popular knowledge representation symbolism within artificial intelligence.

Another mechanism for representing knowledge is the semantic network. A semantic network can be described with the following scenario in which *"Company A sues Spammer B"*. The network will consist of a series of concepts and relations. The network will include the knowledge *"Company A is_a company"*, *"Spammer B is_a man"*, *"Company A initiated the lawsuit"* and much more. There can also be more general facts such as *"Company A a_kind_of organization"*, *"Spammer B a_kind_of person"*, *"Lawsuit a_kind_of legal action"* and so on. Each of these facts can be expressed by means of an arc joining two nodes in a network. The relationships *"Company A is_a company"*, *"Spammer B is_a man"* forms the *"is_a"* hierarchy and *"Company A a_kind_of organization"*, *"Spammer B a_kind_of person"*, *"Lawsuit a_kind_of legal action"* forms the *"a_kind_of"* hierarchy. Any properties that a node in such a hierarchy possesses can be inherited by other objects lower down in the hierarchy. So, if the semantic network contains the information that *"company has name"*, then it can be inferred that this property is also possessed by *"Company A"*. There is a close similarity between a semantic network and predicate logic since any semantic network can be expressed as a series of logical predicates. There are also important differences. All the information relating to a node in semantic network is grouped together but in predicate logic, the facts are just stored as one long series of predicates and so for a large knowledge base, finding related facts may require a long search (Sharples *et al.*, 1989).

As for frame representation, facts are clustered around concepts (Bratko, 2001). A frame is a data structure which consists of slots containing values. In slots, the values can be atomic or references to other frames. A slot that is left unfilled is filled through inheritance. The frame for a company for example, might contain a name slot, business nature slot and so on. Frames suffer from the frame problem of knowledge linking. A script is a type of frame that describes what happens temporally.

### 2.3.3 XI: A Hybrid Representation Language

XI is a language for representing knowledge about individuals, about classes of individuals and about inclusion relations between classes of individuals (Gaizauskas & Humphreys, 1996). At one level XI may be viewed simply as a declarative formalism with its own syntax and semantics based on first-order logic. While from another perspective, it may be viewed as yet another logic-based or frame-based inheritance system in the general tradition of KL-ONE or more generally as a form of semantic network (Gaizauskas & Humphreys, 1997). Classes are represented as unary predicates and individuals are atoms. Attributes are binary predicates, the first argument identifying the class or individual of which the attribute holds and the second being the value.

The basic alphabet of the definitional component of XI is called a term. A term can take the form of:
- a variable;
- an instance symbol (a 0-ary functor); and
- a complex term in the form $f(t_1,...,t_n)$ where $f$ is a functor and $t_i,...,t_n$ are terms.

In XI, complex terms are used to denote classes and attributes. Unary complex terms or class terms in the form $f(t_1)$ are used to denote classes. Binary complex terms or attribute terms in the form $f(t_1,t_2)$ are used to denote attributes. Lastly, if $t$ is a term, then *var(t)* denotes the set of variables appearing in $t$. With the use of connectives $\Rightarrow$ and $\leftarrow$, XI defined three types of expressions namely O-clauses for defining classes and instances, P-clauses for associating attributes with classes and G-clauses for defining goals or queries. G-clause will be explained as part of the derivational component of XI.

O-clauses can take one of the following forms:

- $c \Rightarrow D_1 \& ... \& D_n$ where c is a class term and each $D_j$ is a disjunction of class terms in the form $d_{j,1} \lor ... \lor d_{j,pj}$ and each class term $d_{j,k}$ contains variable occurring in c.

- $e \leftarrow c_1 \& ... \& c_n$ where e is an instance symbol and each $c_i$ is a class term such that $\text{var}(c_1) = ... = \text{var}(c_n)$

For example, using type-1 O-clause, the hierarchical relationship between class *"government"* and its subclasses can be defined as $government(X) \Rightarrow court(X)$. Instances of the class *court(X)* using type-2 O-clause as $g1 \leftarrow court(X)$ can also be defined.

P-clauses take the form $prop(c,V)$ where c is either an instance symbol or a class term and V is a set whose members are of the form:

- $p(t_1, t_2)$ where p is a functor and $t_1$ is c if c is an instance term and *var(c)* otherwise

- $p(t_1, t_2) :- G$ where G is a G-clause

To illustrate the P-clause, consider the attribute *name* for the instance *g1*:

$$prop(g1,[name(g1, mahkamah\_majistret\_shahalam)])$$

## 2.3.4 Basic Reasoning

Reasoning is the ability to manipulate the facts in hand to serve a purpose and in our case, the purpose will be providing answers to questions. Only certain facts can be reasoned and reasoning enforces the correct way of doing so through a set of inferences (Davis *et al.*, 1993). There are three types of basic reasoning namely deduction, induction and abduction (Sharples *et al.*, 1989). Basic reasoning typically comes intact with the knowledge representation formalism.

Deductive reasoning is the process of reaching a conclusion that is guaranteed to follow, if the evidence provided is true and the reasoning used to reach the conclusion is correct. The conclusion also must be based only on the evidence previously provided and cannot contain new information about the subject matter. Deductive reasoning in classical Aristotelian syllogism is typically of the form:

*If all Xs are Ys.*

*A is an X.*

*Therefore A is also a Y.*

Inductive reasoning is the process of reasoning in which a general rule is inferred from some set of specific observations. It provides an account of how generalization is performed but in a less reliable manner. It is to ascribe properties or relations to types based on limited observations of particular tokens or to formulate laws based on limited observations of recurring phenomenal patterns. For example, based on several observations that *"This swan is white"*, we would usually conclude that *"All swans are white"*, which in fact is not true. In logical terms,

*If P is true of a,*

*And P is true of b,*

*…*

*And P is true of n,*

*Then infer that P is true of all other entities of the same kind as a, b, c … n.*

One difference between both types of reasoning is that in deductive reasoning, the evidence provided must be a set about which everything is known before the conclusion can be drawn. Since it is difficult to know everything before drawing a conclusion, deductive reasoning has little use in the real world. This is where inductive reasoning steps in. Given a set of evidence, however incomplete the knowledge is, the conclusion is likely to follow, but one gives up the guarantee that the conclusion follows. However, it does provide the ability to learn new things that are not obvious from the evidence.

Abductive reasoning is the process of explaining by reasoning back from a situation to the state or action that produced it. Like induction, abductive reasoning is not a reliable form of inference. Despite the unreliable nature of the conclusion, many professionals employ this form of reasoning. It is in the form:

*If b (normally) follows from a*

*And b is known to be true,*

*Then infer a to be true.*

## 2.3.5 Reasoning using Semantic Networks and XI Language

The basic inference scheme in semantic networks is based on the mechanism of following links between nodes. While most inference schemes in systems are implicit, it sounds simple, there are two methods to perform the traversal namely path-based inference and node-based inference (Shapiro, 1978)**.** Path-based inference allows an arc or a path of arcs between two given nodes to be inferred from the existence of another specified path between the two same nodes. One the other hand, node-based inferences allows a structure of nodes to be inferred from the existence of an instance of a pattern of node structures. Path-based inference allows inheritance properties of hierarchies in semantic network to be clearly defined. Inheritance also provides a means of dealing with default reasoning in semantic networks. For example, we could represent *roosters are birds*, *typically birds fly and have wings* and *roosters run* in the semantic network depicted in Figure 2.4.

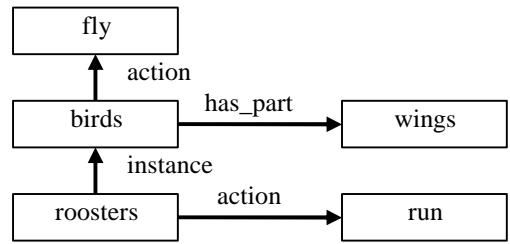

Figure 2.4: A semantic network for default reasoning

From Figure 2.4 also, it is obvious that we can say *roosters have wings* even though it is not explicitly stated because the conclusion was drawn through inheritance such that if *roosters are birds* and *birds have wings*, we can immediately deduce that *roosters have wings*. But we obviously cannot come to the conclusion that *roosters fly*. This is when we have to add exceptions such as *roosters do not fly*. Such exceptions will remove or falsified the previous derivation (i.e. *roosters do not fly*).

If a language is such that adding axioms does not affect previous derivations, then it is monotonic. However, in semantic network, inheritance with exceptions is necessary to make it widely applicable and hence, making semantic networks a non-monotonic logic. More generally, any sort of default reasoning is non-monotonic. A very good example of a classical monotonic system is the first-order logic.

Even though there are non-logical extensions in XI which makes it more powerful, but the reasoning mechanism in XI is still very much constrained by logic-based approaches. Like those of the classical semantic networks, the inference rules in XI are based on inheritance and other hierarchical relationships. There are four types of G-clauses with their own distinctive form:

- Clauses declaring the *a_kind_of* relations between a superclass and a subclass. Let $c_1$ and $c_2$ be class terms or variables: $c_1 \Rightarrow c_2$

- Clauses declaring the *is_a* relations between an instance and some class in the form. Let *e* is an instance symbol or a variable and *c* is a class term or a variable: $e \leftarrow c$

- Clauses declaring that an attribute holds for an instance. Let *e* is an instance symbol or a variable and *p* is an attribute term or a variable: $hasprop(e, p)$

- Conjunctions and disjunctions of the three basic types in the form of $G_1, G_2$ and $G_1; G_2$.

The inference rule to obtain the type-1 G-clause takes the form of:

$c \Rightarrow D_{1,1} \& ...D_{1,n1}$         (type-1 O-clause)

$c_1 \Rightarrow D_{2,1} \& ...D_{2,n2}$         (let $c_1$ be any class term that exists in some $D_{1,j}$)

$\vdots$

$c_{k-1} \Rightarrow D_{k,1} \& ...D_{k,nk}$         (let $c_{k-1}$ be any class term that exists in some $D_{k-1,j}$)

________________________

$c \Rightarrow d$         (let d be any class term in some $D_{k,j}$)

For example, the let *c* be *organization(X)* and $c_1$ be *government(X)* where *government(X)* is a class term that exists in $D_{1,1}$ and because *court(X)* exists in $D_{2,1}$, it can be concluded that class *"court"* is also a kind of *"organization"*.

$organization(X) \Rightarrow government(X) \lor company(X) \lor ngo(X)$
$government(X) \Rightarrow court(X)$
________________________________________________________________
$organization(X) \Rightarrow court(X)$

The second inference rule is used to obtain type-2 G-clause:

$c \Rightarrow d$
$e \leftarrow d_1 \& ...d_m$   (type-2 O-clause where *d* above satisfies $d = d_i$ for some $1 = i = m$)
__________
$e \leftarrow c$

As a continuation of the example above:

$$government(X) \Rightarrow court(X)$$
$$g1 \leftarrow court(X)$$
______________________
$$g1 \leftarrow government(X)$$

It can be concluded that besides being a court, *g1* can also be considered as a governmental unit.

The third inference rule, used to obtain type-3 G-clause, has two forms:

$$props(e,[..., p(e,t),...])$$
____________________________
$$hasprop(e, p(e,t))$$

$$props(c(X),[..., p(X,t),...])$$
$$e \leftarrow c(X)$$
____________________________
$$hasprop(e, p(e,t))$$

Consider the following example:

$$props(court(X),[name(X,t)])$$
$$c1 \leftarrow court(X)$$
____________________________
$$hasprop(c1, name(c1,t))$$

*court(X)* has the property "*names*" and because the instance $c_1$ is a *court(X)*, therefore $c_1$ also has the property "*name*" like its class.

The fourth and last type of inference rule sanctioned by XI language involves the conjunction and disjunction of the previous three rules in the form of:

$$G_1$$
$$G_2$$        $$G_1$$        $$G_2$$
_______     _______     _______
$$G_1, G_2$$     $$G_1; G_2$$     $$G_1; G_2$$

The left-most rule states that in order to come to a conclusion for a conjoined goal, all the individual goals separated by the operator "*,*" must be established first. To establish disjoined goals, illustrated in the remaining two rules, at least one goal in the set of goals grouped together by the operator *";"* needs to be proven true. For example, the following conjoined goal attempts to prove or disprove that *"e1 is a court and has the name mahkamah_majistret_shahalam"*:

$$e1 \leftarrow court(X), hasprop(e1, name(e1, mahkamah\_majistret\_shahalam))$$

For it to be true, the first goal and second goal need to be separately established:

$G_1$: $e1 \leftarrow court(X)$
$G_2$: $hasprop(e1, name(e1, mahkamah\_majistret\_shahalam))$

### 2.3.6 Advanced Reasoning Features

Advanced reasoning is what constitutes intelligent reasoning. Some of the current state-of-the-art in advanced reasoning includes answer explanation, intensional answer description, question relaxation and realization of information fusion (Gaasterland *et al.*, 1992). Answer explanation is performed when users have false or unclear understanding of what he or she is asking for (Benamara & Saint-Dizier, 2003). Question answering systems using basic reasoning capacity for example, will provide a direct answer *"no"* to the question *"Which bus should I take to reach Kota Kinabalu from Kuala Lumpur?"*. However, with the ability of answer explanation, the answer *"there cannot be any bus between Kota Kinabalu and Kuala Lumpur because of the sea"* will be produced instead. Reasoning is needed to detect in a question false presuppositions or misconceptions that conflict with the system knowledge base. False presuppositions occur with respect to the database contents while misconceptions usually occur with respect to the database semantics. Intensional answer description makes generalization or summarization on answers that are too large in scope, making the underlying implications much clearer. The query *"Which students have good marks in web programming?"* for example, can be answered intelligently in ways like *"100% of Class A students, 25% of Class B students..."*. To realize this task, reasoning is needed for cleaning answers when they partly overlap and for determining whether an answer is more specific than the other and for organizing mutually consistent answers. In question relaxation (Benamara & Saint-Dizier, 2004), neighborhood information corresponding to the question is used while reasoning for the answer. For example, the question *"What are the Chinese restaurants available in Penang?"* would have a too small set of solutions and will be more informative if the scope is extended to provide information on hawker centers and restaurants of other kinds. One way to increase the yield of potential answers is to find within the ontology, a set of most appropriate concepts which are conceptually close to the relaxed concept in the initial question. Three types of relaxation techniques available are namely rewriting predicate, broadening of the domain of a variable and breaking a join dependency.

## 2.4 Summary

In this chapter, we presented the theories and current practices in the fields related to question answering namely natural language understanding, and knowledge representation and reasoning. In a nutshell, to achieve natural language understanding in machines, computationally efficient and accurate implementation of the various levels of analysis in the linguistic model is necessary. Currently, many problems in the lower level have been solved and researchers are now redirecting their focus to higher levels such as discourse and pragmatics. As the ultimate aim of natural language understanding is to be able to comprehensively achieve machine representation of the meaning in natural language information, knowledge representation formalisms are essential. Knowledge representation comes in two parts: what and how to represent, and how to manipulate whatever that is represented to serve some purpose. The latter is generally referred to as reasoning capability. Accordingly, the fields of natural language understanding, and knowledge representation and reasoning have everything that are necessary for domain-oriented question answering systems, where the ability to "understand" and to "reason" with whatever that the machines "understood" is all that matters. Some of the existing approaches discussed in this chapter have been considered for use during the solution framework design in Chapter 4.

# Chapter 3

# Review of Prominent Question Answering Systems

## 3.1 Introduction

This chapter presents the reviews of the two groups of prominent question answering systems, which constitutes Phase 1 and 2 of the research methodology. These reviews are essential to support the claims regarding the initial and resultant problems formalized in Chapter 1. The first review tries to demonstrate how the practices by systems based on shallow natural language processing and information retrieval have contributed to the root of the initial problem in this research. The second review studies existing systems that exhibit certain traits of solving the initial problem and to what extent are these current practices based on natural language understanding and reasoning capable of it without introducing additional problems. Two question answering systems from the first approach namely Webclopedia (Hermjakob, 2001) and AnswerBus (Zheng, 2002b) are selected for this review. The next half of the chapter introduces and reviews two question answering systems from the second approach namely START (Katz, 1997) and WEBCOOP (Benamara, 2004). Discussions based on these reviews are done at the end of the chapter to highlight notable contributions that can be applied in this research and pinpoint the common drawbacks. The causes behind these drawbacks and reasons on why these drawbacks are left as it is are also presented.

## 3.2 Shallow Natural Language Processing with Information Retrieval

Despite the various architectural view presented by the author of the respective systems, all the two systems will be scrutinized by mapping their physical functionalities onto the common model based on the work of Pasca *et al.* (2000) for modern-day question answering system which comprises of three phases namely question analysis, information source processing and answer extraction, as shown in Figure 3.1.

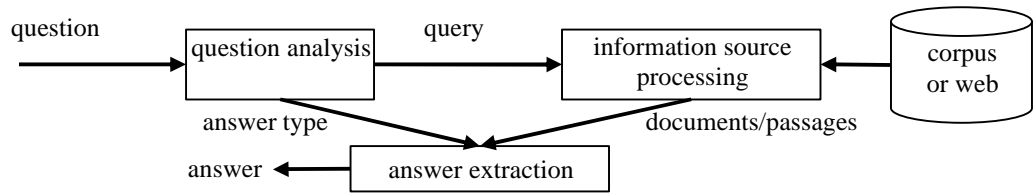

Figure 3.1: General architecture of the question answering system based on shallow natural language processing and information retrieval

## 3.2.1 Webclopedia

Webclopedia[1] (Hermjakob, 2001) consists of eight distinguishable functional components namely question parsing, query formulation, information retrieval engine, document segmentation, segment ranking, candidate answer selection, candidate answer parsing and question-answer matching. The first two components perform the various functions of question analysis. The next three perform functions of information source processing and the last three carry out answer extraction. The architecture is portrayed in Figure 3.2 below.

The phase of question processing involves question parsing and query formulation. In question parsing, the CONTEX (Hermjakob & Mooney, 1997) sentence parser is used to obtain a parse tree consisting of part-of-speech, syntactic categories and constituents' role for questions. Another reason for choosing CONTEX is because of its inbuilt capability to recognize the semantic type of the desired answer or QA type. For example, other than the normal syntactic information, CONTEX will also produce the QA type PROPER-PERSON for the question *"Who is the fourth Prime Minister of Malaysia?"* meaning that the answer must contain names of individuals. BBN's IdentiFinder (Bikel *et al.*, 1999) is used to categorize all proper names in the parsed question into person, organization, role, location. For example, *"fourth Prime Minister"* will be tagged as role and *"Malaysia"* as location. Then, query formulation is done using the parsed question where WordNet is employed to expand the query terms and place all into a Boolean expression. For example, the term *"prime minister"* is expanded to *"(prime&minister) |(head&of&state)|premier|leader"*. The author of Webclopedia (Hovy *et al.*, 2000) acknowledged that it is obvious that such brute force expansion has undesirable effects.

---

[1] An online compilation of questions and answers can be accessed via http://www.isi.edu/natural-language/projects/webclopedia/demo.html. Last retrieved on 24 February 2005.

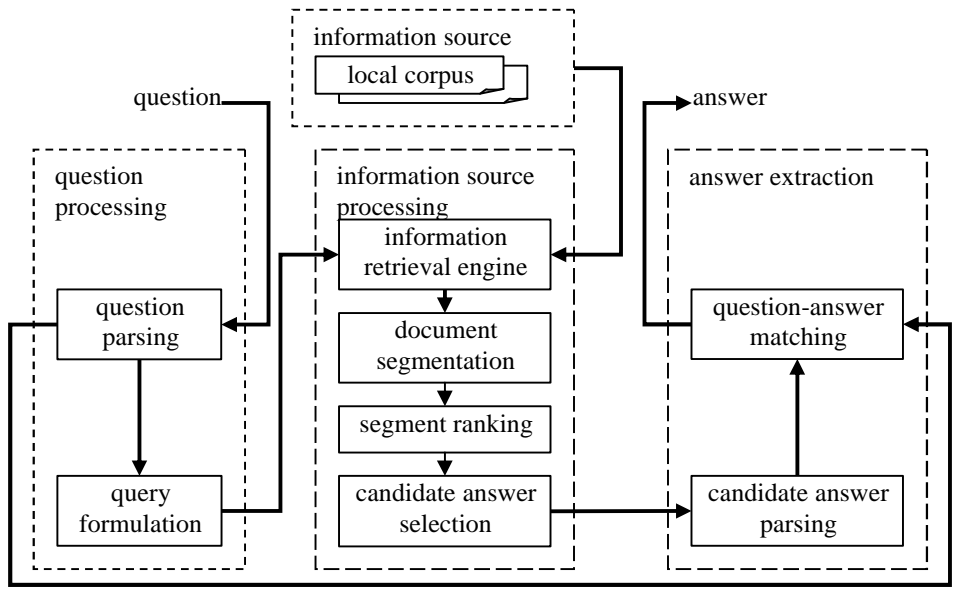

Figure 3.2: Architecture of Webclopedia

After having the search engine query, the activity moves on to information source processing where the information retrieval engine MG (Witten *et al.*, 1999) is used to search the entire local corpora and in the evaluation case, the TREC-9 corpus is used. Based on our previous example, the query *"((prime&minister)|(head&of&state)|premier|leader)&Malaysia"* will be submitted. Then document

segmentation is performed using TextTilling (Hearst, 1994) to split each retrieved document into topical segments followed by segment ranking where scores are assigned if a word in the segment matches with a word in search engine query. Different type of words gets different scores:

- each word in segment that matches a word in question get 2 scores;
- each word in segment that is synonymous with a word in question get 2 scores; and
- others get 0 score

Let $k_i$ be words that has corresponding match or synonym in the question and $w_i$ be the score for $k_i$. Assuming there are $n$ different words in the question and $m$ matches or synonyms for segment $s$, then the score for the segment $s$ is:

$$D(s) = \frac{\sum_{i=0}^{m} w_i}{n}$$

For the final function in information source processing, candidate answer selection is performed by selecting the top hundred segments based on the score $D(s)$ as candidate answers. First function in answer extraction is candidate answers parsing using CONTEX to obtain parse trees for the candidate answers and BBN's IdentiFinder to isolate proper names in the candidate answers into person, organization and location. The actual work load of answer extraction is with the question-answer matching due to the adoption of multiple matching attempts for pinpointing answers.

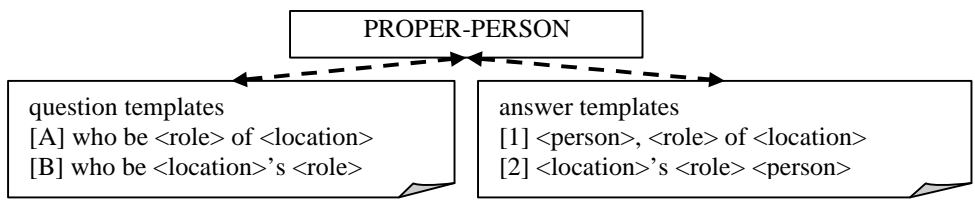

Figure 3.3: Question and answer templates of Webclopedia for QA type PROPER-PERSON

Webclopedia implements what is known QA Typology (Hovy *et al*., 2002a; Hovy *et al*., 2002b) consisting of hierarchically arranged QA types. Each QA type is associated with patterns of expression for both questions and answers in the form of templates.

The first matching approach exploits the QA type and tagged proper names from question parsing whereby attempts to find matches in question templates are performed. The question and answer template for the QA type PROPER-PERSON for example, is shown above in Figure 3.3 and attempts to find matches using the original question with tagged proper names *"fourth prime minister"* and *"Malaysia"* is performed. *"Who is the fourth prime minister of Malaysia?"* will match question template A and *"<role>"* is filled with *"fourth prime minister"* and *"<location>"* with *"Malaysia"*. Then all answer templates under PROPER-PERSON will be instantiated with the value of *"<role>"* and *"<location>"*. The instantiated answer templates will then be matched against all the candidate answers. Another approach is also employed through the collective use of QA types, part-of-speech labels and role from the question parse tree. These are known as Qtarget. For QA types like PROPER-PERSON, tagged proper names from the parsed question are used as Qargs to increase chances of positive matches. Qtarget is matched against all candidate answers' parse tree elements like QA types, part-of-speech labels and role. Qarg is matched against the proper names from the parsed candidate answers. Finally, candidate answers that passes the question-answer matching function are presented as answers.

## 3.2.2 AnswerBus

AnswerBus[2] (Zheng, 2002b) consists of eight distinguishable functional components namely translation, search engine selection, search engine specific query formulation, question classification, web search engine, document segmentation, candidate answer selection and candidate answer scoring. The first four components perform the various functions of question analysis. The next three perform functions of information source processing and the last carries out answer extraction. All the functions are clearly depicted in Figure 3.4.

---

[2] An online demo can be accessed via http://answerbus.com/systems/index.shtml. Last retrieved on 23 February 2005.

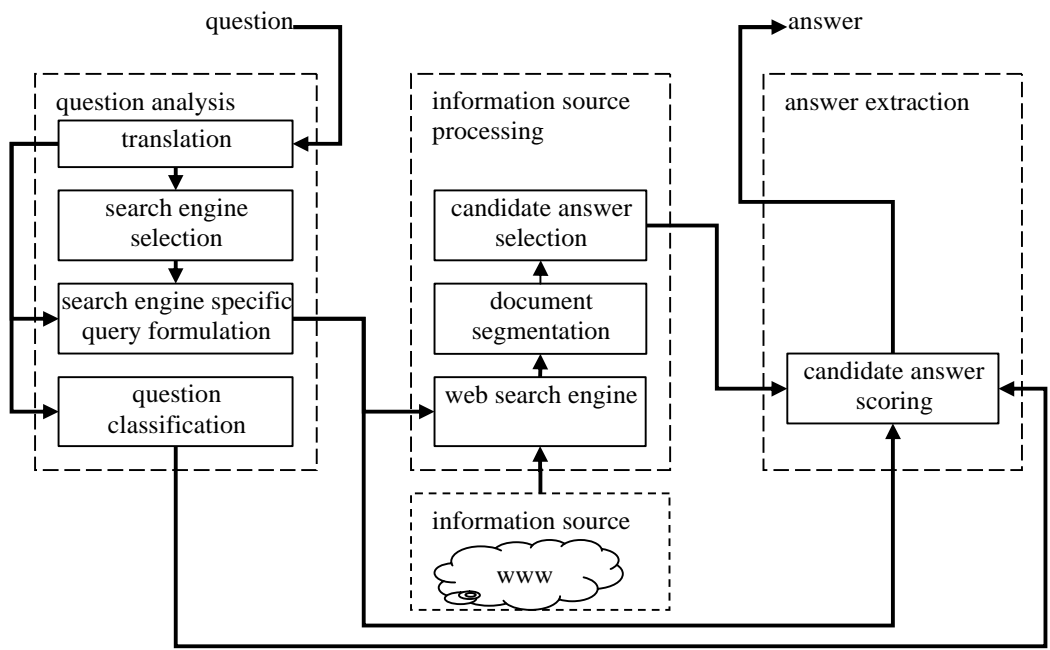

Figure 3.4: Architecture of AnswerBus

The first phase in AnswerBus begins with translation using BabelFish[3] to handle questions in languages other than English. Then, search engine selection is done by finding the corresponding search engine answer count for all words in question from the pre-answered question list. The sentence is first parsed to obtain syntactic information and the token words. For example, the question *"How high is Mount Kinabalu?"*, will yield the corresponding answer count from the list:

---

[3] Can be accessed via http://babel.altavista.com/. Last retrieved on 24 February 2005.

*how - Google(18), Yahoo(19), AltaVista(18)*

*high - Google(8), Yahoo(7), AltaVista(6)*

*is – Google(25), Yahoo(23), AltaVista(24)*

*mount - Google(3), Yahoo(3), AltaVista(3)*

*kinabalu - Google(2), Yahoo(2), AltaVista(1)*

Thus, in the example question, all the answer count will be summed to produce Google(18+8+25+3+2), Yahoo(19+7+23+3+2) and AltaVista(18+6+24+3+1). *N* search engines will be selected based on the counts. With the choice of search engines in hand, search engine specific query formulation will be carried out using various approaches. One of them is functional words deletion whereby words such as prepositions, determiners/pronouns, conjunction, interrogative words are deleted. A word frequency table is compiled in AnswerBus for use as reference in deleting frequently used words in questions. This approach is based on the assumption that words which are more frequently used in a language are less discriminating. Word form modification is also used to convert some words like verbs in questions to another form before these modified words are in the query.

The last function of question analysis is question classification where the question is classified into one of the predefined question type. The implementation in AnswerBus has two level of specificity. First, questions are classified into different question type. Questions like *"How far..."* and *"How close..."* for example, are classified as *DISTANCE*. Then, through the use of a QA specific dictionary, the expected unit of answers for the question is determined. Using the same example, the answer unit for *"How high..."* can be *"mile"*, *"kilometer"* or others and for *"How close"*, the possibilities are *"inch"*, *"centimeter"* and so on.

After submitting the search engine specific queries to the different web search engines, information source processing begins by document segmentation. The retrieved documents are first segmented through the use of some HTML separator tags as sentence boundary. Different formatting exceptions are also taken into consideration. Then, candidate answer selection commences through the use of the following formula $q \geq \sqrt{Q-1} + 1$, where $q$ is the number of matching words in the sentence and $Q$ is the total number of words in the query.

Assuming that after going through the process of query formulation, the previous example of *"How high is Mount Kinabalu?"* becomes *"high&mount&kinabalu"*. Thus, the query has three words making $Q = 3$. This means that the number of words in the segments must be more than or equal to 3. All segments that satisfy the inequality will then be awarded scores based on the number of matching words they contain. Only those segments with non-zero score will be passed on as candidate answers to the next phase.

There is only one function in the answer extraction phase namely candidate answer scoring which heavily depends on the question types and answer unit from question classification. Various influencing factors are used to refine the score from previous phase like basic named-entities extraction, coreference resolution, hit position and search engine confidence and redundancy. Basic named-entities extraction is employed to find entities that match question type. Candidate answers with entities that match question type will receive higher score. Scoring will also be affected by the appearance of the answer unit in the candidate answer. Coreference analysis is used to assign scores to the proceeding sentence of a sentence that has pronouns detected. The score of a candidate answer will also be affected by the position of its original document in the search engine ranking. A candidate answer from a document returned by a search engine with better a confidence score will get a higher score.

Finally, all candidate answers are compared to every other candidate answers and those with higher redundancy are given higher score, assuming that truth will prevail over fictions. The candidate answers will be ranked according to their scores and presented to the users.

## 3.3 Natural Language Understanding with Reasoning

The other approach of question answering based on reasoning and natural language understanding comprises of three parts namely documents understanding to produce facts which will be integrated into the knowledge base, questions understanding and finally, reasoning using facts and rules to look for answers from the knowledge base (Fischer, 2003). The graphical representation of the architecture is depicted in Figure 3.5. The practice of natural language understanding is widely reflected through the use of understanding modules for both question and information source.

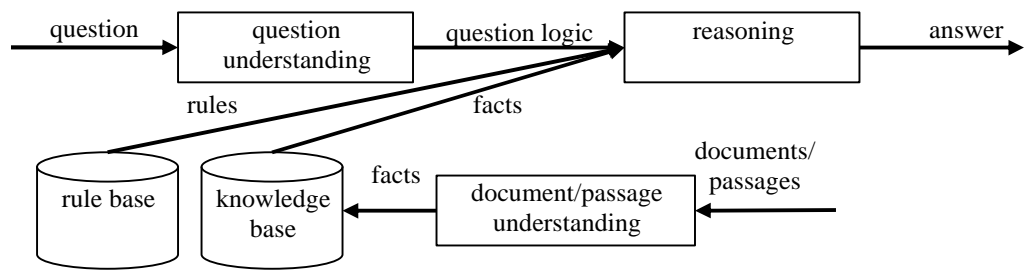

Figure 3.5: General architecture of question answering system based on natural language understanding and reasoning

### 3.3.1 START

START[4] (Katz, 1997) has five distinguishable functions namely sentence parsing, T-expression generation, wh-movement reversal, T-expression matching and answer generation, as illustrated in Figure 3.6 below. The document/sentence understanding phase uses sentence parsing and T-expression generation. Similar to document/sentence understanding, T-expression generation is also a must for question understanding, together with wh-movement transformation. Lastly, due to the simplicity of the reasoning phase of START, only T-expression matching is performed.

---

[4] An online demo can be accessed via http://www.ai.mit.edu/projects/infolab/ailab. Last retrieved on 24 February 2005.

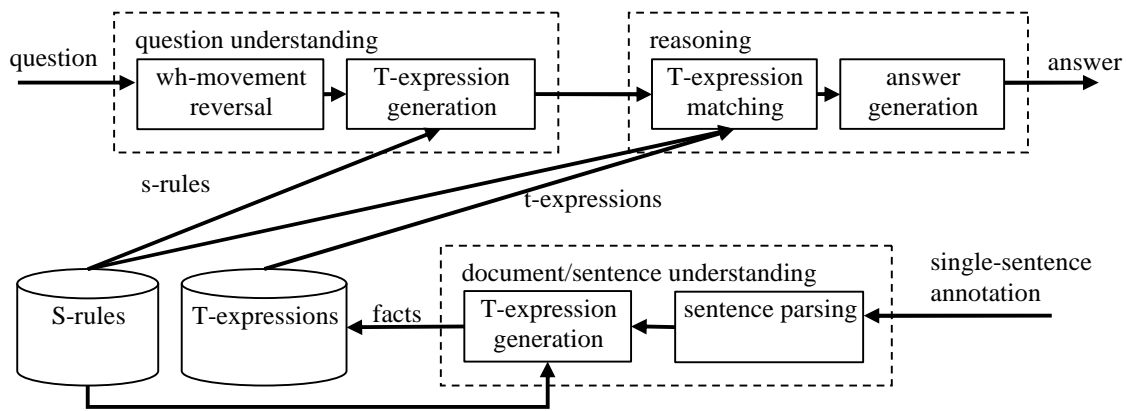

Figure 3.6: Architecture of START

The knowledge base in START consists of ternary expressions or in short, T-expressions in the *<subject relation object>* form. Conceptually, T-expression in START is similar to the ternary expressions adopted by many researches as explained by Lin (2001). START allows any T-expression to take another T-expression as its subject or object. For example, the sentence *"Wilson presented Joe with a gift"* will produce:

*<<Wilson present Joe> with gift>*

As for the rule base, it is made up of rules known as S-rules that make explicit the relationship between alternate realizations of the arguments of verbs in the form of IF-ELSE statements. S-rules are important to handle sentences differing in their surface syntax but similar in meaning. For example, the sentence *"Wilson presented a gift to Joe"* will generate:

<<Wilson present gift> to Joe>

Even though both sentences bring the same meaning, they are not considered similar by the system unless S-rules are used. Only with the below S-rule for the verb *"present"* can the above T-expressions be considered semantically equivalent:

IF <<subj present obj1> with obj2>

THEN <<subj present obj2> to obj1>

S-rules can operate in two modes namely forward and backward. A feature worth noting is the use of semantic class for verbs so that the S-rules can be generalized for use. For example, the verb *"present"*, *"reward"*, *"offer"* and so on belongs to the same semantic class and thus, the above S-rule can be generalized to cater to verbs other than *"present":*

IF <<subj verb obj1> with obj2>

THEN <<subj verb obj2> to obj1>

PROVIDED verb $\in$ act_of_giving class

After having exposure to the concept of S-rules and T-expressions, the functionalities of the systems are presented. In the document/sentence understanding phase, sentence parsing is done to obtain the parse tree structure of sentences. No information concerning the parser was provided by the author of START (Katz & Levin, 1988). Then, by using the grammatical relationships, T-expression generation will be performed by locating the subject, verb and object and tying them together in the *<subj verb obj>* form. Certain other grammatical categories like adjectives, possessive nouns and prepositional phrases are used to create additional T-expressions in which prepositions and several special words will serve as relations. For example, the sentence *"Joe's present"* can be equivalently represented in T-expression as:

*<present related-to Joe>*

The remaining words and other syntactic information from the parsed sentence are recorded in a representational structure known as history, which will be associated with the corresponding T-expression when it is stored in the knowledge-base. Another important point is that the T-expression generation that functions with S-rule in forward mode behaves differently. Consider the following sentence to be converted into T-expression:

*Wilson presented Joe with a gift*

Without the S-rule forward mode, the corresponding T-expression:

*(1) <<Wilson present Joe> with gift>*

would have been incorporated into the knowledge base. Assuming we have the S-rule:

*(2)    IF <<subj present obj1> with obj2>*

*THEN <<subj present obj2> to obj1>*

and it is functioning in forward mode, *(1)* would be intercepted and augmented based on the rule to produce:

*(3) <<Wilson present gift> to Joe>*

which will be immediately added to the knowledge base.

In the next phase where question understanding is carried out, two functions are performed. When a question is asked, a wh-movement reversal is done to undo the effects of the wh-movement transformation used to create English wh-questions. This is done by locating the place in the question where the wh-words like *"who"*, *"whom"* and *"where"* comes from and then relocating the word to that position. For example, in the question:

*(4) "Whom did Wilson present with a gift?"*

the wh-word *"whom"* needs to be relocated to where it belongs, which is between *"present"* and *"with"* resulting to *"Wilson present whom with a gift?"*. Once the question has been transformed into a declarative sentence, T-expression generation is applied on the declarative sentence. In our example, *"Wilson present whom with a gift"* will be transformed to:

*(5) <<Wilson present whom> with gift>*

As the T-expression generation in this phase is identical to the one in document/sentence understanding, if the S-rule forward mode is activated, then attempts to restructure the T-expressions will be performed. In the final phase, which is reasoning, T-expression matching is performed on the T-expression of the question and T-expressions in the knowledge base. If a request which cannot be answered directly comes in, then S-rule operating in backward mode will be triggered.

Assuming only the T-expressions *(3)* exists in the knowledge base and when T-expression *(4)* for question *(5)* is matched against the knowledge base, no results will be returned. When no answers are found, the S-rule in backward mode is triggered. The S-rules are consulted and *(4)* is found to match the IF clause in S-rule *(2)*, causing the T-expression for question *(5)* to be restructured into:

*(6) <<Wilson present gift> to whom>*

The new T-expression for the question is matched against the knowledge base once more and *"whom"* in *(6)* will match *"Joe"* in *(3)*. Finally for answer generation, the two options available are construction of English sentence and use of pointers to HTML pages, with the latter being more common.

## 3.3.2 WEBCOOP

WEBCOOP (Benamara, 2004) has seven distinguishable functions namely information retrieval, document parsing, question parsing, logical matching, query relaxation, intensional answer generation and answer generation. The first two functions are used to fulfill the processing needs of document/sentence understanding phase. Question understanding phase employs only question parsing and as for the reasoning phase, logical matching is used together with various advanced reasoning functions like query relaxation and intensional answer generation. Answer generation is used to transform the logical form of the responses into natural language answers. All the functions are placed together in Figure 3.7.

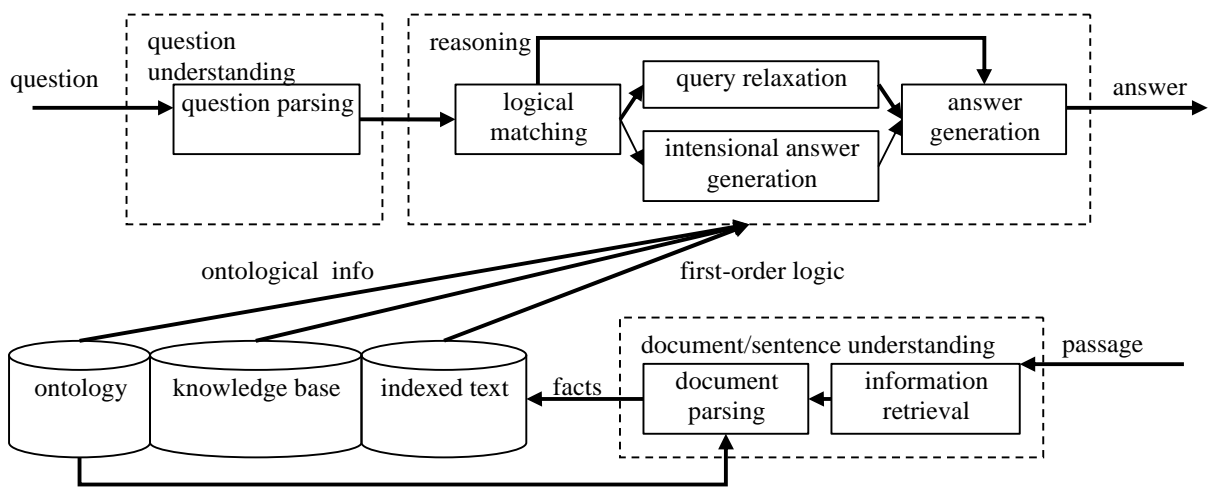

Figure 3.7: Architecture of WEBCOOP

Three important knowledge sources are implemented in WEBCOOP namely ontology, a knowledge base for general-purpose use and an indexed text of domain knowledge. The knowledge base describes the general purpose knowledge about the domain of tourism and contains basic information like country names, famous places, airlines and etc and rules. For example, facts like:

*country(malaysia), capital(malaysia, kuala_lumpur), carrier(malaysia_airline)*

and rules like:

*total_cost(hotel_name, num_nights, total):- fare(hotel_name, rate), total is num_nights*rate.*

The ontology is a conceptual collection of nodes with the associated properties. It defines the scope for extracting domain knowledge. The upper-level of WEBCOOP ontology is depicted in Figure 3.8. Each node is represented by the predicate *"onto-node(concept, properties)"*. The link between nodes can be specified through *"onto-link(subclass, superclass)"*. For example, for the *"hotel"* node, we have *"onto-node(hotel, [fare, capacity])"*. The ontology contains facts that can be interpreted through hierarchical relations and rules as simple constraints for property values. For example, the hierarchical relation between *"hotel"* and *"tourist accommodation"*, which is *"onto-link(hotel, tourist_accomodation)"* can be seen as a fact *"hotel is a kind of tourist accommodation"*. As for the rules, they are specified in the form of property value constraint like *"constraint([hotel(X), capacity(X,C), C>4], fail)"*.

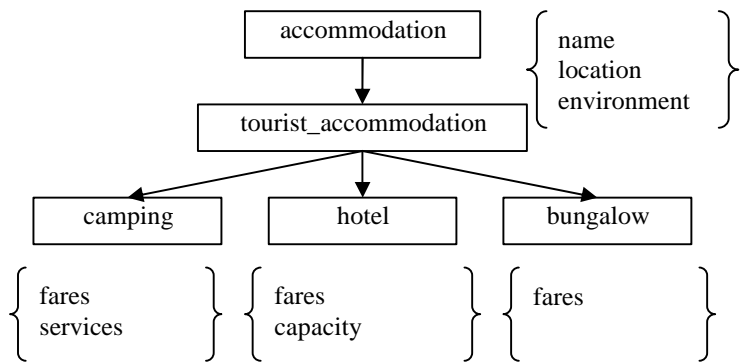

Figure 3.8: Upper-level of WEBCOOP ontology

A feature in WEBCCOP that is unique in conception concerns the use of local grammars which are associated with nodes for domain knowledge extraction from web pages. The basic principle is that each major node of the ontology is associated with a dedicated local grammar that recognizes, in the textual part of a web page, information relevant to the concept associated with that node. Grammar rule that extract knowledge relevant to the *"environment"* property of concept *"accommodation"*, for example:

*environment([at(place, X, Y)]) ? prep(fixed-loc), gap, lex(Y, noun, Sem_Type), {subsume(phys-loc, Sem_Type)}.*

*prep(fixed-loc)* is any preposition that describes a fixed localization (i.e. at, on, near). The *gap* allows the parser to skip any irrelevant string till a word denoting a physical location is found. Indexed texts contains domain knowledge in the form of *text(F, http)* where *F* is a first-order formula that represents knowledge extracted from a web page and with address *http*. For example, indexed texts about the environment of Sunway Hotel have the following form:

*text( at(place, sunway_hotel, Y) ∧ shopping_complex(Y) ∧ theme_park(Y) ∧ localization(sunway_hotel, in(kuala_lumpur)), www.sunway-hotel.com.my ).*

After having an overview of the storage structures of WEBCOOP, the functionalities of WEBCOOP are presented. In document/sentence understanding, an information retrieval function is assumed from the authors' statement that *"We assume that the most relevant documents to the user's question are found using standard information retrieval techniques"*. Moreover, no further details about the information retrieval techniques are provided by the author. Then, a document parsing is performed on the web pages retrieved to obtain the domain knowledge as specified by the local grammars in the ontology. These grammars are written by hand from corpus samples, though automatic construction grammars are currently being studied. The parser can either extract all information that it can reliably find or just those related to pre-selected properties of nodes. The grammar is based on the discontinuous grammar formalism which includes gaps which are variables in the rule that stand for a finite set of words to skip till a certain word or condition is met. The parser runs in a bottom-up fashion, allowing for the recognition of text fragments distributed throughout a whole paragraph or web page. After documents are parsed, the extracted information is transformed into the following logical representation and stored in the indexed text collection.

Question parsing is the only function in the proceeding phase, question analysis. Question parsing produces three important information namely the type of the query either yes/no, Boolean or entity, question focus and construction of its semantic representation in first-order logic. The question parser is a bottom-up parser that produces a conceptual representation of questions. The parser aims to keep track of the terms used in the question as much as possible in order to re-use them in the response. For example, the question *"Give me the Sunway Hotel rates in Kuala Lumpur?"* has the following semantic representation:

*(Quantity, X: listof(rates), hotel(sunway_hotel) $\wedge$ in(place, sunway_hotel, kuala_lumpur) $\wedge$ rates(sunway_hotel, X) ).*

The reasoning phase is perhaps the main focus of WEBCOOP where they attempt to integrate explanations and justifications as part of the natural language response to make the mechanism that led to the answer explicit. The first function in reasoning is logical matching where, given a fragment of text, how can we infer that it is an answer to a question. There are two different ways of doing it namely from the deductive knowledge base where responses are variable instances and from the indexed text base where responses are formulae which unify with the query formula. In the latter case, the unification proceeds as follows. Let $Q$ (conjunction of terms $q_i$) be the question formula and F (conjunction of $f_j$) be a formula associated with an indexed text. $F$ is a response to $Q$ if and only if for all $q_i$ there is an $f_j$ such that:

- $q_i$ unifies with $f_j$; or

- $q_i$ subsumes via the concept ontology $f_j$ (e.g. *means-of-transportation(Y)* subsumes *bus(Y)*); or

- $q_i$ rewrites via rules of the knowledge base into a conjunction of $f_j$. (e.g. *airportof(Z, kuala_lumpur)* rewrites into *airport*(Z) $\wedge$ *localisation*(Z, *in*(*kuala_lumpur*))).

While matching is done, incoherence can be detected due to users' misconceptions that are reflected in their queries. There are two types of incoherence namely failure to satisfy constraint and non-existence of relations or entities. For example, the question *"A hotel room in Kuala Lumpur for 8 persons?"* will fail to have any direct answers because of the constraint:

$$constraint([hotel(X), capacity(X,C), C>4], fail).$$

Instead, an explanation *"A hotel room capacity is less than 4 person"*. Another example question *"Give me the list of highland resorts in Melaka?"* will similarly have incoherence detected because the user presupposes the existence of an entity that does not exist in the knowledge base. An explanation is given as *"There are no highlands in Melaka"*. The logical forms of all the explanations are passed on to answer generation, contributing to the first part of the response.

Depending on the result of logical matching, the second part of the response is generated to provide flexible solutions for the users to go further when there is incoherence or when the set of initial responses is too large. The second part of the response is achieved through the use of advance reasoning techniques of query relaxation and intensional answer generation. Following the question *"Give me the list of highland resorts in Melaka"* for example, due to its incoherence resulting from the matching process, the query relaxation will attempt to provide alternatives or suggestions to the users like *"List of beach resorts in Melaka"*, *"List of city hotels in Melaka"* and so on.

For intensional answer generation, consider the question *"What are the tourist accommodations available Kuala Lumpur?"*. Normal question answering systems would have provided direct answers like *"There are 123 tourist accommodations in Kuala Lumpur"* but with intensional answer generation, the initial response would be segregated into *"56 hotels in Kuala Lumpur"*, *"12 chalets in Kuala Lumpur"* and so on. Again, the logical form of the second stage of response is passed on to answer generation.

Finally, function in the reasoning stage involves answer generation whereby using the logical form of the two stages of response from the previous functions in reasoning, a natural language response is generated and structured with hypertext links in a web page to produce a dynamic response.

## 3.4 Summary

Besides sharing the same aim of exploiting information of the World Wide Web for question answering, the two systems based on shallow natural language processing and information retrieval have the same appearance. They parse questions, not to obtain the meaning representation, but to get the type of the questions and answers and a set of keywords usually consisting of some noun phrases or verbs. Web search engines are used to retrieve web pages based on queries formed by the keywords from the original questions. Then, some segmenting is performed to reduce the size of the texts to lessen the burden of the parser. Again, some keywords from the candidate answers are identified and scoring algorithms are devised based on those keywords and some other factors to figure out which among the many passages answer the question.

Through the way question and information are handled, it is very obvious that the philosophy behind the working of these systems are quantity and not quality. For this approach of question answering, diversity of question and quality of response are factors that must be compromised in order to handle large volume of open-domain information. Even at this stage of technological advancement, there is still no better way to harvest online information than the readily available web search engines, whose credibility have long been proven. Moreover, when higher-level of natural language understanding and reasoning is not possible on open-domain free text, researchers have no better options than to employ statistical methods with mere syntax processing. But, regardless of what the reasons are, the nature of questions and responses has been overly restricted as a result of the adoption of mere low-level natural language processing like syntax processing and information retrieval applications like web search engines.

As for the two systems from the second approach, START and WEBCOOP signify two very different efforts geared towards question answering based on natural language understanding and reasoning. Both START and WEBCOOP employ natural language understanding at the basic level. START follows the convention in understanding natural language and WEBCOOP uses a totally different practice.

Natural language understanding in WEBCOOP is performed through pattern matching using hand-coded dedicated local grammars that are applicable only to properties of concepts in the domain ontology. In short, it is not a linguistic grammar like link grammar or principle-based grammar and is not applicable to general natural language information written in English. Rather, such grammar is domain-specific and will add on to the burden of scaling across multiple domains. Moreover, the meaning extracted using such grammar are stored in the form of indexed text in first-order logic and not some higher-level knowledge representation formalism. The natural language understanding in START is done using some linguistic-based parsers that focuses on grammatical information rather than domain-specific properties. This makes START easily scalable across multiple domains as demonstrated in its online question answering system. Because the information source is limited to sheer hand-coded annotations that represent the actual information, the parser is expected to be straightforward from the absence of the need to handle full semantic and discourse analysis. Thus, the parsing will not be able to scale to real-world information that has posed various hurdles to the field of natural language understanding since the very beginning. As for the reasoning mechanism in WEBCOOP, it can be considered as the true state-of-the-art and is the next move in the reasoning approach for intelligent responses. As for START, the system adopts a rule-based reasoning that deals with the literal matching of ternary expressions and rules. This approach is both effective and simple for first-order logic or ternary expressions but if we decide to employ other more powerful representation formalisms, different reasoning approach is required. Moreover, the representation formalism in START lacks the use of ontological and other domain information, which makes it impossible to introduce advanced reasoning features.

From the traits of both WEBCOOP and START, we can conclude that worthwhile efforts have been attempted that have actually led this approach of question answering to a higher level. The advanced reasoning concepts of WEBCOOP and the use of dependency information between words for ternary expression by START for example, are two important works that have been considered for use in the design of the solution framework in Chapter 4. Nonetheless, their practices towards low-level natural language understanding, choice of inexpressive representation formalisms and conventional reasoning have introduced additional problems related to scalability across domains and to real-life natural language text.

# Chapter 4

# Solution Framework Design

## 4.1 Introduction

Chapter 2 and Chapter 3 have both introduced some related algorithms and practices in question answering and also, its related two fields (i.e. natural language understanding, and knowledge representation and reasoning). This information will be useful in understanding Phase 3 of the research methodology, which is to design a solution framework that satisfies the criteria laid out in the hypothesis. The design of the framework for question answering must put into consideration the following three elements in order to solve the problems:

- full-discourse natural language understanding approach is required to ensure the ease of scalability across domains and to real-life natural language text;
- representation formalism that has good expressive capability for intrinsic properties with support of ontological information, like the semantic network is required for representing rich output from natural language understanding; and
- reasoning approach capable of fully exploiting what the representation formalism has to offer and to work with the advanced reasoning features is required.

Figure 4.1 shows the design of the solution framework, which can be divided into two main parts namely natural language understanding mechanism and reasoning mechanism, and one supporting part which is knowledge base and gazetteer.

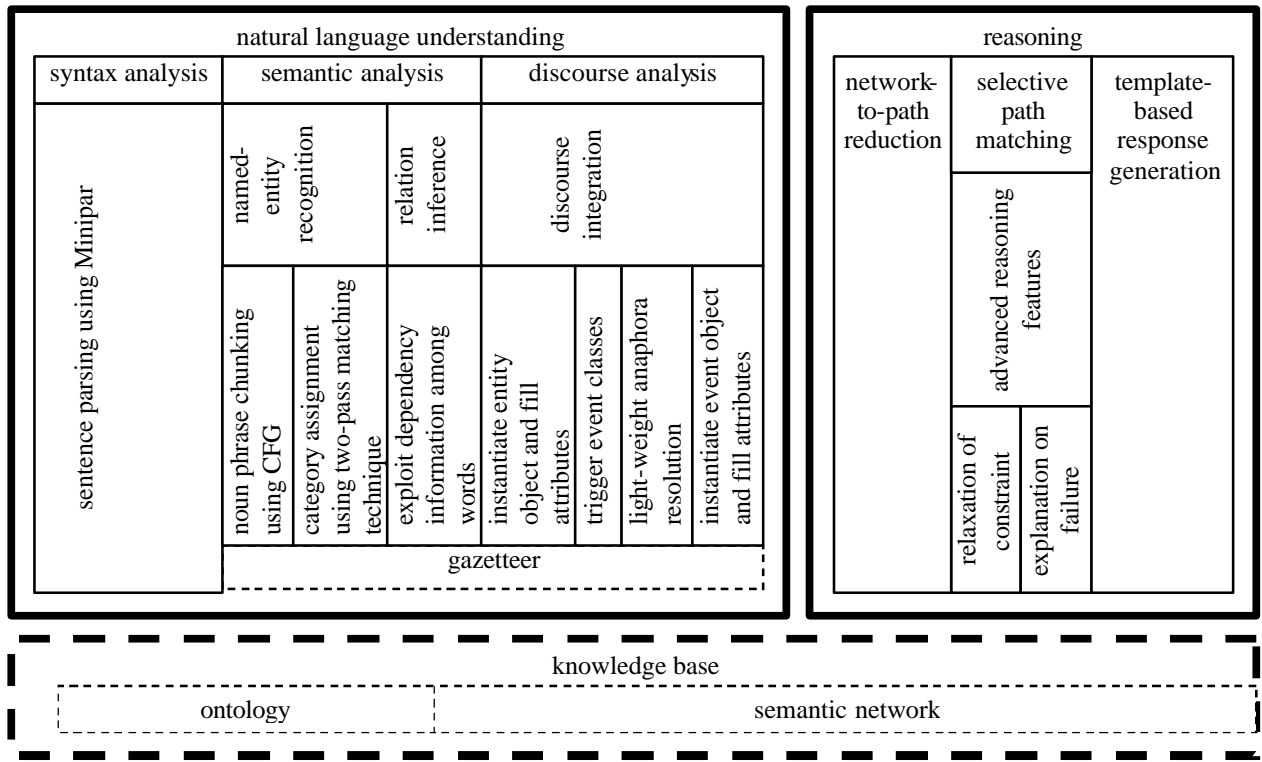

Figure 4.1: An outline of the solution framework

The design of knowledge base and gazetteer makes up the first part of this chapter. Due to the fact that ontology and gazetteer is a domain-oriented structure, their design will be focused in the domain of Cyberlaw. Also, a clear distinction has to be made concerning the three structures namely semantic network, ontology and the gazetteer. The semantic network and ontology constitutes the knowledge base and the language used for presentation in both structures is based on XI language, which was discussed in Chapter 2. XI was chosen as the representation language in this research not only for its simplicity of use, ease of implementation, flexibility and theoretically well-founded, but also for its nature of being a hybrid between frame and semantic network. It allows entities and attributes to be organized around classes, which are later interconnected based on their relationships. The semantic network is used to hold the meanings or facts in the form of network-bound ternary expression that are extracted and unified by the natural language understanding mechanism. As for ontology, it is used to define the top-level classes that are used for the creation of entities or events nodes in the semantic network. The gazetteer is a totally separate structure from the knowledge base. The gazetteer is only used for natural language understanding and even though a separate from knowledge base, the entries in the gazetteer have references to the nodes in the ontology.

The design of the natural language understanding mechanism must take into considerations the various levels of analysis up to the discourse level. Although there are existing concepts or techniques out there for various stages of analysis in natural language understanding but mostly, they are studied separately without care for compatibility in the case where these algorithms are required to be integrated for full natural language understanding. Hence, for this research, a series of algorithms based on actual theories for various stages of analysis that were designed to work seamlessly together are put forward. In syntax analysis, an existing external module for sentence parsing called Minipar, which was discussed in Chapter 2, is used. In the remaining stages of analysis, algorithms are developed, which can be entirely new or just innovative implementations of existing concepts. In semantic analysis, three algorithms were introduced; two cooperative algorithms for named-entity recognition and one algorithm based on existing concepts about ternary expression for relation inference. For named-entity recognition, context-free grammar is used for chunking noun phrases and a two-pass matching method for assigning categories to noun phrases. As for relation inference, it exploits four classes of dependency information between words to identify relations of interest. In discourse analysis, an entirely new four-stage method is developed to unify meanings of different sentences from the same discourse. One of the notable algorithms lies in the third stage for resolving anaphora. Also, note that both semantic and discourse analysis heavily utilize information from the gazetteer.

For the design of the reasoning mechanism, there are three top-level algorithms namely network-to-path reduction, selective path matching and template-based response generation. The idea behind the reasoning mechanism is based on the notion of complexity-reduction whereby a problem of answer discovery that begins with two networks namely query network and semantic network, is later collapsed into two sets of paths by the network-to-path reduction algorithm. From there on, the task of finding the answer is scaled down to the selective matching of the nodes in the paths of both sets, which is performed by the selective path matching algorithm. Also, integrated with the selective path matching are two advanced reasoning features namely relaxation of constraint and explanation on failure to enhance the reasoning process. Lastly, once the desired answer is discovered or during failure, the desired explanation is synthesized, a proper unambiguous response in English is generated using template-based response generation.

## 4.2 Knowledge Base and Gazetteer

### 4.2.1 Design of Ontology for Cyberlaw Domain

In this section, we will discuss the construction of ontology for the Cyberlaw domain using the definitional component of XI language. The process is guided using the result from the study of the reporting pattern of 64 Cyberlaw news articles from ZDNet[1]. News describing Cyberlaw cases usually contain of the information described in Figure 4.2. After having the basic information about Cyberlaw cases, they are structured into a guide with the *"legal_proceeding"* event as the main focus as shown in Figure 4.3.

Figure 4.3 shows the structured guide obtained from the information in Figure 4.2. *"{X, Y}"* in *"<Z>"* shows that both *"X"* and *"Y"* respectively are possible subclasses under *"Z"* and any *"<U>"* in *"<Z>"* denotes the use of instances from class *"U"* to fill the value of an attribute in *"Z"*.

---

[1] The news portal is accessible via http://www.zdnet.com. Last retrieved on 26 March 2005.

| Plaintiff | the party who initiated the legal action |
| Defendant | the party whom an action is brought against in a court of law |
| Judge presiding over the case | a public official authorized to preside and hear legal complications brought before a court of justice |
| Court at which the case proceed | depending on the nature of the case and the jurisdiction of the court, different cases need to be filed in different courts |
| Stage of proceeding | is it a case filing, trial, resolution or an appeal |
| Charges | the accusation brought against the defendant by the plaintiff |
| Case resolution | if the proceeding has reached the stage of resolution, then is it achieved through judgment or settlement |
| Judgment | the judicial decision after both parties has been heard and based on available evidence |
| Settlement | agreement by both parties to settle the dispute outside the court |
| Date of filing | the date on which the legal case is filed |
| Date of resolution | the date on which the legal case is settled, in or outside the court |
| Date of trial | the date on which the trial takes place |

Figure 4.2: Common information included in news on Cyberlaw cases

| <LEGAL_PROCEEDING> | plaintiff: | <PERSON>\|<ORGANIZATION> |
| | defendant: | <PERSON>\|<ORGANIZATION> |
| | proceeding_stage: | {FILING, TRIAL, RESOLUTION, APPEAL} |
| | judge: | <JUDGE> |
| | court: | <COURT> |
| | nature_of_case: | <VARIABLE> |
| | proceeding_date: | <DATE> |
| <RESOLUTION> | prevailing_party: | <PERSON>\|<ORGANIZATION> |
| | losing_party: | <PERSON>\|<ORGANIZATION> |
| <PERSON> | per_fname: | FIRST_NAME |
| | per_lname: | LAST_NAME |
| | per_profession: | PROFESSION |
| | per_desc: | ADDITIONAL_DESCRIPTOR |
| <ORGANIZATION> | org_name: | NAME |
| | org_desc: | ADDITIONAL_DESCRIPTOR |
| | org_type: | {GOVERNMENT, COMPANY, NGO} |
| <LOCATION> | city_name: | NAME |
| | state_name: | NAME |
| | country_name: | NAME |
| | region_name: | NAME |
| <DATE> | day_of_week: | VALUE |
| | day_of_month: | VALUE |
| | month_value: | VALUE |
| | year_value: | VALUE |
| <COURT> | court_type: | TYPE |

Figure 4.3: Structured guide for constructing Cyberlaw domain ontology

A point worth noting is that attributes for entity objects are atomic or in other words, cannot contain objects of other classes. However, for event classes, their attributes can only assume objects from other entity classes as their values. This is why an anomalous class *"variable"* is introduced, to act as a container for further descriptions of the *"legal_proceeding"* event.

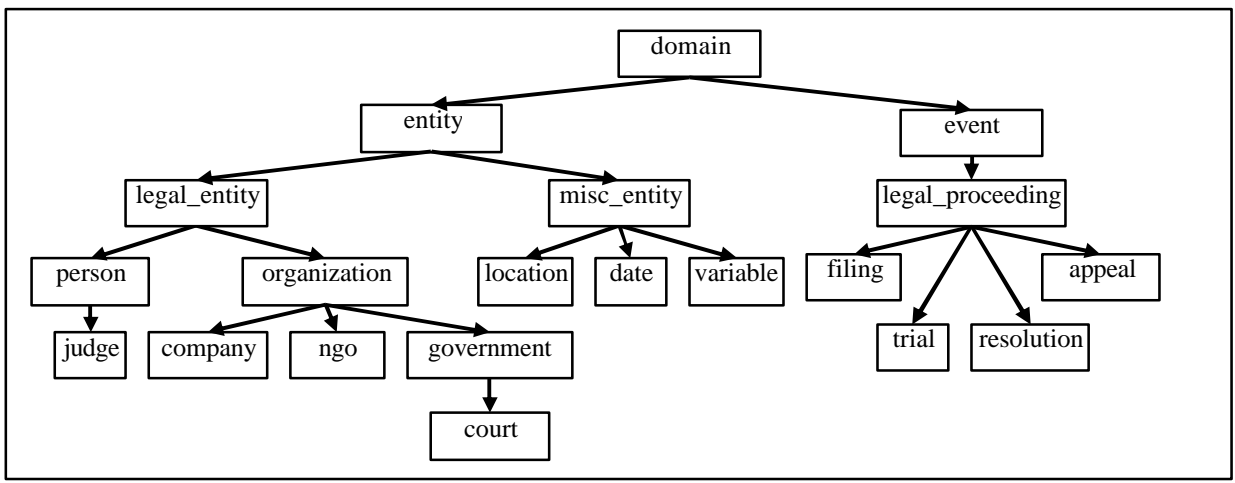

Figure 4.4: Ontology for the Cyberlaw domain

The overall ontology for the Cyberlaw domain is shown in Figure 4.4. Constructing the ontology based on the template in Figure 4.3 is carried out in stages. Firstly, the root class will be *"domain"* and followed by the next level of the ontology that will be based on the standard classes adopted by most other ontologies namely *"entity"* and *"event"*. Under *"entity"*, objects will be divided into *"legal_entity"* and *"misc_entity"* Then, the following levels will be constructed using the classes identified by the template. From the template, those objects that can assume legal responsibilities namely *"person"* and *"organization"* are placed under *"legal_entity"* and the remaining *"location"* and *"date"* are made into subclasses of the class *"misc_entity"* in our ontology. Due to the significance of the person *"judge"* in our domain, a subclass under *"person"* called *"judge"* is introduced without any new attributes of its own.

Under organization, three subclasses namely *"government"*, *"company"* and *"ngo"* which stand for non-governmental organization are brought in. Due to its role as a governmental establishment of justice, the class *"court"* will be placed under *"government"*. As for the event *"legal_proceeding"* in the template, it will be turned into a subclass under the *"event"* class. The four specializations of the stages of legal proceeding are in turn placed under their parent. Finally, all slots in each class in the template will be made into attributes in the ontology.

The hierarchical collection of classes in the ontology is then encoded using the definitional component of XI language as shown in Figure 4.5.

```
domain(X) ⇒ entity(X) ∨ event(X)
entity(X) ⇒ legal_entity(X) ∨ misc_entity(X)
      legal_entity(X) ⇒ person(X) ∨ organization(X)
            person(X) ⇒ judge(X).
            organization(X) ⇒ company(X) ∨ government(X) ∨ ngo(X).
                  government(X) ⇒ court(X).
      misc_entity(X) ⇒ location(X) ∨ date(X) ∨ variable(X)
event(X) ⇒ legal_proceeding(X).
      legal_proceeding(X) ⇒ filing(X) ∨ trial(X) ∨ resolution(X) ∨ appeal(X)
```

Figure 4.5: Classes in Cyberlaw ontology encoded using XI language

Similarly, based on the slots of each class in the template, a collection of attributes for the corresponding classes are constructed. The collection of the attributes is shown in Figure 4.6. Note that only the attributes for the top-level classes and attributes unique to lower-level classes are encoded as the inheritance property in XI language will enable lower-level classes to inherit all attributes from their superclasses. For example, the class *"organization"* has been encoded with attributes such as *"name"* and *"desc"*. Even though the classes *"court"*, which is an indirect subclass of *"organization"*, has been provided with only the attribute *"court_type"*, this does not mean that the class *"court"* does not have any other attributes. It will similarly inherit from its immediate superclass *"government"* which in turn directly inherits from class *"organization"*.

| | | | | |
|---|---|---|---|---|
| attribute( | legal_proceeding(X), | [ | plaintiff(X, person(_);organization(_)), defendant(X, person(_);organization(_)), nature_of_case(X, variable(_)), preside_by(X, judge(_)), occur_at(X, court(_)), occur_on(X, date(_)) | ]) |
| attribute( | resolution(X), | [ | prevailing_party(X, person(_);organization(_)), losing_party(X, person(_);organization(_)) | ]) |
| attribute( | legal_entity(X), | [ | desc(X, _ ) | ]) |
| attribute( | person(X), | [ | per_fname(X, _ ), per_lname(X, _ ), profession(X, _ ) | ]) |
| attribute( | organization(X), | [ | org_name(X, _ ) | ]) |
| attribute( | court(X), | [ | court_type(X, _ ) | ]) |
| attribute( | location(X), | [ | city(X, _ ), state(X, _ ), country(X, _ ), region(X, _ ) | ]) |
| attribute( | date(X), | [ | day_of_week(X, _ ), day_of_month(X, _ ), month(X, _ ), year(X, _ ) | ]) |
| attribute( | variable(X), | [ | var_desc(X, _ ) | ]) |

Figure 4.6: Attributes for the classes in Cyberlaw ontology encoded using XI language

## 4.2.2 Use of Network-Bound Binary Terms for Semantic Network

The semantic network for the solution is a network-bound list of binary complex terms based on XI language in the form of $f(t_1, t_2)$ to denote the relationships between classes, objects and attributes. The functor *f* is used to denote the relationship between an object with its attribute value using the attribute name (e.g. *org_name*) or the relationship between a class and its objects using *is_a*. As for the terms $t_1$ and $t_2$, they can be a class (e.g. *company*), object (e.g. *a2*) or attribute value (e.g. *microsoft*). Based on Table 4.1, there are only two types of binary complex terms allowable in the semantic network depending on the functor and the terms.

An example that uses the "is_a" functor is $is\_a(a2, company)$ and the second one uses an attribute's name $org\_name(a2, microsoft)$.

Table 4.1: Types of binary complex terms in the semantic network

| Functor $f$ | Term $t_1$ | Term $t_2$ | Explanation |
|---|---|---|---|
| is_a | Class | object | relationship between a class and its objects |
| attribute name | Object | attribute value | relationship between an object with its attribute value |

### 4.2.3 Design of Gazetteer for Cyberlaw Domain

The gazetteer is designed to consist of two parts where the first is used for named-entity recognition and the second is for discourse integration.

The first part of the gazetteer consists of common proper nouns like a person first name used for triggering or complete proper names like a country name. The gazetteer currently comprises of names for organizations, person, date and location. The names of companies together with their designations are compiled from Fortune 500 totaling to 865. First names for person are compiled from *"YourBabysName.com"*[2] and names of countries and cities are collected from *"The World Gazetteer"*[3]. Currently, there are 6375 first names and 6962 names of countries and cities in our gazetteer. There are two types of entries for the first part of the gazetteer namely standalone names and triggering words as shown in Table 4.2. Standalone names are stored in the gazetteer in three parts namely the name, a matching pattern and an alias, if applicable. Each entry is marked as *"specific"* to denote that the word is a valid proper name by itself.

---

[2] Site is accessible via http://www.yourbabysname.com/. Last retrieved on 26 March 2005.
[3] Site is accessible via http://www.world-gazetteer.com/st/stata.htm. Last retrieved on 26 March 2005.

Table 4.2: Two types of proper name entries in the gazetteer

|  | Standalone names | Triggering nouns |
|---|---|---|
| Location | Cities, states, countries and regions. E.g.: "Melaka", "Malaysia", etc. | NONE |
| Person | Profession indicators: "Judge", "Jury", "Chairman" and etc. | First names: "Andrew", "Sharmini", "Joe", etc. |
| Company | Company names: "AT&T", "Excite", etc. | Company indicators: "Inc.", "Ltd.", etc. |
| NGO | NGO names: "United Nations", etc. | NGO indicators: "Association", "Group", etc. |
| Government | Government institution names: "Ministry of Defense", "Court", etc. | Government institution indicators: "Ministry", "Agency", etc. |
| Date | Months and day of week: "Monday", "October", etc. | NONE |

As for triggering words, they are stored as individual words in the similar gazetteer but with an extra mark *"generic"* to denote that these words serve only as general indicators that a noun phrase might belong to the corresponding category. Following this, certain standalone names and triggering words may have conditions in the form of predefined patterns that must be fulfilled before finalizing an assignment. Both standalone names and triggering words are assigned with categories to facilitate the identification of possible entity class instances and their attributes. Table 4.3 below shows several possible types of categories for different proper name entries in the gazetteer.

Table 4.3: Categories for different type of proper name entries in gazetteer

| Entries in gazetteer | Category in gazetteer |
|---|---|
| Name of locations like cities, states, countries and regions | location |
| First names | person |
| Profession | person, judge |
| Name of companies & indicators | company |
| Name of ngo & indicators | ngo |
| Name of government institutions & indicators | government, court |
| Day of the week, months | date |

As for the part of the gazetteer used for discourse integration, the designing task involves the identification of special verbs, prepositions and even nouns that will give rise to event classes. These special words are associated to certain events in the ontology and used to relate entities together to form an event of *"legal_proceeding"* or its subclasses. As verbs and prepositions have their own associated subjects and objects, these trigger words can be used to set off the appropriate value for certain attributes in the *"event"* class. The values for the *"nature_of_case"* attribute in *"legal_proceeding"* class for example, can be filled with the noun phrase that occurs after the triggering verbs like *"accuse of"*, *"charge for"* and etc. The same goes for the *"prevailing_party"* and *"losing_party"* attribute in the *"resolution"* event whose values can be filled with noun phrases occurring after verbs like *"side with"* and *"rule against"* respectively. Besides using verbs and prepositions to trigger values to fill event attributes, verbs and nouns are also used to identify specialization under the *"legal_proceeding"* hierarchy. For example, verbs like *"won"*, *"rule"*, *"side with"*, *"sentence"*, *"settle"*, *"throw out"* and *"dismiss"* and nouns like *"ruling"* and *"closing"* are used to instantiate an object of the *"resolution"* event and verb *"plead"* and *"file"* to indicate *"trial"* and *"filing"* event respectively. All these triggering verbs are separated from other proper names in the gazetteer using the indicator *"relation"*.

## 4.3 Natural Language Understanding Mechanisms

The solution to natural language understanding consists of two parts namely the various natural language processing tasks from morphological to semantic level to extract the meaning of individual sentences, and the unification and representation of the meaning of the various sentences into a single discourse using the representation language of choice, XI.

### 4.3.1 Meaning Extraction of Individual Sentences

The collective responsibility of meaning extraction is carried out in two major phases of natural language processing namely morphological and syntactic analysis, and semantic analysis. The lowest level task for meaning extraction, following conventions in natural language understanding, is morphological analysis and followed by syntactic analysis in the next level. These two stages of analysis collectively aim to extract the structure of the sentences before semantic analysis can be done. The functions in these two stages of analysis are usually realized through sentence parsing. Initially, two external sentence parsers for English language namely Link Grammar Parser and Minipar are considered for use. In an evaluation by Wong *et al.* (2004b) to assess the performance, accuracy and quality of output of Link Grammar Parser and Minipar, Minipar turns out to be a better external module for integration. Many researchers like Lin (2001) use the Minipar in their retrieval systems. Next, the design of tasks required in semantic analysis namely word sense disambiguation, named-entity recognition and relation inference is carried out.

Word sense disambiguation is performed to distinguish between similar words or homographs of different senses. The focus will only be on distinguishing senses among homographs belonging to the same grammatical category rather than all the words because the latter has already been solved by sentence parsing. To distinguish between the judge as in *"public official in the court of justice"* and judge as in *"to pass judgment on"* for example, only a sentence parser is required because these two senses exist as different word classes. The case where the senses belong to the same grammatical category, a little more work is required. The word *"charge"* for example, can have different meaning when used as a noun. In the context of attack, the noun charge is used as *"the battle began with a cavalry charge"*. In the context of electrical phenomenon, it is used as *"the battery needed a fresh charge"*. It is obvious from the example that the task of disambiguating word senses can be made a lot easier or even to the extent of solved, when the possibilities of contexts a word can exist in is limited. In the domain of Cyberlaw for example, when the word *"charge"* is encountered, a high possibility that it is of the meaning *"a pleading describing some wrong or offense"* is extremely high. Thus, no explicit word sense disambiguation is carried out in NaLURI because the problem has been implicitly solved through domain constraint where each word with multiple senses is disambiguated when they are associated with classes or attributes of classes.

### 4.3.1.1 Context-Free Grammar for Noun Phrase Chunking

Named-entity recognition is implemented as two parts namely noun phrase chunking and category assignment. Noun phrase chunking algorithm is formalized as a context-free grammar such that a noun phrase is in the language of grammar *G* if it can be derived from it. As shown in Figure 4.7, the grammar *G* constitutes of a tuple of four sets namely *V*, a finite set of variables, *T*, a finite set of terminals, *R*, a finite set of rules and a start variable, $S \in V$.

---

$$G = (V, T, R, S)$$

where

$V = \{$<NOUN_ PHRASE>, <MODIFIER>, <END_MODIFIER>, <HEAD>, <MODIFIER_NOMINAL>, <MODIFIER_ADJ>, <DET>$\}$

$T = H \cup M$

  where  $M = \{$a set of all nouns, adjectives and determiners that modifies a head noun$\}$
        $H = \{$a set of all head nouns$\}$

$S =$ <NOUN_ PHRASE>

$R$:
  <NOUN_PHRASE> → <MODIFIER><HEAD><END_MODIFIER>
   <MODIFIER> → <MODIFIER_NOMINAL><MODIFIER>|
        <MODIFIER_ADJ><MODIFIER>|<DET><MODIFIER>| e
  <END_MODIFIER> → <MODIFIER_NOMINAL><END_MODIFIER>| e
    <HEAD> → *{a set of all head nouns, H}*
 <MODIFIER_NOMINAL> → *{a set of all nouns that modifies a head noun, N| N ∈ M}*
  <MODIFIER_ADJ> → *{a set of all adjectives that modifies a head noun, A| A∈ M}*
    <DET> → *{a set of all determiners of a head noun, D | D ∈ M}*

---

Figure 4.7: Context-free grammar for noun phrase chunking

For example, based on the Minipar parse output in Figure 4.8, *H = {Judge}*, N = *{U.S., William Pauley III}*, *A = {District}* and *D = {The}*.

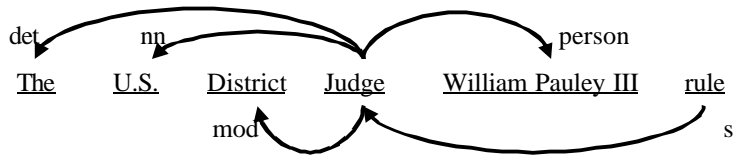

Figure 4.8: Dependency structure for the phrase *"The U.S. District Judge William Pauley III"*

The derivation sequence shown in Figure 4.9 shows that the noun phrase *"The U.S. District Judge William Pauley II"* is indeed in the language of the grammar G.

| | | |
|---|---|---|
| *<NOUN_PHRASE>* | ⇒ | *<MODIFIER><HEAD><END_MODIFIER>* |
| | ⇒ | *<DET><MODIFIER>**Judge**<MODIFIER_NOMINAL><END_MODIFIER>* |
| | ⇒ | *The <MODIFIER_NOMINAL><MODIFIER> **Judge** William Pauley III e* |
| | ⇒ | *The U.S. <MODIFIER_ADJ><MODIFIER> **Judge** William Pauley III* |
| | ⇒ | *The U.S. District e **Judge** William Pauley III* |
| | ⇒ | *The U.S. District **Judge** William Pauley III* |

Figure 4.9: Derivation of the phrase *"The U.S. District Judge William Pauley III"* using a context-free grammar

Context-free grammar is implemented entirely in Perl as a bottom-up parser using the grammatical and dependency information produced by Minipar.

### 4.3.1.2 Two-Pass Method for Category Assignment

The second part in named-entity recognition namely category assignment is carried out with the output of noun phrase chunking in a two-pass method using dependency information and a gazetteer.

As the name implies, the two-pass method used for category assignment operates in two stages of matching with increasing complexity. The two-pass method is an innovative combination of all the existing concepts for category assignment namely list-lookup, triggering and pattern as discussed in Chapter 2. The process of assigning categories is shown in Figure 4.10.

The first pass attempts a direct match for any standalone names in the gazetteer without using any patterns and positive matches will usually prevail for single-word names concerning most dates and locations like *"Monday"* and *"California"* and rarely, certain person and organizations like *"Judge"* and *"Excite"*. Attention was given in the first pass to the matter of case distinction as there are cases where proper names coincide with normal words like the token *"deal"* as in agreement and the token *"Deal"*, a city in England. A special feature employed in both passes in this category assignment technique is the use of aliases for names to cater new matching possibilities without creating redundancies. Some might refer to *"Hewlett-Packard"* for example, by its name while others recognize the company through its short-form *"HP"*. Either way, this new technique allows entries in gazetteer to refer to the same entity with different names without compromising anything.

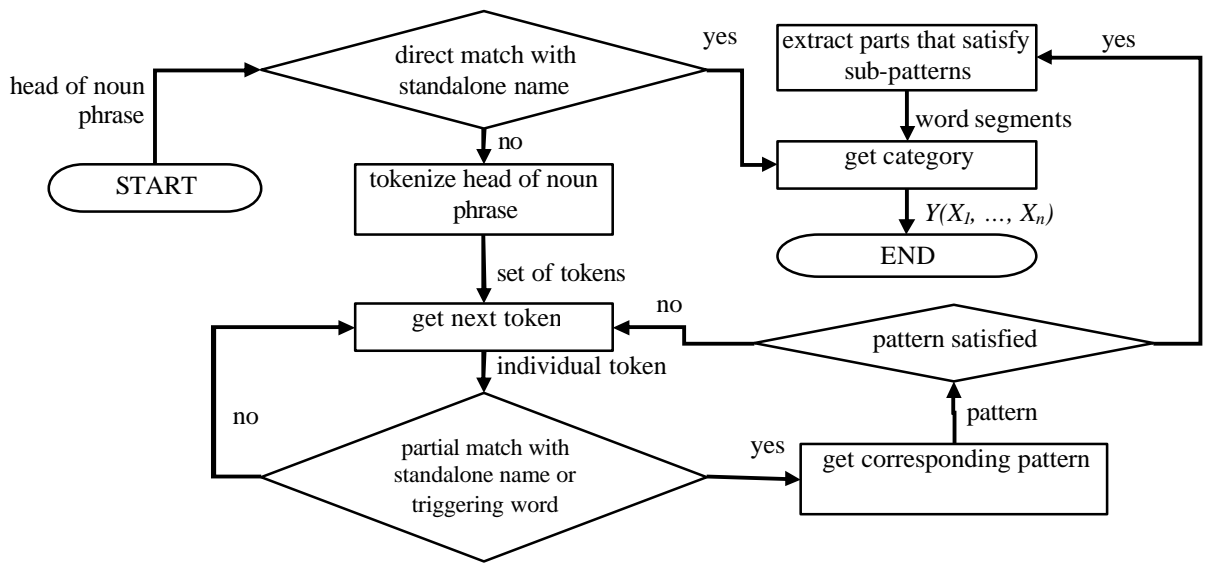

Figure 4.10: Flowchart of the two-pass method

If there are no direct matches for a noun phrase, then the second pass will be executed and usually, names of companies and person like *"Microsoft Corp."* and *"Andrew Garcia"*, consisting of two words will require the second pass for a positive match. Unlike the first pass, the second pass works on both standalone names and triggering words with mandatory fulfillment of corresponding patterns.

Table 4.4: Sample gazetteer entries for company names

| Name | Pattern | Type | Alias |
|---|---|---|---|
| Microsoft | ({TOKEN}(\sCorporation\|\sCorp[\.]?)?) | specific | |
| Excite | ({TOKEN}(\sIncorporated\|\sInc[\.]?)?) | specific | |
| Hewlett-Packard | ({TOKEN}(\sCompany\|\sCo[\.]?)?) | specific | HP |
| Corporation | (([A-Z][\w\d'&\-\.]+\s)+{TOKEN}) | generic | Corp. |

As an example, if the company *"Excite Inc."* appears alone without the *"Inc."* label, the token will have a direct match without any need to proceed to the more complicated pattern match. If the company name appears as *"Excite Inc."*, then there will not be any exact names or aliases for a direct match. For such cases, the token would be broken down in an attempt to achieve any partial matches namely *"Excite"* and *"Inc."*. The first token *"Excite"* would trigger a partial match in the second row of the sample gazetteer in Table 4.4 and the corresponding pattern is retrieved. The *{TOKEN}* variable in the pattern is instantiated with the part *"Excite"* to produce *"Excite(\sIncorporated|\sInc[\.]?)?"* and using a several lines of regular expressions, the original token *"Excite Inc."* is used to match against the instantiated pattern and would produce a positive pattern match. Regular expression was chosen for its strength and expressiveness to handle variations in patterns.

The pattern *(\sIncorporated|\sInc[\.]?)* for example, will enable the recognition of the variants of *"Excite Inc."* like *"Excite Incorporated"* or *"Excite Inc"*[4]. After the positive match, the token would be assigned with the corresponding category of the partial match, *"company"*.

To illustrate the use of aliases and triggering words in the second pass, consider the company name *"Oracle Corp."*. In the second pass, the name would be broken down into *"Oracle"* and *"Corp."*. By referring to Table 4.4 again, the first token *"Oracle"* does not exists, but does this mean that *"Oracle Corp."* cannot be identified as a company. Through the use of triggering word *"Corporation"* and its alias *"Corp."*, the second token *"Corp."* would have a positive partial match and the noun phrase *"Oracle Corp."* will be categorized as a company even though we have no *"Oracle"* in our gazetteer. As for person names, they are identified in the same manner. When a partial match is triggered using the first name, a validating pattern is used to ensure that the trailing last name is in a valid form. Consider the example name *"Andrew Garcia"* where a direct match during the first pass is impossible because we cannot have all possible combinations of first and last names. As before, during the second pass, the name would be tokenized for produce *"Andrew"* and *"Garcia"*. Referring to Table 4.5, the first name *"Andrew"* will trigger a positive partial match and the corresponding pattern *"({TOKEN})\s([A-Z][a-z]+[A-Z\s]\*)"* will be activated and must be fulfilled. The pattern basically states that following the first name, there must be a mandatory last name. All of them must be separated from each other by a white space and must begin with a capital letter and followed by only alphabets and the single quote *"'"*.

Table 4.5: Sample gazetteer entries for person first names

| Name | Pattern | Type | Alias |
|---|---|---|---|
| Andrew | ({TOKEN})\s([A-Z][a-z]+[A-Z\s]*) | generic | |
| David | ({TOKEN})\s([A-Z][a-z]+[A-Z\s]*) | generic | |

---

[4] There is no trailing dot in this variation of Excite Inc.

The phrase *"Andrew Garcia"* satisfies the condition of the pattern and the person's name will be assigned with the category *"person"*. Another important feature about the use of regular expression for pattern matching in the second pass is the ability of the facility to return the extract that satisfies certain sub-patterns. For example, the sub-pattern *"[A-Z][a-z]+"* will return the last name of *"Andrew Garcia"* which is *"Garcia"*. The first and last name can then be used to fill the attributes *"per_fname"* and *"per_lname"* as specified in the ontology. This facility provides additional features to the task of named-entity recognition unlike other techniques.

Another example to illustrate this feature is the phrase *"federal court"*. The substrings *"federal"* and *"federal court"* are returned through the fulfillment of the sub-pattern *"([S|s]upreme\s|[S|s]uperior\s|[F|f]ederal\s|[D|d]istrict\s|[A|a]ppeals\s)"* and the outer most pair of bracket respectively during the regular expression matching using patterns from the *"court"* entry in Table 4.6. The substring can then be used to fill the attribute *"court_type"* obtained from the ontology. Then, the category information assigned to the noun phrase *"Andrew Garcia"* and *"Federal Court"* will be more than just *"person"* and *"court"* respectively.

Table 4.6: Sample gazetteer entries for court

| Name | Pattern | Type | Alias |
|------|---------|------|-------|
| Court | ((([\w\s\.]*) ([S\|s]upreme\s\|[S\|s]uperior\s\|[F\|f]ederal\s\|[D\|d]istrict\s\|[A\|a]ppeals\s)? {TOKEN}) | generic | |

The final output of named-entity recognition contains a category and additional information that will assist in the filling of values of attributes in the form of an n-ary relation $Y(X_1,...,X_n)$ where $X_i$ is a binary relation in the $attr\_pred(x, attr\_value)$ form and $Y$ is the category name assigned to the named-entity. The *"Andrew Garcia"* example will have the final category assigned as *"person(per_fname(X, Andrew),*

*per_lname(X, Garcia))"* and *"federal court"* as *"court( court_type(X, federal), org_name(X, federal court))"*.

In a noun phrase, various modifier nouns can exist, which can be extremely distracting. Due to the feature of our noun phrase chunking that is based on the head-modifier relationship, the head of a noun phrase can be easily identified and thus, allowing only the head to be used for matching in the category assignment phase. To illustrate this, consider the example noun phrase *"U.S. District Judge William Pauley III"*. If the entire phrase were to be used for matching, then a lot of ambiguities and conditions need to be handled. The phrase *"U.S. District Judge William Pauley III"* for example, contains a location identifier *"U.S."* and two person identifiers namely the word *"Judge"* and a person name. This has led to the ambiguity on whether the phrase is referring to a location, a person who is a judge or the person with the name. By exploiting the syntactic information of dependency between words in a phrase, it can be easily pointed out that the phrase is actually referring to a person *"Judge"* and not the country *"U.S."* or some unknown person by the name of *"William Pauley III"*. This subtle but useful information is reflected through the choice of the word *"Judge"* as the head of the phrase in the first place. Thus, by holding to this fact, no matter how the modifiers change from *"U.S."* to *"U.K."* or *"District"* to *"Federal"*, the essence of the phrase is still the same, that is, it points to a judge.

### 4.3.1.3 Relation Inference using Dependency Information

As the task of identifying individuals of entity objects and their attributes are performed by the named-entity recognition component of natural language understanding, instantiating event objects and their attributes are done through relation inference, the final task in semantic analysis. There are four types of relations based on dependency information exploited by relation inference module to identify all possible event objects. The relations are possession, appositive, subject-verb-object and prepositional phrase. This idea of extracting ternary relations using grammatical relationship has been applied by many systems like START as discussed in Chapter 3 but what differs here is how the output is being further utilized.

Possession uses the genitive case, an adjectival form of a noun to show some sort of relationship between itself and what it describes. In a general sense, genitive relationships may be thought of as one thing belonging to, being created from or otherwise deriving from some other thing. Some varieties of possession relations include relationship as in *"Janet's husband"*, subjectivity as in *"my leaving"*, objectivity as in *"the archduke's murder"*, inalienable possession as in *"my height"*, *"his existence"* and *"her long fingers"* and alienable possession as in *"his jacket"* and *"my drink"*. Based on the dependency diagram in Figure 4.11, a possessive relation exists between *"defendant's"* and *"right"* denoted by the *"gen"* link. It can be validly inferred that *"defendant has right"* but not vice versa.

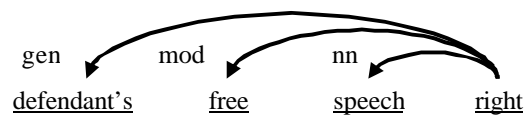

Figure 4.11: Dependency structure of a possessive relation

An appositive is a noun, noun phrase or noun clause which follows a noun or pronoun and renames or describes the noun or pronoun. A simple appositive is an epithet like Alexander the Great. Appositives are often set off by commas. An appositive is denoted through the use of the *"appo"* link and in the example *"Andrew Garcia, a former employee"* in Figure 4.12, an appositive relation can be inferred between *"Andrew Garcia"* and *"a former employee"*.

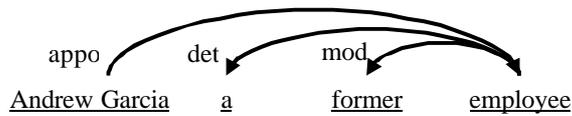

Figure 4.12: Dependency structure of an appositive relation

Next, the most important dependency among words that form the basic structure of an English sentence is subject-verb-object relation, as illustrated in Figure 4.13. The entities and actions encoded in this relation provide the basic information that the complete sentence is trying to deliver. This type of relation is represented in the output of Minipar using the *"s"* and *"obj"* links.

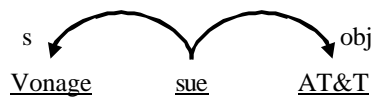

Figure 4.13: Dependency structure of a subject-verb-object relation

In subject-verb-object relation, conjunctions in either subject or object may exist. Please refer to Figure 4.14 to view the example. *"Google"* and *"Electronic Frontier Foundation"* are connected through the *"conj"* link and the subject relation between *"Google"* and *"file"* can be distributed over the conjunctive link to enable us to infer that *"Electronic Frontier Foundation file amicus brief"*.

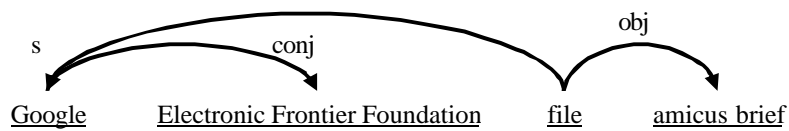

Figure 4.14: Dependency structure of a conjunctive relation

A preposition is a word that establishes a relationship between what is called its object (usually a noun phrase) and some other parts of sentences. The preposition and its object make up a prepositional phrase, which can be used to modify noun phrases and verb phrases in the manner of adjectives and adverbs. In the sentence *"The appeals court rule on Wednesday"* in Figure 4.15 for example, the prepositional phrase *"on Wednesday"* is used to modify the verb *"rule"*. In the sentence *"The appeals court threw out the case against Oracle"*, the prepositional phrase *"against Oracle"* modifies the noun *"the case"*.

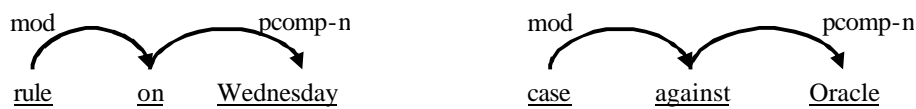

Figure 4.15: Dependency structure of prepositional relation

## 4.3.2 Meaning Unification and Representation using Semantic Network

Meaning extraction produces two important outputs namely entity objects and relations between them from the two sentences of each online news report. Given these outputs, the responsibility of meaning unification and representation is done by discovering event objects using the relations and later using these event objects to encapsulate and relate the entity objects with one another.

Meaning unification and representation is carried out in four steps using the named-entity, verbs and prepositions, patterns and maps for triggering verbs and prepositions and the hierarchical information from the ontology. As this phase involves the integration of meaning from multiple sentences and the need to co-refer between them, the tasks of meaning unification and representation are collectively known as discourse integration, as generally understood in the natural language understanding community. The idea for discourse integration came about from the fact that named-entity and grammatical information is usually underutilized by existing question answering systems. If combined and treated properly, the two pieces of information can yield a more solid representation of meaning for sentences. Moreover, the use of pronoun is very common in English but many question answering systems are not designed to handle it.

**4.3.2.1 Discourse Integration with Anaphora Resolution**

The first step instantiates entity objects and fills their attributes using the information produced by named-entity recognition in the form of $Y(X_1,...,X_n)$. The appropriate class is instantiated based on the category indicator *Y* and the attributes for the new instance can be obtained from the attribute string $X_1,...X_n$. "*company(org_name(X,AT&T))*" for example, can be used to instantiate the class "*company*" to obtain *company(a1)* and fill its attributes using "*org_name(X,AT&T)*" to produce "*org_name(a1,AT&T)*".

The second step involves the use of verbs and prepositions returned by the relation inference phase to trigger possible event classes. These triggering verbs and prepositions together with the related patterns and maps are available in the gazetteer. To illustrate, consider the example verb "*side with*" which will be triggered by the name "*side with*" in the sample gazetteer in Table 4.7. Then, the associated category, pattern and map are returned for use in the next step. The pattern and map are extremely useful not only for plugging in the values of attributes for event objects, but also for performing anaphora resolution.

Table 4.7: Sample gazetteer entries for "*side with*" and "*file against*"

| Name | Category | Pattern | Map |
| --- | --- | --- | --- |
| side with | resolution | {COURT}<RELATION>{PERSON\|ORGANIZATION} | {OCCUR_AT}<RELATION>{PREVAILING_PARTY} |
| file against | legal_proceeding | {PERSON\|ORGANIZATION}<RELATION>{PERSON\|ORGANIZATION} | {PLAINTIFF<RELATION>{DEFENDANT} |

The algorithm resolves the anaphora to the nearest prior named-entity that satisfies the pattern as specified in the gazetteer. Consider the sentence *"A federal court has sided with AT_T over a complex patent lawsuit it filed against Microsoft."* and its corresponding list of named-entities below in Table 4.8. Initially, the anaphora *"it"* and other unknown entities are tagged as *"variable"*. The left column contains the offset of the entities in terms of discourse offset and local offset and *"it"* occurs at discourse offset 0 and local offset 13. The theory adopted by the anaphora resolver is to look back for the first named-entity whose class matches that of the pattern. In the case of anaphora *"it"*, the candidate antecedents are located in offsets before *"0.13"*, that is *"variable(desc(X,complex patent lawsuit))"*, *"company(org_desc(X,_), org_name(X,AT&T))"* and *"court(court_type(X,_),org_desc(X,federal ),org_name(X,federal court))"*.

Table 4.8: Sample named-entities and offsets information for anaphora resolution

| | |
|---|---|
| 0.12 | variable(desc(X,complex patent lawsuit)) |
| 0.13 | variable(desc(X,it)) |
| 0.16 | company(org_desc(X,_),org_name(X,Microsoft)) |
| 0.3 | court(court_type(X,_),org_desc(X,federal ),org_name(X,federal court)) |
| 0.7 | company(org_desc(X,_),org_name(X,AT&T)) |
| 1.14 | variable(desc(X,evidence)) |
| 1.18 | variable(desc(X,AT&T concealed information)) |
| 1.23 | government(org_desc(X,_),org_name(X,U.S. Patent Office)) |
| 1.24 | variable(desc(X,when)) |
| 1.29 | variable(desc(X,its speech compression)) |
| 1.3 | judge(per_fname(X, William),per_lname(X, Pauley III),profession(X,Judge)) |
| 1.9 | date(day_of_month(X,9),day_of_week(X,Monday),month(X,February),year(X,2004)) |

Using offset information alone is not enough as this would mean *"variable(desc(X,complex patent lawsuit))"* is the antecedent. This light-weight anaphora resolution also takes into consideration the context in which the anaphora exists in a simple subject-verb-object relation or prepositional phrase. In the case of *"it"*, it exists in a subject-verb-object relation *"it file against Microsoft"* where the position *"it"* assumes must be an active performer of some task specified through the verb. This constraint is duly specified in the pattern associated with each verb and by the referring back to the sample entries of the gazetteer above, the pattern associated with *"file against"* states that the subject *"it"* must assume the role of a person or organization. This constraint eliminates the two candidates *"variable(desc(X,complex patent lawsuit))"* and *"court(court_type(X,_),org_desc(X,federal ),org_name(X,federal court))"*, leaving the one possible antecedent for *"it"*, *"company(org_desc(X,_),org_name(X,AT&T))"*.

The fourth and final step in discourse analysis is the instantiation of event objects and filling their attributes with entity objects. Initially, the verbs or prepositions that trigger event categories in the second step are used to instantiate event objects. *"file against"* for example, triggers the *"legal_proceeding"* category and thus a new instance *legal_proceeding(e1)* is created. Due to the fact that many different verbs in the same discourse can trigger similar events, such verbs will all point to the same event object instead of creating multiple objects of the same event class. This is followed by the use of maps to relate entity objects to attributes of the newly created event objects. Using the same example, *"{federal court}<sided with>{AT&T}"* employs the map *"{OCCUR_AT}<RELATION>{PREVAILING_PARTY}"* to fill two attributes namely *"occur_at"* and *"prevailing_party"* with entity object *company(e2)* for AT&T and *company(e3)* for Microsoft respectively.

The final output is in logic form and will be integrated into the existing semantic network. The final output for the phrase *"AT&T file against Microsoft"* is shown in Table 4.9 and the corresponding semantic network is depicted in Figure 4.16 below. Objects like *"e2"*, *"e3"* and *"e1"* are by default related to their class *"company"* for the first two and *"legal_proceeding"* for the latter using the edge *"is"*. As for attributes like *org_name(e3,Microsoft), org_name(e2,AT&T), defendant(e1,e3)* and *plaintiff(e1,e2)*, their predicates are used as edges to connect the first argument to the second argument.

Table 4.9: Logical representations of event and entity objects

| legal_proceeding(e1) | | | |
|---|---|---|---|
| **attributes of event objects** | *defendant(e1,e3)* | *company(e3)* | |
| | **entity objects** | **attributes of entity objects** | *org_name(e3,Microsoft)* |
| | *plaintiff(e1,e2)* | *company(e2)* | |
| | **entity objects** | **attributes of entity objects** | *org_name(e2,AT&T)* |

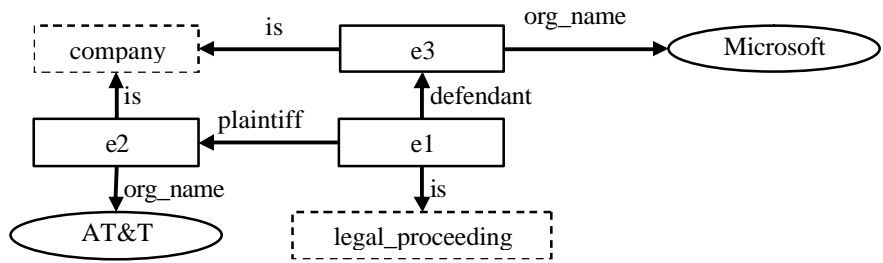

Figure 4.16: Semantic network for *"AT&T file against Microsoft"*

## 4.4 Reasoning Mechanisms

The reasoning mechanism in NaLURI couples the novel idea of complexity reduction during answer discovery in a network-oriented knowledge base with two advanced reasoning features namely relaxation of event constraint and explanation on failure to provide higher standards of responses, way beyond the current conventional factual answers. This idea of a network-based answer discovery came about through the motivation that the power of semantic network cannot be fully tapped through basic inferences and the reasoning in semantic network has yet to incorporate advanced features. Such reasoning mechanisms and the ontological commitment of the knowledge base can be said as a match made in heaven. This statement is well-justified because such advanced reasoning cannot be carried out without the use of domain ontology and knowledge base and only with the adoption of these high-level reasoning capabilities can the ontological information and knowledge base be thoroughly exploited.

The first two parts in this section discusses how the complexity of discovering for answers in network-oriented knowledge base can be handled in terms of complexity reducibility through a series of form reduction. The third part presents the algorithmic ideas for advanced reasoning capabilities that are implemented as part of the answer discovery mechanism. The final part discusses how valid answers are merged and presented in a unified unambiguous natural language response.

### 4.4.1 Query Network

Reasoning in NaLURI works on two types of logical network namely the semantic network and query network, consisting of the networked representation of the meaning of online news and of the question respectively. In essence, the query network is similar to the semantic network as appears in the knowledge base and the derivation of both networks uses the same approach of natural language understanding as described in the previous section. The only difference during the construction of a query network is the use of potential answer marker *X*. The marker can appear only in leaf nodes in the query network. In other words, only attributes of entity objects can be returned as answers.

The placement of the marker is carried out during the discourse integration phase of the natural language understanding process. The placement can either occur naturally on subjects and objects of verbs or explicitly placed for noun triggered events and in the absence of any marker. In the first case, subjects and objects of a verb can be missing in a dependency tree of a question, consider the example of two simple questions, *"Who sues Microsoft?"* and *"Microsoft sues whom?"*. The corresponding dependency trees are shown in Figure 4.17.

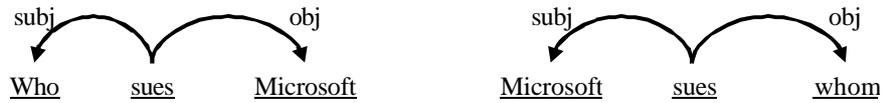

Figure 4.17: Dependency structure of questions

The verb *"sue"* would give rise to a *"legal_proceeding"* event and its associated pattern and map as shown below in Table 4.10. As the wh-word assumes the position of a subject in the first question and object in the second question, the answer marker is placed in the attributes of the event that are mapped to those positions namely *"plaintiff"* in the former and *"defendant"* in the latter.

Table 4.10: Sample gazetteer entries for *"sue"*, *"file on"* and *"filing"*

| Name | Category | Pattern | Map |
| --- | --- | --- | --- |
| sue | legal_proceeding | {LEGAL_ENTITY}<RELATION>{LEGAL_ENTITY} | {PLAINTIFF}<RELATION>{DEFENDANT} |
| file on | filing | {VARIABLE}<RELATION>{DATE} | {}<RELATION>{OCCUR_ON} |
| filing | filing | no pattern | no map |

In the next case, markers are placed using the help of wh-words as a contingency step in the absence of any markers. Consider the question *"When was the filing of the case against Excite by Microsoft?"* where the noun *"filing"* would give rise to the *"filing"* event. Given the event category *"filing"* and the entity category *"date"* indicated by the wh-word *"when"*, NaLURI searches for trigger words in gazetteer that belong to the event category and consist of the entity category in its pattern. Each trigger word in the gazetteer has an associated map for assigning the matching values to their appropriate attributes and using this facility, the corresponding attribute in the map for the matching entity category is assigned with the answer marker *X* as its value. By referring to the sample gazetteer entries above, the trigger word *"file on"* satisfies the requirement of being in the same category as the noun *"filing"* and having *"date"* in its pattern is *"file on"*. Given that, the corresponding map for the sub pattern *"date"*, which is the attribute *"occur_on"* will be assigned the answer marker *X* as its value.

The network shown below in Figure 4.18 is a complete example of the query network for the question *"When did AT&T file its case against Microsoft?"*. The understanding module has placed an answer marker *X* in the date object *"a039"* using the contingency step. The question is analysed in steps and shown as follows. Firstly, the verb *"file"* has the subject *"AT&T"* and the object *"its case"*. Based on the sample entries below in Table 5.11, the verb gives rise to the event *"filing"* and both the subject and object satisfy the pattern, resulting in entity object *"AT&T"* being assigned to the event attribute *"plaintiff"*. This *"plaintiff"* relation is represented as an edge between nodes *"6360"* and *"1b1c0"*. As for the preposition *"against"*, it triggers the general event *"legal_proceeding"* and its direct entity object *"Microsoft"* is assumed as value for the event attribute *"defendant"*, represented as an edge connecting nodes *"bf99"* and *"1b1c0"*.

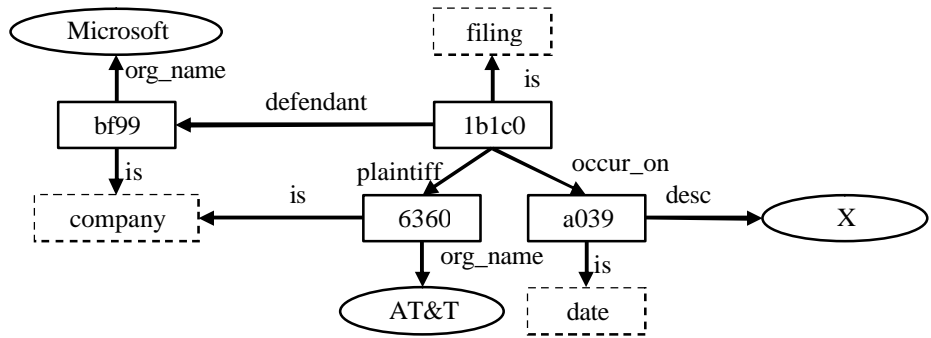

Figure 4.18: Query network for *"When did AT&T file its case against Microsoft?"*

At this stage, no more grammatical relationships can be exploited and the answer marker has not been assigned. The contingency step is executed whereby the gazetteer is explored to find trigger words belonging to the same category as the previous events namely *"filing"* and *"legal_proceeding"* and having the entity category for the wh-word *"when"*, which is *"date"*, in its pattern. Based on the sample gazetteer entries below in Table 4.11, the trigger word *"file on"* for event *"filing"* and *"occur on"* for event *"legal_proceeding"* are found. Consequently, the answer marker *X* is assigned to the map, *"occur_on"* of the corresponding sub-pattern *"date"* for each trigger word.

Table 4.11: Sample gazetteer entries for *"file"*, *"against"*, *"file on"* and *"occur on"*

| Name | Category | Pattern | Map |
|---|---|---|---|
| file | filing | {LEGAL_ENTITY}<RELATION>{VARIABLE} | {PLAINTIFF}<RELATION>{} |
| against | legal_proceeding | {FILING\|VARIABLE\|LEGAL_ENTITY}<RELATION>{LEGAL_ENTITY} | {}<RELATION>{DEFENDANT} |
| file on | filing | {VARIABLE}<RELATION>{DATE} | {}<RELATION>{OCCUR_ON} |
| occur on | legal_proceeding | {VARIABLE}<RELATION>{DATE} | {}<RELATION>{OCCUR_ON} |

## 4.4.2 Answer Discovery using Selective Network Path Matching

After obtaining the query network, the task of answering the question is reduced to discovering the presence of the query network in the whole of semantic network. To perform the discovery, it is necessary to first understand the notion of the three types of nodes in a query or semantic network. A root node is a class node where one or more intermediate nodes are being created and thus, in the network, a root node can only have incoming edges, usually more than one. The second type of node is intermediate node where edges can be from both directions to interconnect other intermediate nodes. The last type of node is the leaf node. The leaf node is similar to root node in the sense that both can only have incoming edges and beyond that, the similarity ends. In the leaf node, there can only be one incoming edge.

To perform the discovery, query and semantic network are collapsed into two sets of all possible paths $Q$ and $S$ respectively. Each set of paths takes the form of $\{P_1,...,P_n\}$ where $P_i$ is a sequence of alternating node and edge in the form $n_{i1}, e_{i1}, n_{i2}, e_{i2}, n_{i3}, e_{i3}, n_{i4}$ that satisfy the rules:

- each path sequence must begins with a leaf node and ends with a root node;
- each path sequence must contains exactly two intermediate nodes; and
- for query network, the path beginning with the leaf node $X$ must not be included in $Q$.

Another set called $A$, consisting of only one element, which is the path in the query network that begins with the leaf node $X$ is introduced. Hence, it can be concluded that the answer may exists and retrievable from the semantic network if:

(1) for each $q_i \in Q$, there exists an $s_j \in S$ such that $q_i$ conditionally matches $s_j$ where the condition of the matching between elements of $Q$ and $S$ implies a literal match between member $n_1, e_1, e_2, e_3$ and $n_4$ of path sequence $q_i$ and of $s_j$.

The problem of discovering the answer has been reduced to finding the path $s_j \in S$ such that:

(2) every nodes and edges except $n_1$, $e_1$, $n_2$ and $n_3$ in $s_j$ matches those in the single element of $A$; and

(3) the node $n_3$ in $s_j$ from (2) must match the $n_3$ of any $s_j$ that conditionally matches $q_i$ in (1).

Consider the sample query network in Figure 4.18 for the question *"When did AT&T file its case against Microsoft?"*. The entire network is collapsed into the set $Q$ below:

$Q = \{$ *"Microsoft, org_name, bf99, defendant, 1b1c0, is, filing",*

*"AT&T, org_name, 6360, plaintiff, 1b1c0, is, filing"* $\}$

$A = \{$ *"X, desc, a039, occur_on, 1b1c0, is, filing"* $\}$

Based on a segment from the knowledge base in Figure 4.19, the set *S* is obtained:

*S* = {   *"Microsoft, org_name, bf99, defendant, 1b1c0, is, filing",*

   *"AT&T, org_name, 6360, plaintiff, 1b1c0, is, filing"*

   *"2002, year, a039, occur_on, 1b1c0, is, filing"*

   *"federal court, org_name, b7, occur_at, 1b1c0, is, filing"*

   *"federal, court_type, b7, occur_at, 1b1c0, is, filing"*        }

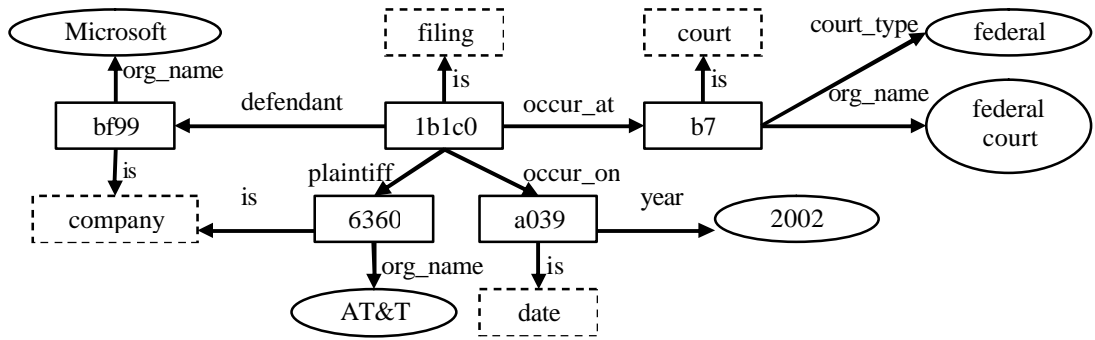

Figure 4.19: A portion of the semantic network from knowledge base

By selectively matching the members of path sequence $q_i$ and $s_j$, all paths in *Q* are satisfied. In the next step, the single element of *A* will be matched against elements of *S* so that the two conditions are met. Lastly, the leaf node *X* of the path sequence in *A* is assigned with the value of the corresponding member $n_{m1}$ of the matching $s_m$. In our case, *X* would be assigned with *"2002"*.

### 4.4.3 Advanced Reasoning Features

The two algorithmic ideas implemented during answer discovery to provide better answers over conventional techniques are discussed in this last section of the reasoning mechanism. The concept of advanced reasoning was initially implemented in an indexed-text first-order logic environment by WEBCOOP as discussed in Chapter 3. Even though not new, our version of advanced reasoning was designed to work in a more powerful environment namely semantic network.

Firstly, relaxation of the event constraint is done to enable the loosening of scope for questions that covers more than one potential event. Consider the example question *"Who presided the case against Microsoft?"* and its corresponding path set *Q* below.

$$Q = \{ \text{ "Microsoft, org\_name, bf99, defendant, 1b1cc, is, legal\_proceeding" } \}$$

The root node *"legal_proceeding"* at the end of the sequence is actually a superclass node that is capable of having many others subclasses namely *"filing"*, *"resolution"* and etc. As the focus of the question is wide, at the level of superclass event, the chances of finding the desired answer is very slim as meanings of news are stored in their most detailed level in the semantic network to leave no room for ambiguities. Knowing this, the literal matching of $n_4$ during the selective path matching is extended so that parent-child relationship is put into consideration. During the matching of $q_i \in Q$ and $s_j \in S$, the node $n_4$ of path $q_i$ is relaxed to enable the subclasses of $n_4$ to pass through. This is to say that any of the below sample path $s_j$ from the semantic network can have a positive match with $q_i$.

$$s_1 = \textit{Microsoft, org\_name, bf99, defendant, b2, is, resolution}$$

$$s_2 = \textit{Microsoft, org\_name, bf99, defendant, b2, is, appeal}$$

$$s_3 = \textit{Microsoft, org\_name, bf99, defendant, b2, is, filing}$$

Secondly, explanations on failures are provided in the absence of any valid answers to clear doubts concerning the status of the knowledge base. Consider the example question *"Which judge presided the ruling of the case by RealNetworks against Microsoft"*. A conventional question answering system would easily answer *"No"* in the absence of any valid answers. Such answers will definitely leave the users with a big question mark concerning the actual implication. The answer *"No"* can have two implications here. First, the system responses *"No"* because of the inexistence of such case between RealNetworks and Microsoft and second, the system can also respond *"No"* because of the absence of information about the judge presiding the closing of the case in its knowledge base. To provide an explanation on why no valid answers were produced, the following conditions are used:

- if there exist at least one $q_i$ in $Q$ that fails to match some $s_j$ in $S$, then this means the event in question does not exist. Using the example question above, we can say that the ruling of case between RealNetworks and Microsoft does not exists; and

- if all $q_i$ in $Q$ holds for some $s_j$ in $S$ and the single element in $A$ fails to find a match in $S$, then this means the event in question exists but the semantic network has inadequate knowledge concerning the event.

### 4.4.4 Template-based Response Generation

In the case of positive answer discovery, responses must be generated in a way that is readable and unambiguous to the users. Each question actually queries one of the many event attributes in the semantic network as denoted by the marker $X$ in the query network. Answer generation exploit this fact and uses a template-based approach where each event attribute is attached to an answer generation guide. For example, the events attribute *"preside_by"* can be attached to the template *"<COURT> judge <JUDGE> chairs the <EVENT> of the case by <PLAINTIFF> against <DEFENDANT> on <DATE>"* in the ontology. Then, the generation of the response for the question querying the attribute *"preside_by"* can be easily done through the instantiation of the associated template in the ontology using the rest of the attributes of the event in the semantic network that contains the value for the marker *X*.

## 4.5 Summary

In this chapter, we have presented a solution framework that abides by the criteria in the hypothesis. This framework constitutes some innovative use of existing practices that were discussed in Chapter 2 and Chapter 3, and also several new algorithms. Table 4.12 summarizes and briefly describes the level of novelty, and the origin or motivation behind each component that are included in the solution framework.

Table 4.12: Novelty of components in solution framework

| components | | | description | novelty | motivation/origin |
|---|---|---|---|---|---|
| natural language understanding | syntax analysis | sentence parsing | employs an external module | common | Minipar (Lin, 1998) |
| | semantic analysis | noun phrase chunking using context-free grammar | the use of context-free grammar for noun-phrase chunking is common | common | Abney (1991) Abney (1995) Chen & Chen (1993) |
| | | named-entity recognition / category assignment using two-pass matching technique | innovatively combine existing list lookup, triggering approach or carefully hand-crafted pattern | innovation | Maynard (2003) Gaizauskas et al. (1995) Appelt *et al.* (1993) |
| | | relation inference / exploit dependency information between words | the use of grammatical relationship between words to generate relation is common | common | Lin (2001) START (Katz, 1997) |
| | discourse analysis | four stage discourse integration with light-weight anaphora resolution | integrating meaning from different sentences of the same discourse plus the ability to resolve anaphora is new | new | • named-entity and grammatical information is usually underutilized. • if combined and treated properly, the two pieces of information can yield a better meaning representation. • many question answering systems are not designed to handle pronoun. |
| reasoning | | network-to-path reduction | the use of network-based technique to discover for answers in a semantic network is new | new | • power of semantic network cannot be fully tapped through basic inferences • the reasoning in semantic network has yet to incorporate advanced features • existing advanced features in Benamara & Saint-Dizier (2003) |
| | | selective path matching | | | |
| | | advanced reasoning features | innovatively adapt advanced reasoning features in a semantic network environment | innovation | |
| | | template-based response generation | the use of template for generating answers in common | common | |

# Chapter 5

# Architecture of NaLURI

## 5.1 Introduction

NaLURI is a prototype system that implements the solution framework discussed in the Chapter 4. All algorithms and data structures discussed in the previous chapter are realized using Perl language with relational database MySQL. The reasons behind the choice of the programming language, database software and other software modules used for the development of NaLURI will also be discussed and presented. The interactions between the interrelated modules and subsystems are also highlighted to demonstrate the consistency and pipelined communications that has become one of the main reasons behind the high performance of NaLURI. Finally, the storage structures that are widely used throughout the system and their unique features that make them work well with the modules will be presented in the last section.

## 5.2 Tools for Development

Perl[1] was chosen for development of the prototype for its strength in handling text, data structures and its stability, which is reflected through the nickname *"C language for the World Wide Web"* provided by the web programming community for Perl. The regular expression capability of Perl, which is heavily utilized during natural language processing, is also a determining factor. Moreover, Perl has a wide range of modules in various fields, especially artificial intelligence contributed by individuals readily available for

---

[1] The binary distribution is available via http://www.activestate.com/Products/ActivePerl/. Last retrieved on 25 March 2005.

download[2]. As the intention is to make NaLURI accessible on the World Wide Web, Perl has the added advantage over other standalone programming languages. Besides, Perl is widely deployed and because of common high-level interfaces, Perl code written on one platform is often more portable to other platforms than the equivalent C code would be. Perl can also be embedded within applications written in other languages to allow for maximum reusability of the framework

As for the storage facility, the relational database MySQL is used for its open-source nature and the strength it has to offer in a small-scale non-commercial environment. It is vital to note that the choice of MySQL must not be debated over other proprietary databases like Microsoft SQL Server, which are meant for large-scale high volume commercial usage, due to the lack of features like replication services, load balancing, levels of locking supported and so on. All of these only make MySQL unacceptable in a large scale commercial environment but despite these shortcomings, MySQL did not pose any difficulty during development. Other external module used for the development of NaLURI is Minipar, a broad-coverage parser for the English language.

## 5.3 System Modules

The functionalities of NaLURI are nicely packed into two main subsystems namely natural language understanding and network-based advanced reasoning as shown in the Figure 5.1. There are three groups of storage structures namely news repository, knowledge base which consists of ontology and semantic network, and gazetteer. Collectively, the system interacts with the environment in three forms namely reading and understanding sentences from Cyberlaw news articles that have been processed to populate the semantic network, receiving question from user, and providing response to user.

---

[2] The modules collection are accessible via http://www.cpan.org

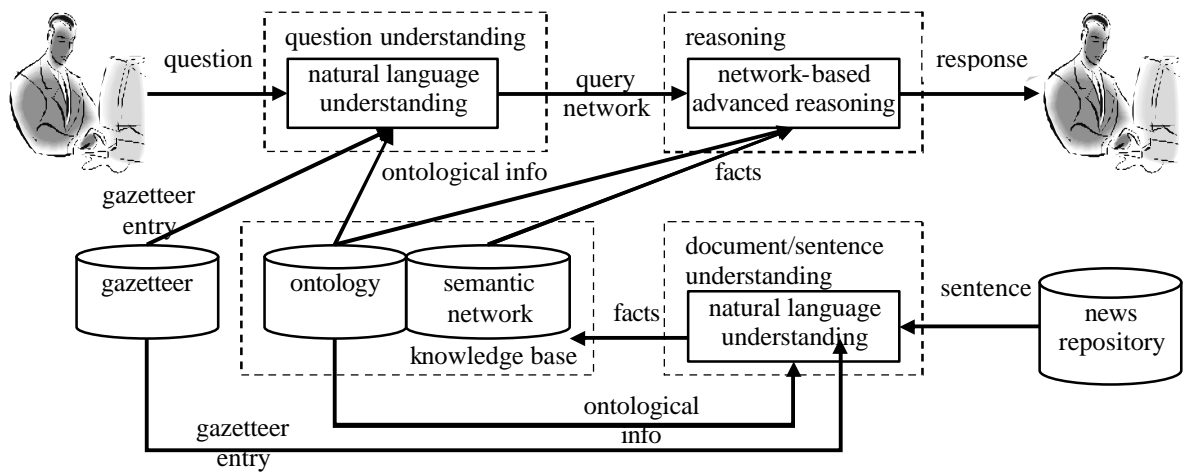

Figure 5.1: Architecture of NaLURI

## 5.3.1 Natural Language Understanding Subsystem

Natural language understanding implements the natural language understanding mechanism discussed in Chapter 4. This subsystem is responsible to read and understand two things: questions from users and sentences of processed Cyberlaw news articles from news repository. In more specific terms, this subsystem is to turn the English representation of the meaning of questions or news content into network representation. The process is carried out in four phases by four natural language processing modules namely sentence parsing, named-entity recognition, relation inference and discourse integration, as shown in Figure 5.2. The input to this subsystem must be in the form of a single English sentence where each can either be the individual sentences of the news content from news repository or the question from user. The output of this subsystem takes the form of a network-bound meaning representation of all the sentences in a discourse.

The sentence parsing module takes a single English sentence and using Minipar, produces the grammatical categories and relations of the sentence. These two types of information are then fed into named-entity recognition and relation inference. Named-entity recognition module implements both the noun phrase chunking and category assignment algorithm discussed in the previous chapter. This module first performs noun phrase chunking and later assigns a category out of the many predefined based on advice from the gazetteer and the ontology. The tagged noun-phrases or named-entities are then passed on to relation inference and together with the output of sentence parsing, relation inference will deduce and extract useful grammatical relations. The output of named-entity recognition and relation inference for different sentences of the same discourse are gathered and used by the discourse integration module to generate a network representation of the meaning of the discourse. This discourse integration module is the one that implements the four-stage approach with light-weight anaphora resolution.

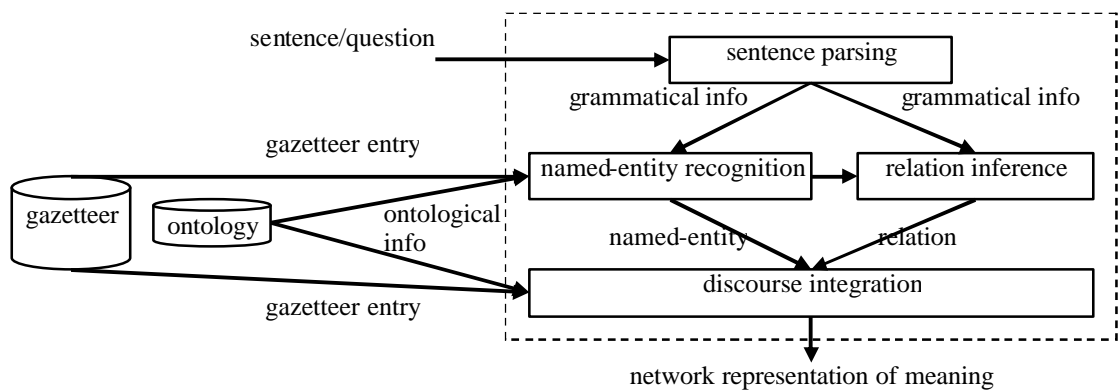

Figure 5.2: Architecture of natural language understanding subsystem

### 5.3.2 Network-based Advanced Reasoning Subsystem

The network-based advanced reasoning subsystem implements ideas in the reasoning mechanisms of the solution framework discussed in Chapter 4. Network-based advanced reasoning subsystem is responsible to discover the valid answer and generate unambiguous response or generate an explanation for users' questions. The process is executed in five phases by five modules namely network-to-path reduction, selective path matching, relaxation of event constraint, explanation on failure and template-based response generation. The modules are illustrated in Figure 5.3.

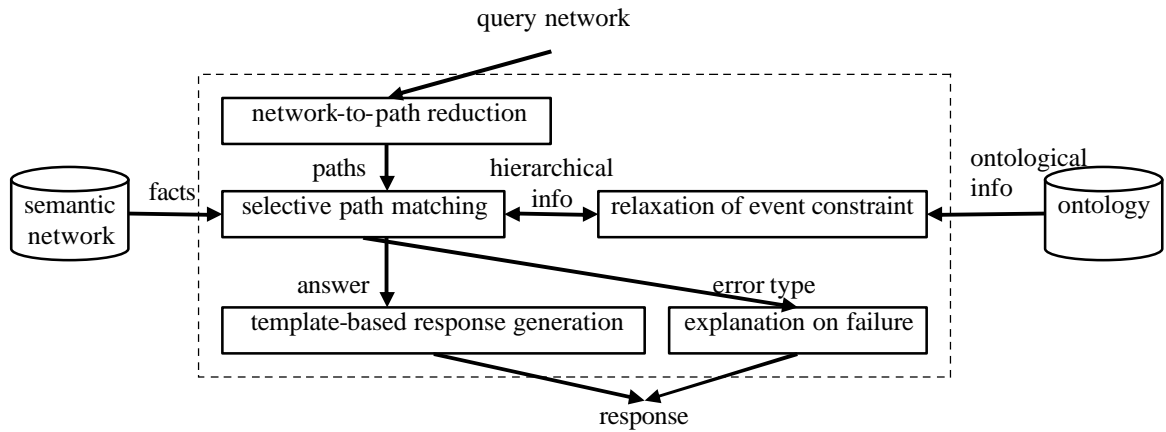

Figure 5.3: Architecture of network-based advanced reasoning subsystem

Network-to-path reduction module collapses the query network into sets of path sequences to reduce the complexity in discovering the answer. The output of network-to-path reduction is two sets of path sequences that will be used by selective path matching module to discover the answer from the semantic network through a series of conditional path unification. To extend beyond literal matching of path sequence, ontological information is utilized to put into consideration events that are hierarchically equivalent. This process is performed by the module relaxation of event constraint. In case of failure during the discovery for a valid answer by selective path matching, an explanation or justification is dynamically generated by the explanation on failure module as an alternative response. This process is carried out based on the context of the question and the current status of the semantic network. If answers can be validly discovered, then readable and unambiguous natural language responses are generated by the template-based response generation module.

## 5.4 Storage Structures

All storage structures are implemented as relational tables using the MySQL database instead of indexed texts or other forms. This is to ensure a uniform interface for accessing the data and quick data read/write.

### 5.4.1 News Repository

The news repository holds all the meta-information about the news content from online news portal like ZDNet. The collection of meta-information consists of identification string, URL, title, author, date, content and category as shown in Figure 5.4. The manner through which the news contents are obtained and maintained is beyond the scope of this research. More information is available through the work of Wong *et al.* (2004a).

| meta_id | meta_url | meta_title | meta_day | meta_month | meta_year | meta_conte | meta_categ | meta_root | meta_status | meta_timest |
|---|---|---|---|---|---|---|---|---|---|---|
| 2619a560ec0 | http://zdnel | HP declares | 16 | January | 2004 | <paragraph: | earning | f988759b7C | set | 000000090! |
| 0ff770eec3f0 | http://zdnel | Red Hat rel | 16 | January | 2004 | <paragraph: | software | f988759b7C | set | 000000090! |
| 9a4f540bb53 | http://zdnel | Cast Iron Sy | 23 | January | 2004 | <paragraph: | acquisition | f988759b7C | set | 000000090! |
| 3a0da7f197f8 | http://zdnel | Former trade | 22 | January | 2004 | <paragraph: | employment | f988759b7C | set | 000000090! |
| 3b2f345ca32l | http://zdnel | Multimedia ( | 15 | January | 2004 | <paragraph: | hardware | f988759b7C | set | 000000090! |
| f508677d9e8l | http://zdnel | J.R.R. Tolki | 14 | January | 2004 | <paragraph: | cyberlaw | f988759b7C | set | 000000090! |
| 8d59a6a8501 | http://zdnel | Intel funds E | 26 | January | 2004 | <paragraph: | hardware | f988759b7C | set | 000000090! |
| d043be12ce9 | http://zdnel | Boingo roan | 26 | January | 2004 | <paragraph: | wireless-tec | f988759b7C | set | 000000090! |
| 7ba80f35269 | http://zdnel | Web ad sal | 16 | January | 2004 | <paragraph: | business | f988759b7C | set | 000000090! |
| 41581dede29 | http://zdnel | Top sales e | 14 | January | 2004 | <paragraph: | employment | f988759b7C | set | 000000090! |
| 0642ae563aa | http://zdnel | LookSmart ( | 21 | January | 2004 | <paragraph: | managemer | f988759b7C | set | 000000090! |
| 6f33059a85a! | http://zdnel | Wavecom l | 23 | January | 2004 | <paragraph: | business | f988759b7C | set | 000000090! |

Figure 5.4: Screenshot of the news repository implementation

### 5.4.2 Semantic Network

The semantic network contains the meaning of the news content from news repository that is produced by the natural language understanding subsystem. The meaning is represented as a network in the form of a collection of binary relations. The functor of the relation will form the edge in the network and both the arguments will be the adjoining nodes. The screenshot of the implementation of the semantic network is shown in Figure 5.5.

| sn_node1 | sn_edge | sn_node2 |
|---|---|---|
| 1721dfc1fb8e7a90377a5b49b1bb35893 | is | resolution |
| 1721dfc1fb8e7a90377a5b49b1bb35893 | prevailing_party | 6360a84768ba80879c4e29b9ff643d4d |
| 6360a84768ba80879c4e29b9ff643d4d | is | company |
| 6360a84768ba80879c4e29b9ff643d4d | org_name | AT&T |
| 1721dfc1fb8e7a90377a5b49b1bb35893 | nature_of_case | dc0821fd329ae3f4368c60b140747cbc |
| dc0821fd329ae3f4368c60b140747cbc | is | variable |
| dc0821fd329ae3f4368c60b140747cbc | var_desc | complex patent lawsuit |
| 1721dfc1fb8e7a90377a5b49b1bb35893 | defendant | bf99a71cb5cc77b2f65088e1692f75c0 |

Figure 5.5: Screenshot of the semantic network implementation

### 5.4.3 Ontology

Ontology consists of two types of entries that collectively define the ontology of NaLURI namely class entry and attribute entry. The entire class entries collection defines the hierarchy of the ontology. As shown in the following screenshot Figure 5.6, the class entries have the *"onto_type"* field marked as *"class"* and the attribute entries marked as *"attribute"*. The class hierarchy begins with the invisible parent node *{START}* and followed by *"domain"* and so on. A special field worth noting is *"onto_response"* which is a response template attached only to event attributes for template-based response generation.

| onto_id | onto_parent | onto_level | onto_type | onto_ | onto_response |
|---|---|---|---|---|---|
| legal_proceeding | occur_on(X,date(_)) | 3 | attribute | 0 | <EVENT> took place on <ANSWER> w |
| location | city(X,_) | 4 | attribute | 1 | |
| location | state(X,_) | 4 | attribute | 2 | |
| location | country(X,_) | 4 | attribute | 3 | |
| location | region(X,_) | 4 | attribute | 4 | |
| date | day_of_week(X,_) | 4 | attribute | 1 | |
| date | day_of_month(X,_) | 4 | attribute | 2 | |
| date | month(X,_) | 4 | attribute | 3 | |
| date | year(X,_) | 4 | attribute | 4 | |
| court | court_type(X,_) | 6 | attribute | 2 | |
| person | per_lname(X,_) | 4 | attribute | 3 | |
| variable | var_desc(X,_) | 4 | attribute | 1 | |
| resolution | losing_party(X, person) | 4 | attribute | 0 | |
| domain | {START} | 1 | class | 0 | |
| entity | domain | 2 | class | 0 | |

Figure 5.6: Screenshot of the ontology implementation

## 5.4.4 Gazetteer

The gazetteer is a huge list structure containing trigger words for use in named-entity recognition and discourse integration, and can also be seen as the backbone for many other natural language understanding tasks. Each entry in the gazetteer consists of the word (i.e. noun, verb or preposition), category of the word with respect to the ontology, mandatory pattern that must be satisfied, nature of the word, an alias and a map. Figure 5.7 shows the segment of gazetteer for use during discourse integration. These trigger words are used to give rise to events and the corresponding maps are used to fill the attributes of the events. As denoted by the *"g_type"* field, the trigger word for use in discourse integration can be verb, preposition and noun.

| g_name | g_category | g_pattern | g_type | g_alia | g_map | g_id |
|---|---|---|---|---|---|---|
| filing | filing | no pattern | relation->noun | | | eb51e2297 |
| trial | trial | no pattern | relation->noun | | | 593fa6e234 |
| closing | resolution | no pattern | relation->noun | | | 329a47e25 |
| ruling | resolution | no pattern | relation->noun | | | c0cab01dd |
| in | legal_proceeding | {LEGAL_EN | relation->prep | | {}RELATION{NATURE_OF_CASE} | 38565a4c3 |
| over | legal_proceeding | {LEGAL_EN | relation->prep | | {}RELATION{NATURE_OF_CASE} | dea3bc1e2 |
| with | legal_proceeding | {LEGAL_EN | relation->prep | | {}RELATION{PLAINTIFF&&DEFENDAN | 0ce2a356e |
| against | legal_proceeding | {FILING|VAF | relation->prep | | {}RELATION{DEFENDANT} | 3eb75acf72 |
| by | legal_proceeding | {VARIABLE| | relation->prep | | {}RELATION{PLAINTIFF} | b75a8a000 |
| sentence | resolution | {JUDGE|CO| | relation->verb | | {PRESIDE_BY|OCCUR_AT}RELATION | 64fe48fa7a |
| file against | legal_proceeding | {LEGAL_EN | relation->verb | | {PLAINTIFF}RELATION{DEFENDANT} | d7c5adc65 |
| rule on | resolution | {JUDGE|CO| | relation->verb | | {PRESIDE_BY|OCCUR_AT}RELATION | b8d77ed26 |
| initiate | legal_proceeding | {LEGAL_EN | relation->verb | | {PLAINTIFF}RELATION{} | 117fb664ee |
| side with | resolution | {JUDGE|CO| | relation->verb | | {PRESIDE_BY|OCCUR_AT}RELATION | bf4b109656 |

Figure 5.7: The portion of the gazetteer for use in discourse integration

This last screenshot in Figure 5.8 shows the portion of the gazetteer for use during named-entity recognition. Unlike the portion of the gazetteer for discourse integration, this portion contains only trigger words of type noun. The *"g_type"* field specifies whether the noun is a generic trigger word like first name or last name of a person or an actual specific standalone name like *"Malaysia"*.

| g_name | g_category | g_pattern | g_type | g_alias | g_map | g_id |
|---|---|---|---|---|---|---|
| Association | ngo | ([\w\d'&\-\.\s]+{TOKEN}[\w\s\.]* | generic | association | 1,2 | a1b0d709fcb0603b9 |
| Marilyn | person | ({TOKEN})\s([A-Z][a-z]+[A-Z\s]*) | generic | | 1,2,3,4 | b84dc6f02dcbda483 |
| Fujitsu | company | ([\w\s]*)({TOKEN}(\sLimited\sLtd | specific | | 2,1 | fde02a44673244fd4l |
| International Busin | company | ([\w\s]*)({TOKEN}(\sCorporation\ | specific | IBM | 2,1 | ae257a7f8f84f9914f |
| William | person | ({TOKEN})\s([A-Z][a-z]+[A-Z\s]*) | generic | | 1,2,3,4 | 4022881949e858899 |
| Michael | person | ({TOKEN})\s([A-Z][a-z]+[A-Z\s]*) | generic | | 1,2,3,4 | 2ce76612fe20eca27 |
| Office | government | ([\w\d'&\-\.\s]+{TOKEN}) | generic | office | 1,2 | 428f75ba9249c9f96l |
| Justice | judge | ([\w\s\.']*?)({TOKEN})([\s]?[\w]*)( | generic | justice | 4,3,1,2 | 23609b93066abc8c9 |
| Playboy | company | ([\w\s]*)({TOKEN}(\sCompany\sC | specific | | 2,1 | 12a72a02384b147c1 |
| Mandrakesoft | company | ([\w\s]*)({TOKEN}(\sCompany\sC | specific | | 2,1 | 253231af1f5132cc63 |
| Hearst Holdings | company | ([\w\s]*)({TOKEN}(\sIncorporated | specific | | 2,1 | a2065f875e52ef9111 |
| Peoplesoft | company | ([\w\s]*)({TOKEN}(\sIncorporated | specific | | 2,1 | 709639473f99fba46 |

Figure 5.8: The portion of the gazetteer for use in named-entity recognition

## 5.5 Summary

In this chapter, we have provided an overview of the implementation of the prototype in accordance to the solution framework. The flow of control from the very beginning when questions are asked to the end when responses are generated was made explicit to promote a clearer view of the functionalities of this new question answering system. The main development tools for the prototype are Perl language and MySQL database, and the reasons why these tools are chosen are also justified.

At this stage, we believe that this question answering prototype known as NaLURI will serve the purpose of this research, which is to extend the type of questions supported, and capable of producing better quality of answers. This belief is founded on the fact that the prototype was developed in accordance to the solution framework and the framework was carefully designed based on the criteria stated in the hypothesis. In order to prove this belief and consequently, the hypothesis, two evaluations are performed, and the process and results are documented in the next chapter.

# Chapter 6

# Evaluation and Result

## 6.1 Introduction

The evaluation of question answering systems has been largely reliant on the TREC corpus and it works relatively well with non-dynamic responses. It gets more difficult to evaluate NaLURI as there is no baseline or comparable systems in the field of news on Cyberlaw cases. Furthermore, due to the dynamic nature of the responses, there is no right or wrong answer as there are always responses to justify the absence of an answer. Besides, developing a set of test questions is easier said than done because unlike the open-domain evaluations, where test questions can be mined from question logs like Encarta, no question sets are at the disposal for domain-oriented evaluations. The predicament remains in the preparation of question sets of different domain for evaluation without giving rise to any fairness issue. People will tend to be very skeptical with the use of different question sets in one evaluation that compares an array of systems. Question like, *"I find that the questions used for evaluating XXX are more difficult that those used for evaluating YYY. This will certainly make YYY better"* will usually arise.

The evaluation criteria namely response time and response quality are used for evaluation in this research. These two criteria attempt to look at the practicality issue of question answering approaches from two dimensions namely time and quality of answer. In this chapter, we will evaluate the prototype implemented in Chapter 5 using quantitative approach for response time and black-box approach for response quality.

## 6.2 Response Time

Response time concerns only with how much time is required to provide an answer or response to a question. The actual response is irrelevant in this evaluation. The response time of NaLURI is evaluated using statistical measures (i.e. quantitative approach) with respect to other existing question answering systems that are available online namely AnswerBus and START. Besides being accessible online, these two systems are also chosen for their approach, allowing representation from each of the current two approaches towards question answering.

### 6.2.1 Assumptions during Performance Evaluation

Evaluating response time in this research is extremely challenging, if not impossible due to networking, computing and other inherent factors:

- Servers that host the three question answering systems (i.e. AnswerBus, START and NaLURI) are located at different physical locations (i.e. two at different states in the United States and one in Malaysia). Latency related to the distance, the different network infrastructure and traffic during peak period may effect the response time. At this level of research, it is not possible to move all three servers to one particular location;

- Specifications of the three servers are unknown and are very likely to be different. System hosted on more powerful machine may have the advantage over system on server with poorer specifications. At this level of research, it is not possible to host all three systems on the same server or on three different servers with the same specification;

- The loads of the three servers are very likely to be different at any particular point in time. System hosted on server that have low burden may have the advantage over busy servers. At this level of research, it is not possible to make constant the burden of all the three servers; and

- The three systems do not belong to the same class of approach and each has different level of processing demand imposed by additional components.

In short, the evaluation of response time in this research can never be fair. Instead, moves are taken to *minimize* the impact of the factors mentioned above:

- NaLURI is uploaded to a web server in Malaysia and made publicly available. Then all the three systems are accessed using the same computer running on an AMD Athlon XP-m 848MHz and connected using a 46.6 kbps dial-up connection. This is to make sure that the server that hosts NaLURI is not in the same local network as the client machine used to probe the system;
- After each question is submitted, a lag of 10 seconds is introduced before the next question. This is to ensure that the performance of the remote question answering system would not be affected by its work on the previous question;
- The two servers in the United States are accessed during the off-peak hours from GMT 08:00 to 09:00 (16:00 to 17:00 Malaysian time and 03:00 to 04:00 United States eastern seaboard time). This is to ensure that miscellaneous traffics that might have impact on these remote servers' performance are brought to the lowest level possible; and
- The same set of questions based on Cyberlaw cases is used for evaluation. This is to ensure that all the systems under evaluation are faced with questions of similar difficulty. AnswerBus and START are designed to handle open-domain information and by definition, they must be able to handle questions on Cyberlaw too. Moreover, the purpose of this response time evaluation is merely to asses the speed of each system in generating responses, regardless of whether the responses produced by AnswerBus and START are correct or not.

By adopting the precautionary measures to minimize the impact due to various factors, we can now assume that the network and computing factors, and also the difficulty of question issue are negligible.

### 6.2.2 Quantitative Approach for Performance Evaluation

A set of 45 test questions concerning Cyberlaw cases is prepared and used to query all three systems remotely over the World Wide Web. Like what is carried out during the performance evaluation of AnswerBus, a computer program is developed to send the questions one by one to AnswerBus, START and NaLURI and the response time is recorded. The results of previous evaluation by researchers of AnswerBus are used in this evaluation as a benchmark for ensuring that the results from this evaluation do not deviate too significantly. The response time for each question submitted to AnswerBus, START and NaLURI is collected and analyzed for average and standard deviation. The patterns of the response time for the three systems are depicted in the following graph in Figure 6.1. For details on the response time of each question, please refer to Appendix A.

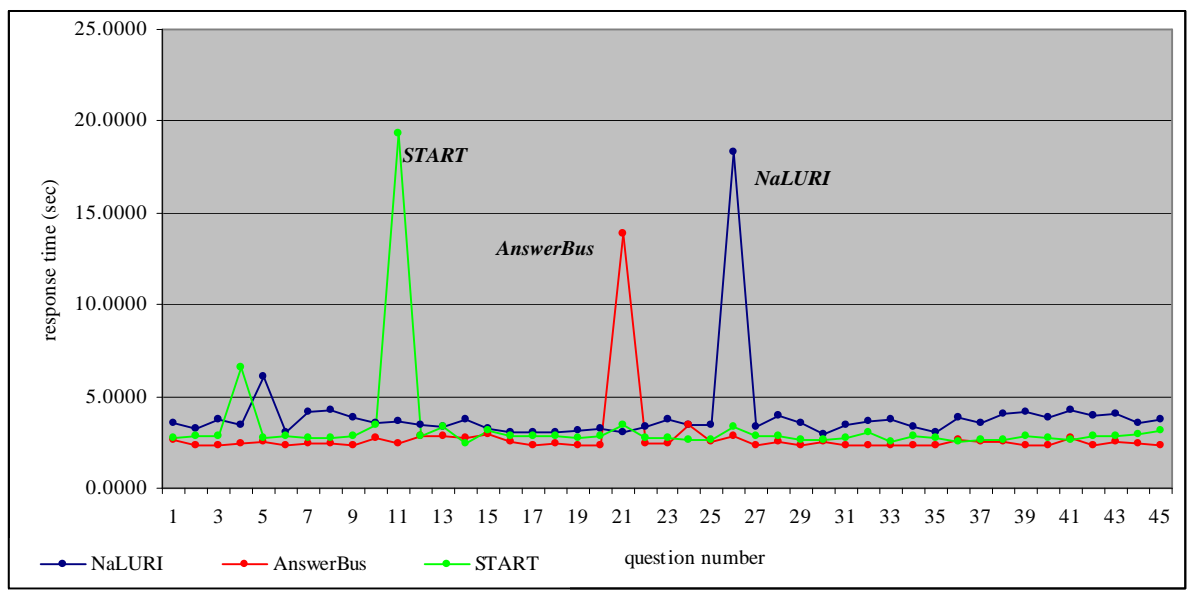

Figure 6.1: Response time for AnswerBus, START and NaLURI

The maximum, minimum and average response time and also the standard deviation obtained from this evaluation are displayed in Table 6.1. The results obtained from the evaluation by the researchers of AnswerBus (Zheng, 2002a) on START and AnswerBus is also provided in the following Table 6.2 to show that the evaluation for response time carried out in this research is actually in line with theirs.

Table 6.1: Average response time and standard deviation for AnswerBus, START and NaLURI

| Systems | Maximum Time | Minimum Time | Average Time | Standard Deviation |
|---|---|---|---|---|
| AnswerBus | 13.8656 | 2.3033 | 2.7469 | 2.8599 |
| START | 19.3377 | 2.4234 | 3.2798 | 6.2859 |
| NaLURI | 18.3106 | 2.8989 | 3.9357 | 4.9586 |

Table 6.2: Average response time and standard deviation for AnswerBus and START by researchers of AnswerBus

| Systems | Maximum Time | Minimum Time | Average Time | Standard Deviation |
|---|---|---|---|---|
| AnswerBus | 15.06 | 3.79 | 7.20 | 3.07 |
| START | 62.07 | 2.02 | 9.84 | 7.45 |

The response time from our evaluation is shown to have the nearly similar outliers as the result produced by the researchers of AnswerBus, where the response time for the AnswerBus system is within the range of 2.3033 to 13.8656 in our evaluation and 3.79 to 15.06 in their evaluation. The standard deviations also exhibit some similarity between the two evaluations. The only major difference is the average time that might be caused by the number of questions used for evaluation. In the evaluation by researchers of AnswerBus, a total of TREC-8's 200 questions are used whereas in our evaluation, only 45 questions of TREC-10 are used for AnswerBus and START. The considerable similarity of the statistical measures between this evaluation and the one done by researchers of AnswerBus has reinforced the validity of this evaluation.

## 6.3 Response Quality

This evaluation is to ensure that the question answering approach does not only produce responses in the shortest time possible, but also generates accurate responses that directly answers the question, and not merely passages or snippets extracted from the full document. In this evaluation, we assess and rank AnswerBus, START and NaLURI for their ability in generating compact and direct responses in face of various anomalies.

### 6.3.1 Considerations for Alternative Measures

To evaluate NaLURI and other systems alike in an absolutely fair and controlled environment for response quality, an alternative measure that is agreed upon by members of the understanding and reasoning community is required. This measure should be capable of handling information in the form of knowledge base and classification of response extending beyond the logical correct/incorrect.

The new measure must put into consideration the three crucial factors related to the inherent nature of question answering systems based on natural language understanding and reasoning:

- systems based on understanding and reasoning uses knowledge base as information source and there are no numerical measurements for such unit of information. In systems where information retrieval is their backbone, the unit of information has always been a document. We can say, *"out of the three documents retrieved, two answers the question"*. But can we actually say *"two out of the three meaning or knowledge produced answers the question"*?; and
- responses generated by such systems are subjective and a scale whereby everyone in the community of understanding and reasoning agrees on measuring the quality of responses of such nature need to be produced. A scale where everyone can actually refer and say that a response to a question is 45% correct for example, is needed.

- preparation of the questions set must put into consideration that the peer systems under evaluation may or may not be from the same domain, for example, there are two systems to be evaluated where one supports the biological disease domain while the other handles agricultural domain. Given that, how are we going to craft or prepare the questions in a way to prevent any controversy concerning the fairness of the evaluation?

All in all, only with the presence of new and non-refutable metrics can the formal evaluation for this new question answering approach be done. Until then, the validity of comparing and evaluating question answering systems based on understanding and reasoning will always be a topic of debate. A formal evaluation is crucial to promote further research interest and growth in this area.

### 6.3.2 Black-box Approach for Quality Evaluation

For this research to lead the way and craft some new numerical evaluation methods for systems like WEBCOOP and NaLURI is not feasible. In fact, it is a whole new research question by itself. Instead, for this dissertation, a black-box approach is employed based on the work of Allen (1995), Nyberg & Mitamura (2002), Diekema *et al.* (2004) as discussed in Chapter 1 for evaluating response quality. This approach is further refined by proposing a response classification scheme and a scoring mechanism. To demonstrate this approach, three question answering systems that represent different level of response generation complexity have been selected, namely AnswerBus, START and NaLURI.

To begin with, this black-box approach requires a set of questions that can sufficiently examine the response generation strength of all systems under evaluation. For this purpose, 45 questions of various natures on the Cyberlaw domain is prepared. The set constitutes of only factual questions in two forms. The first takes the form of interrogative sentences in both active (e.g. *Who did the court ruled against in* cases involving Microsoft?) and passive voice (e.g. *The court ruled against whom in cases involving Microsoft?*) that can either uses wh-words (e.g. *Who sues Microsoft?* and *How many companies were involved in cases with Microsoft?*) or true/false (e.g. *Is it true that the judge James Ware has sided with Microsoft in an antitrust lawsuit by Realnetworks?*). The second takes the form of imperative sentence (e.g. *List the companies that were involved in lawsuits with Microsoft*). These questions will be used to probe the systems and the actual responses are gathered for later use. Details of the questions and responses for the three systems are available in Appendix B, C and D.

For this approach, a classification scheme that consists of categories to encompass all possible types of response from all systems under evaluation is proposed. This scheme consists of three category codes and is designed based on the quality of responses as perceived by general users and is not tied down to any implementation detail of any systems. This makes the scheme generally applicable to all evaluations of question answering systems with different approaches. Under this scheme, two general categories *BQ_?* and *LQ_?*, where *?* is systems initial, which represent the best and lowest quality response for each system, and one dynamic category *Oj_?*, where *j* is an integer, which represents other evaluation-specific criteria, is defined. Evaluators can create as many new categories as required by the types of systems under evaluation using *Oj_?*. The *Oj_?* category not only makes this scheme expandable but also dynamic because as technology progresses, the response generation capability of systems may increase and in such cases, evaluators can define evaluation-specific categories. For this evaluation, *O1_?* for quality of response in the event of no answer and *O2_?* for response that suggest possible spelling mistake are defined. In this evaluation, the initials for AnswerBus, START and NaLURI are *A*, *S* and *N* respectively.

Next, using these codes, the evaluators will try to observe and classify each response into one of the categories. The classification is done based on the manual observation by evaluators who are guided by the criteria of each category. If the evaluator comes across a response that is generated by system *a* for example, and the response appears to be an uninformative attempt to notify the user that no valid answer can be found, then the response can be classified as *O1_a*. This is to say that system *a* generates uninformative response in the event of no answer.

The responses generated by AnswerBus are scrutinized, and from the nature of their responses, they can be grouped into three relevant categories:

- BQ_A: responses with some elements that meet the questions' requirements (e.g. question 10 in Table 6.3);
- LQ_A: responses with completely irrelevant information (e.g. question 37 in Table 6.3);
- O1_A: uninformative responses in the event of no answers (e.g. question 43 in Table 6.3); and
- O2_A: not relevant.

Table 6.3: Part of the responses by AnswerBus

| | Question | Response |
|---|---|---|
| question 10 | List the companies who sued Microsoft? | • The United States Department of Justice (DOJ) and the 19 US states who successfully sued Microsoft over Anti-Trust violations have requested that the US court spilt Microsoft in to two separate companies.<br>• None of these companies sued Microsoft when it began giving away a free codec in Windows Media Player, because they could not be sure their technology was in there. |
| question 37 | Which judge presided the ruling of the case by RealNetworks against Microsoft? | • Grimaldi, Microsoft Judge Says Ruling at Risk, Wash.<br>• A month ago, lawyers for the accused ISV ElcomSoft asked a US District Judge Ronald Whyte, who presided over the Sun-Microsoft Java suit, to drop the case because the DMCA is unconstitutional. |
| question 43 | Who presided the closing of the spamming case against Howard Carmark? | I'm not sure what you mean. |

Then, the responses generated by START are studied. Similarly, based on the nature of the responses produced, the responses can be grouped into three relevant categories:

- BQ_S: responses that provide accurate and direct answers to questions;
- LQ_S: not relevant;

- O1_S: uninformative responses in the event of no answers (e.g. question 38 in Table 6.4); and
- O2_S: responses that suggest possible spelling mistake (e.g. question 29 in Table 6.4).

Table 6.4: Part of the responses by START

|  | Question | Response |
|---|---|---|
| question 29 | When was the closing of the caset against Microsoft? | The word CASET may be misspelled |
| question 38 | Who chaired the closing of the case by RealNetworks against Microsoft? | Unfortunately, I don't know who chaired the closing of the case against Microsoft. |

Lastly, based on the nature of the responses produced by NaLURI, the responses are grouped into three relevant categories:

- BQ_N: responses that provide accurate and direct answers to questions (e.g. question 43 in Table 6.5);
- LQ_N: not relevant;
- O1_N: informative responses in the event of no answers (e.g. question 32 in Table 6.5); and
- O2_N: responses that suggest possible spelling mistake (e.g. question 29 in Table 6.5).

Table 6.5: Part of the responses by NaLURI

|  | Question | Response |
|---|---|---|
| question 29 | When was the closing of the caset against Microsoft? | There are some spelling errors in the question. When was the closing of the **caset** against Microsoft |
| question 32 | When was the filing of the case against Microsoft by RealNetworks? | There is no such filing event involving RealNetworks as plaintiff. |
| question 43 | Who presided the closing of the spamming case against Howard Carmark? | Attorney General Eliot Spitzer New York chaired the resolution of the case |

After classification of the responses is done, a scoring mechanism is used to determine which system produces the best overall quality of responses. A pair-wise relative comparison is performed and points are assigned based on superiority of responses of the same category. If there are $n$ systems under evaluation, then there should be $_nC_2 = k$ pairs. Let $?_i$ represents the pair of system $?_{xi}$ and $?_{yi}$. To perform the scoring, a table is constructed as shown in Table 6.6 where the column header represents all the $?_1, ?_2,..., ?_k$ pairs. The row header consists of the two general categories *BQ_?* and *LQ_?* and other evaluation-specific categories *Oj_?*.

Table 6.6: Template for scoring mechanism

| Category | $?_1$ | | $?_2$ | | ... | $?_k$ | |
|---|---|---|---|---|---|---|---|
| | $?_{x1}$ | $?_{y1}$ | $?_{x2}$ | $?_{y2}$ | | $?_{xk}$ | $?_{yk}$ |
| BQ_? | | | | | | | |
| LQ_? | | | | | | | |
| Oj_? | | | | | | | |
| Total | | | | | | | |

Then for every $?_i$, $BQ\_?_{xi}$ with $BQ\_?_{yi}$, $LQ\_?_{xi}$ with $LQ\_?_{yi}$ and other $Oj\_?_{xi}$ with $Oj\_?_{yi}$ are compared. The rules for superiority comparison and assigning of score are as follows:

- if the description of the responses for $?_{xi}$ is better than $?_{yi}$ under a particular category, then $?_{xi}$ is assigned with 1 and $?_{yi}$ is assigned with 0 under the same category;

- if the description of the responses for $?_{xi}$ is inferior compared to $?_{yi}$ under a particular category, then $?_{xi}$ is assigned with 0 and $?_{yi}$ is assigned with 1 under the same category; and

- if the description of the responses for $?_{xi}$ is the same as $?_{yi}$ under a particular category, then both $?_{xi}$ and $?_{yi}$ are assigned with 0 under the same category.

After filling up all the cells in the score table, summation of scores for every $?_{xi}$ and $?_{yi}$ under all categories is performed. Provided next are a few examples to demonstrate the working behind the scoring mechanism. The best quality responses of AnswerBus, BQ_A has the possibility of containing irrelevant elements, whereas responses generated by START are always correct and directly answer the questions.

Due to this, the best quality responses from START, which belongs to BQ_S, are a level higher than the best quality responses of AnswerBus, BQ_A. Hence, for the pair "*START vs. AnswerBus*", START will be assigned with one point. In the case of ties, like other categories O_1S and O_1A which demonstrate the same quality of responses in the event of no answers, no points will be given for either side of the pair "*START vs. AnswerBus*". Consider another example where the responses from O_2S, which attempt to alert the users of possible spelling mistake, make START an additional level higher than AnswerBus. This provides START with another additional point in the pair "*START vs. AnswerBus*". The comparison will be done on all the three systems, thus giving three possible pairs.

From Table 6.7, it can be observed that AnswerBus has a total score of 0 + 0 = 0, NaLURI with a total score of 3 + 1 = 4 and START with a total score of 0 + 2 = 2.

Table 6.7: Scoring table for quality evaluation using pair-wise relative comparison

| Category | AnswerBus vs. NaLURI | | START vs. NaLURI | | START vs. AnswerBus | |
|---|---|---|---|---|---|---|
| | **AnswerBus** | **NaLURI** | **START** | **NaLURI** | **START** | **AnswerBus** |
| BQ_ / LQ_ | 0 | 1 | 0 | 0 | 1 | 0 |
| O_1 | 0 | 1 | 0 | 1 | 0 | 0 |
| O_2 | 0 | 1 | 0 | 0 | 1 | 0 |
| **Total** | **0** | **3** | **0** | **1** | **2** | **0** |

## 6.4 Implications and Validity of the Results

In this section, conclusions are drawn concerning how the introduction of components in natural language understanding and advanced reasoning makes question answering better in terms of response quality without compromising response time. These components can either originate from new ideas or innovative use of existing concepts. Referring to Table 6.8, the complexity and the demanding nature of the system increases as more components are included, but at the same time, the quality of responses produced also improves. This can be attributed to the fact that more and more computation is performed on decreasing amount of information in an attempt to exploit more aspects of natural language to achieve richer meaning representation.

Table 6.8: Natural language understanding and advanced reasoning components in AnswerBus, START and NaLURI

| natural language understanding, advanced reasoning components and other features | AnswerBus | START | NaLURI |
|---|---|---|---|
| sentence parsing | v | v | v |
| named-entity recognition | v | x | v |
| relation extraction | x | v | v |
| anaphora resolution | x | x | v |
| semantic unification | x | x | v |
| semantic representation | x | v | v |
| traceable answer discovery | x | v | v |
| explanation on failure | x | x | v |
| dynamic answer generation | x | v | v |
| Ontology | x | x | v |

Firstly, looking at the results from the response time evaluation, it can be observed that AnswerBus achieved the fastest average response time of 2.7469 seconds. As for START and NaLURI, their average response time hover over the range of 3.2 to 3.9 seconds. These figures do not come as a surprise as AnswerBus is a system of a totally different class as compared to START and NaLURI but this is not what this evaluation is trying to show. It is vital to make it clear that it is not an intention to demonstrate that NaLURI is faster in any sense because with the amount of processing demand shown in Table 6.8, it is impossible. Instead, the graph have reveals that response time of NaLURI is actually on par with other systems practicing approach which is less demanding in nature. Again, the fact that the response time may be tainted by miscellaneous factors is not defended, even after various precautionary steps have been taken. In addition to the above revelation, the results have also shown that the response time of NaLURI is also consistent despite the uncertainty in the type of questions through relatively low standard deviation as compared to START. This is an important characteristic for systems like NaLURI that is designed to handle questions of various natures, unlike other existing question answering systems that are only capable of tackling wh-questions.

As for response quality, NaLURI ranked first with 4 points, followed by START with 2 points and lastly, AnswerBus with 0 points. This makes the responses generated by NaLURI conditionally better in relative comparison with START and AnswerBus. The condition is to assume that the evaluators' classification is consistent throughout, and the set of questions used for evaluation is exhaustive enough to trigger all possible responses. In the case of new systems being added to the evaluation, the classification and scoring process needs to be redone.

The approach of evaluating the response quality through response classification scheme and a scoring mechanism has revealed to us that the lack or addition of components has great impact on the response quality. Please refer to Table 6.8 for the list of components implemented by each of the three systems evaluated. One of the criteria that have contributed to the higher score of NaLURI for instance, is the capability of the system in generating dynamic responses to suit the various anomalous situations. Useful responses for example, can be dynamically generated by NaLURI to cater to the condition when no answers are available. This ability can be attributed to the inclusion of the two advanced reasoning components namely explanation on failure and dynamic answer generation. Such useful responses can help the users to clear any doubts related to the actual state of the knowledge base. This is obviously a desirable trait for a question answering system. Table 6.9 clearly shows how each of the categories of responses are achieved through the different approach towards question answering that implements diverse components in information retrieval, natural language understanding and reasoning.

Table 6.9: Relation between quality of responses and components in question answering

| Categories of responses | AnswerBus | START | NaLURI |
|---|---|---|---|
| responses with some elements that meet the questions' requirements, while the rest are irrelevant materials. | achieved through mere sentence parsing and information retrieval | n/a | n/a |
| responses that provide accurate and direct answers to questions | n/a | achieved through higher-level of natural language understanding and reasoning | achieved through higher-level of natural language understanding and reasoning |
| quality of responses in the event of no answers | uninformative due to the lack of advanced reasoning | uninformative due to the lack of advanced reasoning | informative due to the use of advanced reasoning |
| responses that suggest possible spelling mistake | n/a | achieved through additional linguistic feature | achieved through additional linguistic feature |

After having concluded that NaLURI is comparatively better than the other two systems, skeptical thoughts may arise. Firstly, thoughts may arise concerning the domain of the question. People may be very tempted to question that the evaluation is inclined towards NaLURI because the question set is prepared in the same domain as NaLURI, which is Cyberlaw. However, one question stands out, which is, what is the domain of START and AnswerBus? As Zheng (2002b, p.1) said "*AnswerBus is an open-domain question answering…*" while START is capable of handling many domains based on the statement "*our system answers millions of natural language questions about places (e.g., cities, countries, lakes, coordinates, weather, maps, demographics, political and economic systems), movies (e.g., titles, actors, directors), people (e.g., birth dates, biographies), dictionary definitions, and much, much more…*" by Katz *et al.* (2002, p.3). Hence, we do not see any problem in START and AnswerBus handling Cyberlaw questions. Secondly, thoughts may arise concerning the nature of the question. People may question that the evaluation is unfair towards START and AnswerBus because the nature of the questions used to evaluate vary greatly and cover beyond wh-questions. The researchers however, would like the readers to recall that the aim of this evaluation is to assess and rank systems of any approach based on the quality of responses generated. How can these systems be ranked if we merely use wh-questions, knowing that given the present state of question answering technology, handling wh-questions is no more a challenge? Hence, benchmark for question answering systems has to progress with time by considering various state-of-the-art factors instead of dwelling in the past.

Even though the core of this black-box approach is justifiable and have been proposed by many researchers as discussed in Chapter 1, but like every other evaluation measures, limitations may exist that actually threatens the accuracy and fairness of the metrics. There are several inherent limitations in this black-box approach of evaluation. Firstly, the approach relies on classification of large collection of responses generated by question answering systems. Due to the subjective nature of manual classification that can be affected by many factors, results may vary depending on the evaluator. Hence, the consistency of evaluator is extremely important to minimize human errors and contradictory results. Secondly, the evaluation relies on grouping of responses based on the traits exhibited. As there are infinite ways of asking questions, this makes the range of responses generated to be very big. This makes the task of classification over this range of responses very time-consuming, if not impossible. If a set of samples is taken for evaluation, then the representation from every unique class of responses may not be available. This will cause the score for that particular system to be inaccurate.

One of the possible enhancements on the evaluation approach is to refine and widen the scope of the scoring mechanism. Firstly, the categories of responses can be extended to allow more types of responses to be included in the evaluation. Secondly, each category of responses can be further refined to make the evaluation more specific. For example, the category in the event of no answer (i.e. $O\_1$) can be further segregated to make new categories handle the various possible situations that may cause the system not being able to find the answer. These changes will turn the scoring mechanism into a better representation of the actual question answering system. Another possible improvement is to formalize the ways to prepare the question set, the scoring mechanism and to properly define the rules and guidelines in performing the classification. This will help the evaluators in preparing questions that cover a wider variety of possible responses and ensuring consistency throughout the classification process.

## 6.5 Summary

In this chapter, the steps and assumptions involved in evaluating the prototype developed in Chapter 5 is presented. The prototype is evaluated for two things: response time and response quality. Firstly, the response time evaluation is to ensure that the introduction of complex natural language understanding and advanced reasoning into question answering does not slow down the ability of the system in generating responses to the extent not tolerable by users. The results from response time evaluation have shown that despite the various negligible fairness issues, NaLURI actually performs on par with other question answering systems which are less demanding in nature. This implies that the practicality of the system has not been affected despite the heavy processing required by the components in NaLURI. Secondly, the response quality evaluation is to verify that the introduction of natural language understanding and advanced reasoning can indeed solve the problem of question and response restriction. The results from the response quality evaluation have shown that NaLURI is more capable of handling wider range of questions, and the responses generated are not only direct and compact, but also informative in various anomalous cases.

In a nutshell, it is shown that the innovative introduction of natural language understanding and advanced reasoning in the form of the three criteria stated in the hypothesis have both solved the problems addressed in this research, and also takes care of practicality factors. Consequently, from the experiments conducted in the form of the design, implementation and evaluation of a prototype system, the evidence we have is necessary and sufficiently strong for us to accept the hypothesis formulated in this research.

# Chapter 7

# Conclusions

## 7.1 Summary and Contributions

This dissertation has bought forward an initial problem concerning the question and responses restrictions in existing question answering systems, and the resultant problems that ensue from the lack of ideal integration of natural language understanding and reasoning. Accordingly, a hypothesis is formulated stating that the problems can be solved by introducing a practical approach for question answering which combines full-discourse natural language understanding, powerful and expressive representation formalism like semantic network in the Cyberlaw domain, and network-based reasoning that supports advanced reasoning. The hypothesis was finally proven to be valid and accepted after a series of solution framework design, prototype implementation and evaluation.

The acceptance of the hypothesis of this research signifies that the practical introduction of understanding and reasoning into question answering can indeed improve the overall quality of domain-oriented question answering systems in terms of the diversity of question supported and also quality of response. Users are no longer constrained and are allowed to ask questions beyond the use of wh-words and obtain responses that exhibit intelligence. This approach also allows the system to understand real-life information and reason with facts from any domain (given that the ontology, gazetteer and semantic network for that particular domain are available) with minimal fuss.

During the course of designing the solution framework, several contributions were accomplished:

- introduction of two cooperative algorithms for named-entity recognition namely a context-free grammar for noun phrase chunking and two-pass method for category assignment, and an entirely novel approach for discourse analysis using a four-stage discourse integration with light-weight anaphora resolution.
- introduction of a new network-based reasoning approach founded on three algorithms namely network-to-path reduction, selective path matching with advanced reasoning features, and template-based response generation.

The algorithmic contributions above are highly applicable for future researches in the field of intelligent and dynamic responses for question answering because of the modular nature of the algorithms and the proper documentations.

All in all, the true significance of this research extends beyond the algorithmic contributions. The practical integration of natural language understanding and reasoning into question answering attempted by this research, whose practicality was founded through a series of evaluation, will eliminate the idea that the introduction of these two elements into question answering will raise much practical issues. This will pave way for more future researches in this practical approach towards question answering based on natural language understanding and reasoning.

## 7.2 Limitations and Proposed Solutions

Two limitations of the current prototype and the reasons behind such drawbacks will be presented in this section. Then, possible solutions are proposed that will pave way to future work and enhancement.

One of the limitations of NaLURI is the need to structure the question to be grammatically correct. As the processing by Minipar is the first step in understanding of a sentence and its output which is mandatory for further processing, any discrepancies in the output caused by grammatically incorrect questions will produce undesirable results in the subsequent stages. Consider the question posted by a user that contains some grammatical error *"List of judges who presided cases file by Microsoft?"* for example.

| offset | grammatical categories | grammatical relationships | modifier | head (head offset) |
|---|---|---|---|---|
| | OUTPUT OF SENTENCE PARSING | | | |
| 1 | N | wh | List | case (6) |
| 2 | Prep | mod | of | list (1) |
| 3 | N | pcomp-n | judge | of (2) |
| 4 | N | wh | who | fin (E0) |
| 5 | V | i | preside | fin (E0) |
| 6 | V | i | case | fin (E1) |
| 7 | N | obj | file | case (6) |
| 8 | Prep | mod | by | file (7) |
| 9 | N | pcomp-n | Microsoft | by (8) |

Figure 7.1: Erroneous Minipar output due to grammatically incorrect question

The error is explicitly reflected in the corresponding output of the question produced by Minipar as shown in Figure 7.1. Please take note of the wrong part-of-speech assigned to token 6 and 7. As verb is an extremely important indicator of new relations, the absence of *"file"* as a verb has caused a missing event.

Such undesirable situation can be said as a problem inherent to all systems that practice full natural language understanding and can seriously affect the robustness of the system in handling questions produced by humans where one or two grammatical errors are sometimes unavoidable. To overcome this drawback, this research proposes a yet to be implemented solution through the use of a spelling and grammar checker. All questions will need to pass through this checker to make sure no words have been spelt wrongly and no grammatical errors exist. Only then will the error-free question be sent to Minipar for parsing.

Another limitation that is implicit to all systems based on full natural language understanding is the reliant on external lists and dictionary. The main lists collection in NaLURI, which is also the only collection, is known as gazetteer, where the term is normally understood by the natural language understanding community. The gazetteer consists of two important types of words namely nouns for named-entity recognition and nouns, verbs and prepositions for discourse integration. As the gazetteer is manually handcrafted, if during any stage of named-entity recognition, a required trigger word does not exists, then the system will not be able to tag the noun phrase. If the noun *"Singapore"* is encountered during named-entity recognition for example, and the gazetteer does not contain this name, then the system will not be able to recognize this noun as a valid named-entity of the category *"country"*. The same goes for the module of discourse integration that is reliant on the other portion of the gazetteer. While a solution in the long term is available namely to automate the discovery and maintenance of the ontology and the gazetteer, a shorter and more realistic way out is to have some sort of feedback mechanism whereby any unrecognizable words during named-entity recognition and discourse integration are immediately logged and presented for maintenance actions. The justification for such action is that even though the named-entity recognition might not be perfect during the first try because of the missing word in gazetteer, but during the next try, the recognition mechanism will be performing one word better than the previous run. In short, this solution to the problem will in a way make the named-entity recognition and discourse integration continuously improving in accuracy.

## 7.3 Future Work

Firstly, by realizing the limitations of NaLURI presented in the previous section, two possible future works that can bring NaLURI to a higher level of robustness namely the introduction of a spelling and grammar check, and a feedback mechanism to manually upgrade the gazetteer, is identified.

Secondly, in terms of the reasoning capability of NaLURI, the existing advanced reasoning and response generation capability is planned to be strengthened by implementing additional features like generating intensional responses when the number of direct responses is very large or too small and also look for more advanced natural language response generation techniques to replace the current template-based approach.

Lastly, which is also the most challenging would be the research on the automated development and maintenance of the ontology and the gazetteer. This work will bring the dream of having question answering systems based on natural language understanding and reasoning portable across multiple domains a step closer to reality. This work will not be achieved easily due to the need to call for a whole new research to study the overall requirements and together with it, an accompanying dissertation.

# References

Abney, S. (1991). Parsing by chunks. In *Principle-Based Parsing: Computation and Psycholinguistics* (Berwick, R., Abney, S. & Tenny, C., eds.), pp. 257-278. Boston: Kluwer Academic Publishers.

Abney, S. (1995). *Dependency grammars and context-free grammars*. Retrieved 5 May 2005, from http://www.vinartus.net/spa/publications.html.

Alfonseca, E., Manandhar, S. & DeBoni, M. (2001). A prototype question answering system using syntactic and semantic information for answer retrieval. In *Proceedings of the 10$^{th}$ Text Retrieval Conference* (Voorhees, E. & Harman, D., eds,), pp. 680-686.

Allen, J. (1995). *Natural language understanding*. California: Benjamin/Cummings.

Anon. (2004). *Incident statistics 2004*. Retrieved 27 April 2005, from http://www.niser.org.my/statistics.html.

Appelt, D., Bear, J., Israel, D., Hobbs, J. & Tyson, M. (1993). FASTUS: A finite-state processor for information extraction from real-world text. In *Proceedings of the 13$^{th}$ International Joint Conference on Artificial Intelligence* (Ruzena, B., eds,), pp. 1172-1178.

Benamara, F. (2004). Cooperative question answering in restricted domains: the WEBCOOP experiment. In *Proceedings of the ACL Workshop on Question Answering in Restricted Domains*.

Benamara, F. & Saint-Dizier, P. (2003). Dynamic generation of cooperative natural language responses. In *Proceedings of the EACL Workshop on Natural Language Generation.*

Benamara, F. & Saint-Dizier, P. (2004). Advanced relaxation for cooperative question answering. In *New Directions in Question Answering* (Maybury, M., ed.), p. 263-274. Massachusetts: MIT Press.

Bikel, D., Schwartz, R. & Weischedel, R. (1999). An algorithm that learns what's in a name. *Machine Learning, 34*(1-3):1-20.

Bobrow, D., Thompson, H., Winograd, T., Norman, D., Kay, M. & Kaplan, R. (1977). GUS, a frame-driven dialog system. *Artificial Intelligence, 8*(2):155-173.

Bratko, I. (2001). *PROLOG: Programming for artificial intelligence*. England: Addison-Wesley.

Breck, E., House, D., Mani, I., Burger, J. & Light, M. (1999). Question answering from large document collections. In *Proceedings of the AAAI Fall Symposium on Question Answering Systems*.

Brill, E. (1992). A simple rule-based part of speech tagger. In *Proceedings of the 3rd Conference on Applied Natural Language Processing*, pp. 152-155.

Buckley, C. (1985). *Implementation of the SMART information retrieval system*. Technical Report (TR85-686), Cornell University.

Burke, R., Kulyukin, V., Schoenberg, S., Lyninen, S., Hammond, K. & Tomuro, N. (1997). Question answering from frequently-asked question files: Experiences with the FAQ finder system. *AI Magazine, 18*(2):57-66.

Burkert, G. & Forster, P. (1991). Representation of semantic knowledge with term subsumption languages. In *Proceedings of the ACL Workshop on Recent Advances in Natural Language Processing and Information Retrieval*.

Cardie, C., Buckley, C., Pierce, D. & Ng, V. (2000). Examining the role of statistical and linguistic knowledge sources in a general-knowledge question-answering system. In *Proceedings of the 6th Conference on Applied Natural Language Processing*, pp. 180-187.

Chen, H. & Chen, K. (1993). A probabilistic chunker. In *Proceedings of the 6th Research on Computational Linguistics Conference*, pp. 99-117.

Chomsky, N. (1985). *Syntactic structures*. The Hague: Mouton.

Chomsky, N. (1995). *The minimalist program*. Massachusetts: MIT Press.

Chung, H., Han, K., Rim, H., Kim, S., Lee, J., Song, Y. & Yoon, D. (2004). A practical QA system in restricted domains. In *Proceedings of the ACL Workshop on Question Answering in Restricted Domains*.

Clark, P. (1989). Knowledge representation in machine learning. In *Machine and Human Learning* (Kodratoff, Y. & Hutchinson, A., eds.), p. 35-49. London: Kogan Page.

Clarke, C., Cormack, G. & Lynam, T. (2001). Exploiting redundancy in question answering. In *Proceedings of the 24th International Conference on Research and Development in Information Retrieval*.

Cutting, D., Pedersen, J., Sibun, P. & Kupiec, J. (1992). A practical part-of-speech tagger. In *Proceedings of the 3$^{rd}$ Conference on Applied Natural Language Processing*, pp. 133-140.

Davis, R., Shrobe, H. & Szolovits, P. (1993). What is a knowledge representation?. *AI Magazine, 14*(1):17-33.

Diekema, A., Yilmazel, O. & Liddy, E. (2004). Evaluation of restricted domain question-answering systems. In *Proceedings of the ACL Workshop on Question Answering in Restricted Domains*.

Facemire, J. (1989). A proposed metric for the evaluation of natural language systems. In *Proceedings of the IEEE Energy and Information Technologies in the Southeast*.

Fischer, D. (2003). *Natural language understanding: State of the art and some current research*. Retrieved 5 May 2005, from http://dan.f3c.com/nlu/.

Gaasterland, T., Godfrey, P. & Minker, J. (1992). An overview of cooperative answering. *Journal of Intelligent Information Systems, 1*(2):123-157.

Gaizauskas, R. & Humphreys, K. (1996). XI: A simple Prolog-based language for cross-classification and inheritance. In *Proceedings of the 7$^{th}$ International Conference on Artificial Intelligence*, pp. 86-95.

Gaizauskas, R. & Humphreys, K. (1997). *Using a semantic network for information extraction*. Technical Report (CS-97-03), University of Sheffield.

Gaizauskas, R., Humphreys, K., Wilks, Y., Cunningham, H. & Wakao, T. (1995). Description of the LaSIE system as used for MUC-6. In *Proceedings of the 6$^{th}$ Message Understanding Conference*, pp. 207-220.

Green, B., Chomsky, C., Laughery, K. & Wolf, A. (1963). Baseball: An automatic question answerer. In *Computers and Thought* (Feigenbaum, E. & Feldman, J., eds.), p. 207-216. Massachusetts: AAAI Press.

Grossman, M. (1999). *What is cyberlaw?*. Retrieved 30 March 2004, from http://www.mgrossmanlaw.com/articles/1999/what_is_cyberlaw.htm.

Guida, G. & Mauri, G. (1984). A formal basis for performance evaluation of natural language understanding systems. *Computational Linguistics, 10*(1):15-30.

Hearst, M. (1994). Multi-paragraph segmentation of expository text. In *Proceedings of the 32$^{nd}$ Annual Meeting of the Association for Computational Linguistics*, pp. 9-16.

Hendrix, G., Sagalowicz, D., Sacerdoti, E. & Slocum, J. (1978). Developing a natural language interface to complex data. *Transactions on Database Systems, 3*(2):105-147.

Hermjakob, U. (2001). Parsing and question classification for question answering. In *Proceedings of the ACL Workshop on Open-Domain Question Answering*.

Hermjakob, U. & Mooney, R. (1997). Learning parse and translation decisions from examples with rich context. In *Proceedings of the 35<sup>th</sup> Annual Meeting of the Association for Computational Linguistics*, pp. 482-489.

Hirschman, L. & Gaizauskas, R. (2001). Natural language question answering: The view from here. *Natural Language Engineering, 7*(4):275-300.

Hobbs, J. (1977). Resolving pronoun references. In *Readings in Natural Language Processing* (Grosz, B., Jones, K. & Webber, B., eds.), p. 339-352. California: Morgan Kaufmann.

Hovy, E., Hermjakob, U., Junk, M., Lin, C. & Gerber, L. (2000). Question answering in Webclopedia. In *Proceedings of the 9th Text Retrieval Conference*.

Hovy, E., Hermjakob, U., Ravichandran, D. & Gerber, L. (2002a). *Qtargets used in Webclopedia*. Retrieved 25 August 2004, from http://www.isi.edu/natural-language/projects/webclopedia/Taxonomy/taxonomy_toplevel.html.

Hovy, E., Ravichandran, D. & Hermjakob, U. (2002b). A question/answer typology with surface text patterns. In *Proceedings of the Conference on Human Language Technology*.

Hu, R. & Atwell, E. (2003). A survey of machine learning approaches to analysis of large corpora. In *Proceedings of the Workshop on Shallow Processing of Large Corpora*.

Ide, N. & Véronis, J. (1998). Word sense disambiguation: The state of the art. *Computational Linguistics, 24*(1):1-41.

Itai, A. & Dagan, I. (1990). Automatic processing of large corpora for the resolution of anaphora references. In *Proceedings of the 13th International Conference on Computational Linguistics*, pp. 330-332.

Katz, B. (1997). Annotating the world wide web using natural language. In *Proceedings of the 5th Conference on Computer Assisted Information Searching on the Internet*.

Katz, B., Felshin, S. & Lin, J. (2002). The START multimedia information system: Current technology and future directions. In *Proceedings of the International Workshop on Multimedia Information Systems*.

Katz, B. & Levin, B. (1988). Exploiting lexical regularities in designing natural language systems. In *Proceedings of the 12th International Conference on Computational Linguistics*.

King, M. (1996). Evaluating natural language processing systems. *Communications of the ACM, 39*(1):73-79.

Kupiec, J. (1993). MURAX: A robust linguistic approach for question answering using an on-line encyclopedia. In *Proceedings of the 16th International Conference on Research and Development in Information Retrieval*, pp. 181-190.

Kwok, C., Weld, D. & Etzioni, O. (2001). Scaling question answering to the web. In *Proceedings of the 10th International Conference on World Wide Web*.

Liakata, M. (2000). *Pronoun resolution using syntactic information & an induced inference mechanism*. Thesis (Phd). University of Oxford.

Lin, D. (1993). Principle-based parsing without overgeneration. In *Proceedings of the 31st Annual Meeting of the Association for Computational Linguistics*.

Lin, D. (1994). PRINCIPAR: An efficient, broad-coverage, principle-based parser. In *Proceedings of the 15th International Conference on Computational Linguistics*.

Lin, D. (1998). Dependency-based evaluation of MINIPAR. In *Proceedings of the 1st International Conference on Language Resources and Evaluation*.

Lin, J. (2001). *Indexing and retrieving natural language using ternary expressions*. Thesis (Master of Engineering). Massachusetts Institute of Technology.

Lin, J., Sinha, V., Katz, B., Bakshi, K., Quan, D., Huynh, D. & Karger, D. (2003). What makes a good answer? The role of context in question answering. In *Proceedings of the 9th International Conference on Human-Computer Interaction*.

Mauldin, M. (1994). Chatterbots, tinymuds and the Turing test: Entering the Loebner prize competition. In *Proceedings of the 12<sup>th</sup> National Conference on Artificial Intelligence*.

Maybury, M. (2003). Toward a question answering roadmap. In *Proceedings of the AAAI Spring Symposium on New Directions in Question Answering*, pp. vii-xi.

Maynard, D. (2003). Multi-source and multilingual information extraction. *Expert Update, 6*(3):11-16.

Maynard, D., Wilks, Y., Cunningham, H., Tablan, V. & Ursu, C. (2001). Named entity recognition from diverse text types. In *Proceedings of the 4<sup>th</sup> International Conference on Recent Advances in Natural Language Processing*.

Mitkov, R. (1998). Anaphora resolution: The state of the art. In *Proceedings of the 17<sup>th</sup> International Conference on Computational Linguistics*.

Mitkov, R. (2001). Outstanding issues in anaphora resolution. In *Computational Linguistics and Intelligent Text Processing* (Gelbukh, A., ed.), p. 110-125. Springer-Verlag.

Moldovan, D., Pasca, M., Surdeanu, M. & Harabagiu, S. (2002). Performance issues and error analysis in an open-domain question answering system. In *Proceedings of the 40<sup>th</sup> Annual Meeting of the Association for Computational Linguistics*.

Moreale, E. & Vargas-Vera, M. (2003). *A question-answering system using argumentation.* Technical Report (KMI-TR-132), The Open University.

Mueller, E. (1999). *Prospects for in-depth story understanding by computer*. Retrieved 5 August 2004, from http://www.signiform.com/erik/pubs/storyund.htm.

Nyberg, E. & Mitamura, T. (2002). Evaluating QA systems on multiple dimensions. In *Proceedings of the Workshop on QA Strategy and Resources*.

Pasca, M., Harabagiu, S. & Maiorano, S. (2000). Experiments with open-domain textual question answering. In *Proceedings of the 18<sup>th</sup> International Conference on Computational Linguistics*.

Shapiro, S. (1978). Path-based and node-based inference in semantic networks. In *Proceedings of the Workshop on Theoretical Issues in Natural Language Processing*.

Sharples, M., Hutchison, C., Young, D., Torrance, S. & Hogg, D. (1989). *Computers and thought: A practical introduction to artificial intelligence*. Massachusetts: MIT Press.

Sleator, D. & Temperley, D. (1993). Parsing English with a link grammar. In *Proceedings of the 3<sup>rd</sup> International Workshop on Parsing Technologies*.

Srivastava, A. & Rajaraman, V. (1995). A vector measure for the intelligence of a question-answering (Q-A) system. *IEEE Transactions on Systems, Man and Cybernetics, 25*(5):814-823.

Strzalkowski, T., Lin, F., Perez-Carballo, J. & Wang, J. (1999). Evaluating natural language processing techniques in information retrieval. In *Natural Language Information Retrieval* (Strzalkowski, T., ed.), p. 113-142. Boston: Kluwer Academic Publishers.

Voorhees, E. (2003). Overview of TREC 2003. In *Proceedings of the 12<sup>th</sup> Text Retrieval Conference*.

Weizenbaum, J. (1966). ELIZA – a computer program for the study of natural language communication between man and cachine. *Communications of the ACM, 9*(1):36-45.

Wilks, Y. (1985). Right attachment and preference semantics. In *Proceedings of the 2<sup>nd</sup> Conference of the European Chapter of the Association for Computational Linguistics*, pp. 89-92.

Winograd, T. (1972). Procedures as a representation for data in a computer program for understanding natural language. *Cognitive Psychology, 3*(1):1-191.

Witten, I., Bell, T. & Moffat, A. (1999). *Managing gigabytes: compressing and indexing documents and images (2<sup>nd</sup> ed)*. San Francisco: Morgan Kaufmann.

Wong, W., Goh, O.S., Mohammad-Ishak, D. & Shahrin, S. (2004a). Online cyberlaw knowledge base construction using semantic network. In *Proceedings of the IASTED International Conference on Applied Simulation and Modeling* (Hamza, M., eds,), pp. 347-352.

Wong, W., Goh, O.S. & Mokhtar, M. (2004b). Syntax preprocessing in cyberlaw web knowledge base construction. In *Proceedings of the International Conference on Intelligent Agents, Web Technologies and Internet Commerce* (Mohammadian, M., eds,), pp. 174-184.

Woods, W. (1973). Progress in natural language understanding: An application to lunar geology. In *Proceedings of the National Conference of the American Federation of Information Processing Societies*.

Zahri, Y. & Ahmad-Nasir, M. (2003). *Cyber threats: Myths or reality?*. Retrieved 27 April 2005, from http://www.niser.org.my/resources/cyber_threats.pdf.

Zheng, Z. (2002a). AnswerBus question answering system. In *Proceedings of the Conference on Human Language Technology*.

Zheng, Z. (2002b). Developing a web-based question answering system. In *Proceedings of the 11th International Conference on World Wide Web*.

Zweigenbaum, P. (2003). Question answering in biomedicine. In *Proceedings of the 10th Conference of the European Chapter of the Association for Computational Linguistics*.

# Appendix A – Response Time for Individual Question

| | Question | Response Time | | |
|---|---|---|---|---|
| | | NaLURI | AnswerBus | START |
| 1 | AT&T accused Microsoft of committing what crime? | 3.5751 | 2.5837 | 2.7540 |
| 2 | Did Vonage initiate any legal actions against Microsoft? | 3.2547 | 2.3434 | 2.7940 |
| 3 | How many companies were involved in cases with Microsoft? | 3.7153 | 2.3033 | 2.8341 |
| 4 | How many times did AT&T lose in cases initiated by it? | 3.4750 | 2.3934 | 6.6095 |
| 5 | How many times did Microsoft win in cases against it? | 6.0587 | 2.5537 | 2.7640 |
| 6 | Ira Zar was accused of what? | 3.0844 | 2.3133 | 2.8141 |
| 7 | Is it true that the judge James Ware has sided with Microsoft in an antitrust lawsuit by Realnetworks? | 4.1560 | 2.4135 | 2.7339 |
| 8 | List of judges who preside over cases file by Microsoft? | 4.2161 | 2.4135 | 2.7139 |
| 9 | List the companies sued by AT&T? | 3.8455 | 2.3734 | 2.7840 |
| 10 | List the companies who sued Microsoft? | 3.5151 | 2.7755 | 3.4249 |
| 11 | List the judges presiding cases file against Microsoft? | 3.6152 | 2.3934 | 19.3378 |
| 12 | List the judges presiding cases involving Microsoft? | 3.4850 | 2.8741 | 2.8341 |
| 13 | Microsoft sued whom? | 3.3548 | 2.8441 | 3.3148 |
| 14 | Name the companies sued by Microsoft | 3.7654 | 2.7039 | 2.4235 |
| 15 | Name the companies who filed complex patent lawsuit against Microsoft? | 3.2246 | 2.9743 | 3.1846 |
| 16 | Name the companies who sued Microsoft? | 3.0043 | 2.5036 | 2.8641 |
| 17 | RealNetworks accused Microsoft of committing what crime? | 3.0744 | 2.3033 | 2.8341 |
| 18 | The court ruled against whom in cases involving Microsoft? | 2.9943 | 2.3933 | 2.8641 |
| 19 | The court sided with whom in cases by Microsoft? | 3.1345 | 2.3734 | 2.7540 |
| 20 | The court sided with whom in cases involving Microsoft? | 3.2747 | 2.3534 | 2.8040 |
| 21 | What are the companies engaged in cases by Vonage? | 2.9943 | 13.8657 | 3.4049 |
| 22 | What are the organizations engaged in cases with AT&T? | 3.3707 | 2.3934 | 2.7640 |
| 23 | When did AT&T file its case against Microsoft? | 3.7254 | 2.3834 | 2.7339 |
| 24 | When did Excite filed its case against Microsoft? | 3.4850 | 3.4550 | 2.6738 |
| 25 | When did Microsoft file its case against Excite? | 3.3949 | 2.5336 | 2.6037 |
| 26 | When was the case by AT&T against Microsoft resolved? | 18.3106 | 2.7940 | 3.3849 |
| 27 | When was the case by RealNetworks against Microsoft resolved? | 3.3564 | 2.3133 | 2.7840 |
| 28 | When was the case by Vonage against AT&T closed? | 3.9813 | 2.5336 | 2.7940 |
| 29 | When was the closing of the case against Microsoft? | 3.5710 | 2.3133 | 2.6438 |
| 30 | When was the closing of the case filed by Microsoft? | 2.8990 | 2.5136 | 2.6037 |
| 31 | When was the filing of the case against Excite by Microsoft? | 3.4537 | 2.3334 | 2.7740 |
| 32 | When was the filing of the case against Microsoft by RealNetworks? | 3.6785 | 2.3033 | 3.0844 |
| 33 | When was the filing of the case against Microsoft? | 3.7143 | 2.3133 | 2.5737 |
| 34 | When was the lawsuit against Microsoft by RealNetworks concluded? | 3.3325 | 2.3033 | 2.8541 |
| 35 | When was the ruling of the conspiracy case against Ira Zar? | 3.0153 | 2.3634 | 2.7540 |
| 36 | Which company sues Microsoft? | 3.8917 | 2.5837 | 2.5437 |
| 37 | Which judge presided the ruling of the case by RealNetworks against Microsoft? | 3.5123 | 2.5236 | 2.6238 |
| 38 | Who chaired the closing of the case by RealNetworks against Microsoft? | 3.9992 | 2.5336 | 2.6138 |
| 39 | Who did the court side with in the case against Microsoft? | 4.1729 | 2.3233 | 2.8241 |
| 40 | Who files the complex patent lawsuit against Microsoft? | 3.8917 | 2.3434 | 2.7640 |
| 41 | Who is the judge presiding the case against AT&T by Vonage? | 4.2104 | 2.7339 | 2.6138 |
| 42 | Who preside over the case against Microsoft? | 3.8982 | 2.3563 | 2.7857 |
| 43 | Who presided the closing of the spamming case against Howard Carmark? | 4.0749 | 2.4931 | 2.8675 |
| 44 | Who sues Microsoft? | 3.5649 | 2.4695 | 2.9038 |
| 45 | Who won in the case by Vonage against AT&T? | 3.7847 | 2.3549 | 3.1828 |

# Appendix B – Responses from AnswerBus

| | Question | Response |
|---|---|---|
| 1 | AT&T accused Microsoft of committing what crime? | I found no answer for your question. |
| 2 | Did Vonage initiate any legal actions against Microsoft? | I found no answer for your question. |
| 3 | How many companies were involved in cases with Microsoft? | I found no answer for your question. |
| 4 | How many times did AT&T lose in cases initiated by it? | I'm not sure what you mean. |
| 5 | How many times did Microsoft win in cases against it? | I'm not sure what you mean. |
| 6 | Ira Zar was accused of what? | Sorry, I don't know the answer to your question. |
| 7 | Is it true that the judge James Ware has sided with Microsoft in an antitrust lawsuit by RealNetworks? | I'm not sure what you mean. |
| 8 | List of judges who preside over cases file by Microsoft? | I found no answer for your question. |
| 9 | List the companies sued by AT&T? | Following is a list of those being sued by various record companies, and the district court in which the suit was filed. |
| 10 | List the companies who sued Microsoft? | • The United States Department of Justice (DOJ) and the 19 US states who successfully sued Microsoft over Anti-Trust violations have requested that the US court spilt Microsoft in to two separate companies.<br>• None of these companies sued Microsoft when it began giving away a free codec in Windows Media Player, because they could not be sure their technology was in there. |
| 11 | List the judges presiding cases file against Microsoft? | I'm not sure what you mean. |
| 12 | List the judges presiding cases involving Microsoft? | Sorry, I don't know the answer to your question. |
| 13 | Microsoft sued whom? | • Microsoft sued over media player By Byron Acohido, USA TODAY SEATTLE RealNetworks on Thursday accused Microsoft of anti-competitive tactics to restrict consumer choice in the PC media player market.<br>• Microsoft Sued Over HomeAdvisor Site Microsoft Corp. and Moore USA in Philadelphia Wednesday claiming the companies illegally used his patented point and click real estate property locator in two of their offerings.<br>• This time Microsoft is being sued over "Yourself! |
| 14 | Name the companies sued by Microsoft | • Microsoft Sued Over HomeAdvisor Site Microsoft Corp. and Moore USA in Philadelphia Wednesday claiming the companies illegally used his patented point and click real estate property locator in two of their offerings.<br>• None of these companies sued Microsoft when it began giving away a free codec in Windows Media Player, because they could not be sure their technology was in there.<br>• BTG plc on Wednesday sued Microsoft and Apple over alleged patent violations that cover the companies' Web-based software update services. |
| 15 | Name the companies who filed complex patent lawsuit against Microsoft? | • The companies have amicably resolved the pending patent-infringement lawsuit based on mutually satisfactory terms that dismiss the outstanding litigation of the patent-related lawsuit originally filed last year by Timeline.<br>• 2 Oct 2003 - Los Angeles - a class-action lawsuit has been filed against Microsoft for myriad security problems and remedies too complex for the public to understand.<br>• The Burst.com lawsuit against Microsoft, originally filed in June of 2002, asserting patent, trade secret and antitrust claims, has been settled. |
| 16 | Name the companies who sued Microsoft? | • Microsoft Sued Over HomeAdvisor Site Microsoft Corp. and Moore USA in Philadelphia Wednesday claiming the companies illegally used his patented point and click real estate property locator in two of their offerings.<br>• None of these companies sued Microsoft when it began giving away a free codec in Windows Media Player, because they could not be sure their technology was in there.<br>• BTG plc on Wednesday sued Microsoft and Apple over alleged patent violations that cover the companies' Web-based software update services. |
| 17 | RealNetworks accused Microsoft of committing what crime? | Sorry, I don't know the answer to your question. |
| 18 | The court ruled against whom in cases involving Microsoft? | • We have found four antitrust cases involving arrangements in which a software program is tied to the purchase of a software platform--two district court cases and two appellate court cases, including one from this court.<br>• Thailand Law Forum, Thai Supreme Court overturns a lower court's decision in a landmark case involving the sale of pirated Microsoft software programs.<br>• Thai Supreme Court overturns lower court's decision in a landmark case involving the sale of pirated Microsoft software programs<br>• That is fundamentally what the court of appeals said in June of 1998 in the first case involving Microsoft.<br>• The Justice Department was dealt a major setback in its case when a federal appeals court ruled earlier this year that Microsoft's inclusion of Internet Explorer in the Windows operating system was not an antitrust violation.<br>• Moreover, in a similar case about 15 months ago, the Appeals Court ruled that Microsoft had the right to include whatever it wanted in the PC Operating System, including its browser.<br>• After hearing the case, the Wuhan court ruled that anyone can send Microsoft corrected mistakes, and that this kind of conduct can not be regarded as a trade secret.<br>• The courts ruled that the use of dirty words in a sincere and honest book did not make the book dirty. Since the 1950s many obscenity cases involving books, magazines, and film have been brought before the Supreme Court.<br>• In those cases where the defendant claimed that use of the tied good made the tying good more valuable to users, the Court ruled that the same result could be achieved via quality standards for substitutes of the tied good.<br>• The Verizon and Rogers Cable cases, along with an earlier Canadian case involving an attempted class action suit against Microsoft, emphasize the importance of contractual certainty. |
| 19 | The court sided with whom in cases by Microsoft? | • Clearly, we are pleased that the Appeals Court unanimously sided with Microsoft and rejected the government's invitation to delay this process," said Jim Cullinan, a Microsoft spokesman, from company headquarters in Redmond, Washington.<br>• Lindows had changed its product names to Linspire in Europe last year, when a European court sided with Microsoft over cases brought up in several European countries.<br>• The appeals court sided with Microsoft stating it would decide the question without waiting for Jackson.<br>• A three-judge panel of the circuit court has sided with Microsoft (MSFT) in a related antitrust case also brought by the Justice Department.<br>• Microsoft appealed that injunction, and the court of appeals sided with the company. |
| 20 | The court sided with whom in cases involving Microsoft? | • We have found four antitrust cases involving arrangements in which a software program is tied to the purchase of a software platform--two district court cases and two appellate court cases, including one from this court.<br>• In one of the few cases where a victimized company had the resources to sue Microsoft, Stac proved in court that Microsoft had stolen their technology.<br>• appeals court, which has sided with Microsoft in earlier rulings, and would contribute "to a quick and effective resolution of the case," Klein told a news conference.<br>• Not only did the appellate judges say the first judge erroneously issued the injunction against Microsoft, but also that he should not have appointed a special master to advise the court about technical issues involving the case.<br>• Just two days before the launch of Windows 98, a federal appeals court has sided with Microsoft in part of the government's antitrust case against the company.<br>• The PayPal case, along with two other earlier U.S. cases involving disputes with America Online, sided with consumers by not enforcing company-drafted clauses that mandated arbitration or court appearances in far-off jurisdictions.<br>• In W two recent Fairfield County cases involving eminent domain, the stare Supreme Court sided with businesses m Stamford and Bridgeport.<br>• Both a federal judge and a federal appeals court have previously sided with the Milford school district, but other courts -- in other cases -- have upheld religious groups' rights to use public buildings for meetings.<br>• The decision stands in contrast to some court cases, such as that involving LucentSucks.com, in which courts have ruled that such addresses constitute fair comment.<br>• The Verizon and Rogers Cable cases, along with an earlier Canadian case involving an attempted class action suit against Microsoft, emphasize the importance of contractual certainty. |
| 21 | What are the companies engaged in cases by Vonage? | I found no answer for your question. |
| 22 | What are the organizations engaged in cases with AT&T? | Sorry, I don't know the answer to your question. |
| 23 | When did AT&T file its case against Microsoft? | • As was anticipated by Judge Motz, who granted a two-week stay (until February 4) to give the appellate court time to consider Microsoft's appeal, Microsoft has filed a notice with the district court that it does intend to appeal.<br>• Burst.com creator of video and audio delivery software for IP networks, filed its lawsuit against Microsoft in June 2002, claiming that Microsoft stole Internet-based video patented technology for its Windows Media Player product.<br>• Microsoft filed 17 lawsuits againts piracy<br>• WinInsider | Microsoft filed 17 lawsuits againts piracy<br>• Burst.com filed its lawsuit against Microsoft : 15 March 2005<br>• In 1998 the United States and a group of states filed a civil complaint against Microsoft alleging antitrust violations.<br>• The Justice Department and 20 states filed their antitrust lawsuit against Microsoft in May 1998, but only 19 states pursued the case.<br>• [January 27, 2004] Applications for patents related to XML processing have been filed by Microsoft with various patent offices worldwide.<br>• One of the key pillars of this initiative was kicked off today as Microsoft announced it has filed 13 civil suits in the U.S. |
| 24 | When did Excite filed its case against Microsoft? | Sorry, I don't know the answer to your question. |
| 25 | When did Microsoft file its case against Excite? | I'm not sure what you mean. |
| 26 | When was the case by AT&T against Microsoft resolved? | • The April agreement between Sun and Microsoft resolved several legal issues that called for the companies to work together over the next 10 years.<br>• and Microsoft last April resolved their legal issues in a $1.6 billion settlement that calls for the two companies to work together over the next 10 years.<br>• A problem with Operations Console when using Service Pack 4 for Microsoft Windows 2000 is documented in APAR SE11182. Microsoft resolved this problem with a hotfix. |
| 27 | When was the case by RealNetworks against Microsoft resolved? | Sorry, I don't know the answer to your question. |
| 28 | When was the case by Vonage against AT&T closed? | • Vonage VoIP Forum - Voice Over Internet Phone Company Closed $105 Million In Funding<br>• Voice-Over-Internet Phone Company Vonage Said Wednesday That It Has Closed A $105 Million Round Of Funding<br>• Voice-over-Internet phone company Vonage said Wednesday that it has closed a $105 million round of funding.<br>• Vonage closed $105 million in a fourth round of venture funding, bring total investment in the company to $208 million.<br>• In February 2004, Vonage closed a $40.0 million series C financing round led by 3i and co-led by Meritech Capital Partners.<br>• To fuel its expansion, Vonage closed a $35 million investment round in late November led by New Enterprise Associates, bringing the total capital raised by the company to $65.3 million. |

| 29 | When was the closing of the caset against Microsoft? | Sorry, I don't know the answer to your question. |
|---|---|---|
| 30 | When was the closing of the case filed by Microsoft? | I found no answer for your question. |
| 31 | When was the filing of the case against Excite by Microsoft? | I'm not sure what you mean. |
| 32 | When was the filing of the case against Microsoft by RealNetworks? | Sorry, I don't know the answer to your question. |
| 33 | When was the filing of the case against Microsoft? | <ul><li>The breakdown of financials by division was published for the first time in Microsoft's Form 10-Q filing to the Securities Exchange Commission, presumably as a side-effect of corporate America's attempt at a post-Enron clean-up.</li><li>Download Electronic Filing Compliance Program (Microsoft Access 2000 or 2002 Required)</li><li>Microsoft's 118-page appellate filing this week raises a number of issues, both technical and legal, that will frame the next phase of its antitrust case.</li><li>We believe that Microsoft's share of server units grew modestly in fiscal 2004, while Linux distributions rose slightly faster on an absolute basis," the filing reads.</li><li>Consider the following two sentences, taken verbatim from the filing: "Since the release of Internet Explorer 1.0 in July 1995, Microsoft has distributed every version of Windows with Internet Explorer included.</li></ul> |
| 34 | When was the lawsuit against Microsoft by RealNetworks concluded? | I'm not sure what you mean. |
| 35 | When was the ruling of the conspiracy case against Ira Zar? | Now, former CA chief financial officer Ira Zar faces as many as 20 years in prison, and former senior financial executives David Rivard and David Kaplan face as many as 10 years in prison for their part in the fraud conspiracy. |
| 36 | Which company sues Microsoft? | <ul><li>South Korean Company Sues Microsoft - South Korea sues Microsoft!</li><li>Wired News: Mouse Company Sues Microsoft Welcome to Wired News. Skip directly to: Search Box, Section Navigation, Content.</li><li>Mouse Company Sues Microsoft</li><li>Trackbacks and Pingbacks for Anti-piracy company sues Microsoft on CNET News.com</li></ul> |
| 37 | Which judge presided the ruling of the case by RealNetworks against Microsoft? | <ul><li>Grimaldi, Microsoft Judge Says Ruling at Risk, Wash.</li><li>A month ago, lawyers for the accused ISV ElcomSoft asked a US District Judge Ronald Whyte, who presided over the Sun-Microsoft Java suit, to drop the case because the DMCA is unconstitutional.</li></ul> |
| 38 | Who chaired the closing of the case by RealNetworks against Microsoft? | Sorry, I don't know the answer to your question. |
| 39 | Who did the court side with in the case against Microsoft? | LXer: Jeremy Allison has history on his side...: Microsoft Court Loss Might Not Help Open Source, Samba Leader Says |
| 40 | Who files the complex patent lawsuit against Microsoft? | <ul><li>2 Oct 2003 - Los Angeles - a class-action lawsuit has been filed against Microsoft for myriad security problems and remedies too complex for the public to understand.</li><li>Unfortunately, almost immediately after signing the agreement with us, Microsoft instituted a lawsuit claiming extremely broad rights to sublicense the patent(s), Osenbaugh continued.</li><li>The dispute goes back to May 2001, when AT T sued Microsoft for infringing on a patent related to a compression technology used to reduce the size of digital speech files.</li><li>Microsoft settles most of patent lawsuit with AT T</li><li>While it did leap back into prominence, it did so not through technology but by prevailing in the settlement of a patent lawsuit against Microsoft, to the tune of US $440 Million -- thereby giving its new owners a cash profit.</li></ul> |
| 41 | Who is the judge presiding the case against AT&T by Vonage? | Sorry, I don't know the answer to your question. |
| 42 | Who preside over the case against Microsoft? | <ul><li>Perhaps most important, from Microsoft's standpoint, is that Judge Jackson will not preside over future proceedings in the case.</li><li>Microsoft Chairman Bill Gates will preside over the Oct. 24 event, which will take place in Central Park.</li></ul> |
| 43 | Who presided the closing of the spamming case against Howard Carmark? | I'm not sure what you mean. |
| 44 | Who sues Microsoft? | <ul><li>Former Smartphone Maker Sues Microsoft</li><li>Pulp Fiction writer sues Microsoft over virtual yoga</li><li>Sun Sues Microsoft For Breach Of Java Contract</li><li>RealNetworks sues Microsoft on antitrust</li><li>PC Plus - Netscape sues Microsoft</li><li>Novell sues Microsoft over WordPerfect</li></ul> |
| 45 | Who won in the case by Vonage against AT&T? | <ul><li>One thing Vonage won t do is mention VOIP to its new prospective markets.</li><li>Texas Instruments has won Vonage's blessing as its preferred provider of VoIP silicon and software.</li><li>I've never had a PPV call-in problem with Vonage -- that type of call is only 9600 baud and seems to get thru even when a long, large download won't (I use serial PPP for the latter, and recommend it).</li><li>At home vonage won't go throuhg the tunnel unless it is behind the computer that has established the tunnel.</li><li>I am not liking Vonage too much at this point, I don't use my ``land line'' enough to warrent dealing with a company that won't let me choose the phone I would like to use.</li></ul> |

# Appendix C – Responses from START

| | Question | Response |
|---|---|---|
| 1 | AT&T accused Microsoft of committing what crime? | I don't know what crime AT&T accused Microsoft because it commits. |
| 2 | Did Vonage initiate any legal actions against Microsoft? | The word VONAGE may be misspelled. |
| 3 | How many companies were involved in cases with Microsoft? | I don't have this information. |
| 4 | How many times did AT&T lose in cases initiated by it? | I do not understand this sentence. Could you try again? |
| 5 | How many times did Microsoft win in cases against it? | Unfortunately, I don't know if Microsoft won in cases against itself. |
| 6 | Ira Zar was accused of what? | I do not understand this sentence. Could you try again? |
| 7 | Is it true that the judge James Ware has sided with Microsoft in an antitrust lawsuit by RealNetworks? | The word SIDED may be misspelled |
| 8 | List of judges who preside over cases file by Microsoft? | You asked me to add the following assertion to my knowledge base: LIST OF JUDGES WHO PRESIDE OVER CASES FILE BY MICROSOFT. At present, however, remote users are only allowed to ask questions. Please send mail to Boris@csail.mit.edu for more information. |
| 9 | List the companies sued by AT&T? | Unfortunately, I don't have that information. |
| 10 | List the companies who sued Microsoft? | I don't have that information. |
| 11 | List the judges presiding cases file against Microsoft? | Unfortunately, I don't have that information. |
| 12 | List the judges presiding cases involving Microsoft? | I don't have that information. |
| 13 | Microsoft sued whom? | Sorry, I don't have that information. |
| 14 | Name the companies sued by Microsoft | I don't have this information. |
| 15 | Name the companies who filed complex patent lawsuit against Microsoft? | Sorry, I don't have this information. |
| 16 | Name the companies who sued Microsoft? | Unfortunately, I don't have this information. |
| 17 | RealNetworks accused Microsoft of committing what crime? | I don't know what crime RealNetworks accused Microsoft because it commits. |
| 18 | The court ruled against whom in cases involving Microsoft? | You asked me to add the following assertion to my knowledge base: THE COURT RULED AGAINST WHOM IN CASES INVOLVING MICROSOFT. At present, however, remote users are only allowed to ask questions. Please send mail to Boris@csail.mit.edu for more information. |
| 19 | The court sided with whom in cases by Microsoft? | The word SIDED may be misspelled |
| 20 | The court sided with whom in cases involving Microsoft? | The word SIDED may be misspelled |
| 21 | What are the companies engaged in cases by Vonage? | The word VONAGE may be misspelled |
| 22 | What are the organizations engaged in cases with AT&T? | Unfortunately, I don't have that information. |
| 23 | When did AT&T file its case against Microsoft? | I don't know when AT&T filed its case against Microsoft. |
| 24 | When did Excite filed its case against Microsoft? | I don't understand. Please try another sentence. |
| 25 | When did Microsoft file its case against Excite? | Please try another sentence. I didn't understand this one. |
| 26 | When was the case by AT&T against Microsoft resolved? | Sorry, I just don't know when the case by AT&T against Microsoft was resolved. |
| 27 | When was the case by RealNetworks against Microsoft resolved? | I don't know when the case by RealNetworks against Microsoft was resolved. |
| 28 | When was the case by Vonage against AT&T closed? | The word VONAGE may be misspelled |
| 29 | When was the closing of the caset against Microsoft? | The word CASET may be misspelled |
| 30 | When was the closing of the case filed by Microsoft? | I don't understand. Won't you try to say it in another way? |
| 31 | When was the filing of the case against Excite by Microsoft? | I'm not quite sure what you mean by that. Perhaps you could try it a bit differently. |
| 32 | When was the filing of the case against Microsoft by RealNetworks? | I did not understand the word "realnetworks". Please try using a different word. |
| 33 | When was the filing of the case against Microsoft? | I'm not following what you're saying. Please try it a different way. |
| 34 | When was the lawsuit against Microsoft by RealNetworks concluded? | I don't know the answer. Sorry. |
| 35 | When was the ruling of the conspiracy case against Ira Zar? | Please try to say this sentence in a different way. I don't understand it. |
| 36 | Which company sues Microsoft? | Unfortunately, I wasn't told which company sues Microsoft. |
| 37 | Which judge presided the ruling of the case by RealNetworks against Microsoft? | I don't know which judge presided the ruling of the case against Microsoft. |
| 38 | Who chaired the closing of the case by RealNetworks against Microsoft? | Unfortunately, I don't know who chaired the closing of the case against Microsoft. |
| 39 | Who did the court side with in the case against Microsoft? | I'm not quite sure what you mean by that. Perhaps you could try it a bit differently. |
| 40 | Who files the complex patent lawsuit against Microsoft? | Unfortunately, I don't know who files the complex patent lawsuit against Microsoft. |
| 41 | Who is the judge presiding the case against AT&T by Vonage? | The word VONAGE may be misspelled. |
| 42 | Who preside over the case against Microsoft? | Sorry, I just don't know who presides over the case against Microsoft. |
| 43 | Who presided the closing of the spamming case against Howard Carmark? | The word SPAMMING may be misspelled |
| 44 | Who sues Microsoft? | I don't know. |
| 45 | Who won in the case by Vonage against AT&T? | The word VONAGE may be misspelled |

# Appendix D – Responses from NaLURI

| | Question | Response |
|---|---|---|
| 1 | AT&T accused Microsoft of committing what crime? | • Trademark lawsuit<br>• Complex patent lawsuit |
| 2 | Did Vonage initiate any legal actions against Microsoft? | The entities in question exist independently and not together in the event legal_proceeding. so we do not have information about the legal_proceeding defendant which Microsoft. |
| 3 | How many companies were involved in cases with Microsoft? | 5 |
| 4 | How many times did AT&T lose in cases initiated by it? | None |
| 5 | How many times did Microsoft win in cases against it? | 1 |
| 6 | Ira Zar was accused of what? | Conspiracy and fraud |
| 7 | Is it true that the judge James Ware has sided with Microsoft in an antitrust lawsuit by RealNetworks? | Yes, it is true |
| 8 | List of judges who preside over cases file by Microsoft? | John Wayne chaired the filing of the case where Microsoft is the plaintiff |
| 9 | List the companies sued by AT&T? | Microsoft |
| 10 | List the companies who sued Microsoft? | RealNetworks and AT&T |
| 11 | List the judges presiding cases file against Microsoft? | • Rob Lowe chaired the filing of a case<br>• Judge James Ware chaired the resolution of a case<br>• Judge William Pauley III of U.S. District chaired the resolution of a case |
| 12 | List the judges presiding cases involving Microsoft? | • Rob Lowe chaired the filing of the case where Microsoft is the defendant<br>• John Wayne chaired the filing of the case where Microsoft is the plaintiff<br>• Judge James Ware chaired the resolution of the case where Microsoft is the defendant<br>• Judge William Pauley III U.S. District chaired the resolution of the case where Microsoft is the defendant |
| 13 | Microsoft sued whom? | Excite |
| 14 | Name the companies sued by Microsoft | Excite |
| 15 | Name the companies who filed complex patent lawsuit against Microsoft? | AT&T |
| 16 | Name the companies who sued Microsoft? | RealNetworks and AT&T |
| 17 | RealNetworks accused Microsoft of committing what crime? | Antitrust lawsuit |
| 18 | The court ruled against whom in cases involving Microsoft? | • Court ruled against Microsoft in a case by AT&T<br>• Court ruled against Microsoft in a case against Excite.<br>• Court ruled against RealNetworks in a case by Microsoft. |
| 19 | The court sided with whom in cases by Microsoft? | Excite |
| 20 | The court sided with whom in cases involving Microsoft? | • Court sided with Excite in a case by Microsoft.<br>• Court sided with AT&T in a case against Microsoft.<br>• Court sided with Microsoft in a case by RealNetworks. |
| 21 | What are the companies engaged in cases by Vonage? | AT&T |
| 22 | What are the organizations engaged in cases with AT&T? | Vonage and Microsoft |
| 23 | When did AT&T file its case against Microsoft? | Filing took place on 2002 |
| 24 | When did Excite filed its case against Microsoft? | There is no such filing event involving Excite as plaintiff. |
| 25 | When did Microsoft file its case against Excite? | Filing took place on 2003 |
| 26 | When was the case by AT&T against Microsoft resolved? | Resolution took place on Monday 9 February 2004 |
| 27 | When was the case by RealNetworks against Microsoft resolved? | Resolution took place on Wednesday 11 March 2004 |
| 28 | When was the case by Vonage against AT&T closed? | Resolution took place on Monday 19 February 2003 |
| 29 | When was the closing of the caset against Microsoft? | There are some spelling errors in the question. When was the closing of the **caset** against Microsoft |
| 30 | When was the closing of the case filed by Microsoft? | Filing took place on 2003 against Excite |
| 31 | When was the filing of the case against Excite by Microsoft? | Filing took place on 2003 |
| 32 | When was the filing of the case against Microsoft by RealNetworks? | There is no such filing event involving RealNetworks as plaintiff. |
| 33 | When was the case against Microsoft? | Filing took place on 2002 by AT&T |
| 34 | When was the lawsuit against Microsoft by RealNetworks concluded? | Resolution took place on Wednesday 11 March 2004 |
| 35 | When was the ruling of the conspiracy case against Ira Zar? | Resolution took place on Wednesday 7 April 2004 |
| 36 | Which company sues Microsoft? | RealNetworks and AT&T |
| 37 | Which judge presided the ruling of the case by RealNetworks against Microsoft? | Judge James Ware chaired the resolution of the case |
| 38 | Who chaired the closing of the case by RealNetworks against Microsoft? | Judge James Ware chaired the resolution of the case |
| 39 | Who did the court side with in the case against Microsoft? | AT&T |
| 40 | Who files the complex patent lawsuit against Microsoft? | AT&T |
| 41 | Who is the judge presiding the case against AT&T by Vonage? | Judge George Wilson chaired the resolution of the case |
| 42 | Who preside over the case against Microsoft? | • Rob Lowe chaired the filing of the case where Microsoft is the defendant.<br>• Judge James Ware chaired the resolution of the case where Microsoft is the defendant.<br>• Judge William Pauley III U.S. District chaired the resolution of the case where Microsoft is the defendant. |
| 43 | Who presided the closing of the spamming case against Howard Carmark? | Attorney General Eliot Spitzer New York chaired the resolution of the case |
| 44 | Who sues Microsoft? | RealNetworks and AT&T |
| 45 | Who won in the case by Vonage against AT&T? | Vonage |